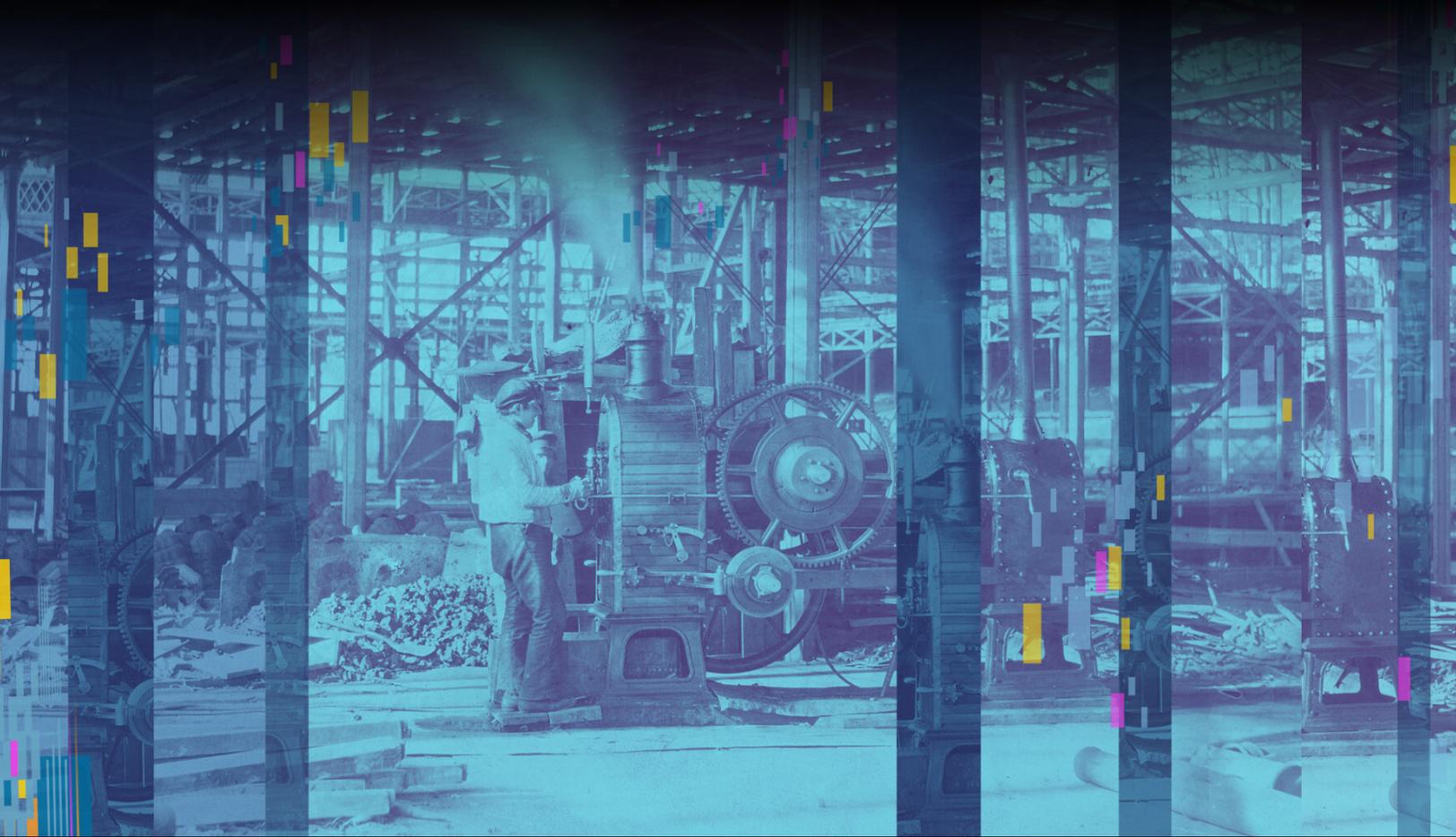

# AI Risk-Management Standards Profile for General-Purpose AI (GPAI) and Foundation Models

ANTHONY M. BARRETT | JESSICA NEWMAN | BRANDIE NONNECKE | NADA MADKOUR

DAN HENDRYCKS | EVAN R. MURPHY | KRYSTAL JACKSON | DEEPIKA RAMAN

**Version 1.1, January 2025**

For the latest public version of this document, see: https://cltc.berkeley.edu/publication/ai-risk-management-standards-profile-v1-1

For the QUICK GUIDE: An Introductory Resource for the AI Risk-Management Standards Profile for General-Purpose AI (GPAI) and Foundation Models, see: https://cltc.berkeley.edu/wp-content/uploads/2025/01/Berkeley-Profile-v1-1-Quick-Guide.pdf

For Retrospective Test Use of the AI Risk-Management Standards Profile for General-Purpose AI (GPAI) and Foundation Models V1.1 Draft Guidance, see: https://cltc.berkeley.edu/wp-content/uploads/2025/01/Berkeley-Retrospective-Test-Use-of-Profile-v1-1.pdf

For Mapping of the AI Risk-Management Standards Profile for General-Purpose AI (GPAI) and Foundation Models V1.1 Guidance to Key Standards and Regulations, see: https://cltc.berkeley.edu/wp-content/uploads/2025/01/Berkeley-Mapping-of-Profile-Guidance-v1-1-to-Key-Standards-and-Regulations.pdf

# AI Risk-Management Standards Profile for General-Purpose AI (GPAI) and Foundation Models

**ANTHONY M. BARRETT**[†] • **JESSICA NEWMAN**[†] • **BRANDIE NONNECKE**[††] • **NADA MADKOUR**[†]

**DAN HENDRYCKS**[†††] • **EVAN R. MURPHY**[†] • **KRYSTAL JACKSON**[†] • **DEEPIKA RAMAN**[†]

† AI Security Initiative, Center for Long-Term Cybersecurity, UC Berkeley

†† CITRIS Policy Lab, CITRIS and the Banatao Institute; Goldman School of Public Policy, UC Berkeley

††† Berkeley AI Research Lab, UC Berkeley

All affiliations listed are either current, or were during main contributions to this work or a previous version.

**Version 1.1, January 2025**

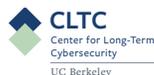

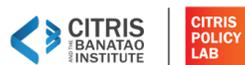

## ABSTRACT

Increasingly multi-purpose AI models, such as cutting-edge large language models or other "general-purpose AI" (GPAI) models, "foundation models," generative AI models, and "frontier models" (typically all referred to hereafter with the umbrella term "GPAI/foundation models" except where greater specificity is needed), can provide many beneficial capabilities but also risks of adverse events with profound consequences. This document provides risk-management practices or controls for identifying, analyzing, and mitigating risks of GPAI/foundation models. We intend this document primarily for developers of large-scale, state-of-the-art GPAI/foundation models; others that can benefit from this guidance include downstream developers of end-use applications that build on a GPAI/foundation model. This document facilitates conformity with or use of leading AI risk management-related standards, adapting and building on the generic voluntary guidance in the NIST AI Risk Management Framework and ISO/IEC 23894, with a focus on the unique issues faced by developers of GPAI/foundation models.

## NOTES ON THIS VERSION

Changes between this Version 1.1 Profile and the Version 1.0 Profile (Barrett, Newman et al. 2023) include:

- Terminology and scope refinements throughout this document
  - » Most notable is that most instances of "general purpose AI systems (GPAIS)" were changed to "GPAI/foundation models" to better reflect our greater relative focus on upstream GPAI models and foundation models, rather than downstream AI systems incorporating GPAI/foundation models.
- Additional resources for:
  - » Red-teaming and benchmark capability evaluations (Measure 1.1)
  - » Transparency (Measure 2.9) and documentation (Measure 3.1)
  - » Governance and policy tracking (Govern 1.1)
  - » Training data audits (Manage 1.3, Measure 2.8)
  - » Model weight protection (Measure 2.7)
- Added actions and resources from the NIST Generative AI Profile, NIST AI 600-1, released July 2024
- Expansion on risks:
  - » Manipulation and deception (Map 5.1)
  - » Sandbagging during hazardous-capabilities evaluations (Govern 2.1, Map 5.1)
  - » Situational awareness (Map 5.1)
  - » Socioeconomic and labor market disruption (Map 5.1)
  - » Possible intractability of removing backdoors (Map 5.1, Measure 2.7)
- In Roadmap in Appendix 3, updates on issues to address in future versions of Profile:
  - » Interpretability and explainability methods appropriate for architectures and scales of LLMs and other GPAI/foundation models
  - » Agentic AI systems
- New supporting documentation:
  - » **Profile Quick Guide**, a short introductory resource designed to complement the full profile (Barrett et al. 2025b). To access, please see: https://cltc.berkeley.edu/wp-content/uploads/2025/01/Berkeley-Profile-v1-1-Quick-Guide.pdf
  - » **Retrospective Test Use of Profile Guidance** document with testing on new foundation models (Barrett et al. 2025c). For the Profile V1.1, the Retrospective Test Use of Profile Guidance has been separated from the main Profile content. To access please see: https://cltc.berkeley.edu/wp-content/uploads/2025/01/Berkeley-Retrospective-Test-Use-of-Profile-v1-1.pdf
  - » **Mapping of Profile Guidance V1.1 to Key Standards and Regulations** document with added mappings to new regulations (e.g., the finalized EU AI Act) and commitments (e.g., the Frontier AI Safety Commitments) (Barrett et al. 2025a). For the Profile V1.1, the Mapping of Profile Guidance has been separated from the main Profile content. To access please see: https://cltc.berkeley.edu/wp-content/uploads/2025/01/Berkeley-Mapping-of-Profile-Guidance-v1-1-to-Key-Standards-and-Regulations.pdf

# Contents

# Executive Summary

Increasingly multi-purpose AI models, such as state-of-the-art large language models or other **"general purpose AI" (GPAI) models**, "**foundation models,**" generative AI models, and **"frontier models"** (typically all referred to hereafter with **the umbrella term "GPAI/foundation models"),** can provide many beneficial capabilities, but also risks of adverse events such as large-scale manipulation of people through AI model-generated misinformation or disinformation or other events with harmful impacts at societal scale.

This document provides an AI risk-management standards **Profile**, or a targeted set of risk-management practices or controls specifically for identifying, analyzing, and mitigating risks of GPAI/foundation models. This Profile is designed to complement the broadly applicable guidance in the NIST AI Risk Management Framework (AI RMF) or a related AI risk-management standard such as ISO/IEC 23894.

We intend this Profile primarily for use by **developers of large-scale, state-of-the-art GPAI/ foundation models**. For GPAI/foundation model developers, this Profile facilitates conformity with or use of leading AI risk management-related standards, and aims to facilitate compliance with relevant regulations such as the EU AI Act, especially for aspects related to GPAI/foundation models. (However, this Profile does not provide all guidance that may be needed for applications or AI systems incorporating GPAI/foundation models in particular industry sectors or applications.) Others who can benefit from the use of this guidance include: downstream developers of end-use applications or AI systems that build on a GPAI/foundation model; evaluators of GPAI/foundation models; and the regulatory community. This Profile can provide GPAI/ foundation model deployers, evaluators, and regulators with information useful for evaluating the extent to which developers of such AI models have followed relevant best practices. Normalizing the use of best practices such as those detailed in this Profile can help ensure developers of GPAI/foundation models can be competitive without compromising on practices for AI safety, security, accountability, and related issues. Ultimately, this Profile aims to help key actors in the value chains of increasingly general-purpose AI models and systems to achieve outcomes of maximizing benefits, and minimizing negative impacts, to individuals, communities, organizations, society, and the planet. That includes protection of human rights, minimization of negative environmental impacts, and prevention of adverse events with systemic or catastrophic consequences at societal scale.

The NIST AI RMF "core functions," or broad categories of activities, apply as appropriate across AI system lifecycles, and we provide corresponding guidance in related sections of this Profile: "Govern" (Section 3.1) for AI risk management process policies, roles, and responsibilities; "Map" (Section 3.2) for identifying AI risks in context; "Measure" (Section 3.3) for rating AI trustworthiness characteristics; and "Manage" (Section 3.4) for decisions on prioritizing, avoiding, mitigating, or accepting AI risks.

Users of this Profile should place high priority on the following risk management steps and corresponding Profile guidance sections. (Appropriately applying the Profile guidance for the following steps should be regarded as the baseline or minimum expectations for users of this Profile; users of this Profile can exceed the minimum expectations by also applying guidance in other sections.)

- **Check or update, and incorporate, each of the following high-priority risk management steps when making go/no-go decisions,** especially on whether to proceed on major stages or investments for development or deployment of cutting-edge large-scale GPAI/foundation models (Manage 1.1).

- **Take responsibility for risk assessment and risk management tasks for which your organization has access to information, capability, or opportunity to develop capability sufficient for constructive action, or that is substantially greater than others in the value chain** (Govern 2.1).
  - » We also recommend applying this principle throughout other risk assessment and risk management steps, and we refer to it frequently in other guidance sections.

- **Set risk-tolerance thresholds to prevent unacceptable risks** (Map 1.5).
  - » For example, the NIST AI RMF 1.0 recommends the following: "In cases where an AI system presents unacceptable negative risk levels — such as where significant negative impacts are imminent, severe harms are actually occurring, or catastrophic risks are present — development and deployment should cease in a safe manner until risks can be sufficiently managed" (NIST 2023a, p.8).

- **Identify reasonably foreseeable uses, misuses, and abuses for a GPAI/foundation model** (e.g., automated generation of toxic or illegal content or disinformation, or aiding with proliferation of cyber, chemical, biological, radiological, or nuclear weapons), and identify reasonably foreseeable potential impacts (e.g., to fundamental rights) (Map 1.1).

- **Identify whether a GPAI/foundation model could lead to significant, severe, or cata-strophic impacts,** e.g., because of correlated failures or errors across high-stakes deploy-ment domains, dangerous emergent behaviors or vulnerabilities, or harmful misuses and abuses (Map 5.1).

- **Use red-teams and adversarial testing** as part of extensive interaction with GPAI/founda-tion models to identify dangerous capabilities, vulnerabilities, or other emergent properties of such systems (Measure 1.1).

- **Track important identified risks** (e.g., vulnerabilities from data poisoning and other at-tacks or objectives mis-specification) even if they cannot yet be measured (Measure 1.1 and Measure 3.2).

- **Implement risk-reduction controls as appropriate** throughout a GPAI/foundation model lifecycle, e.g., independent auditing, incremental scale-up, red-teaming, structured access or staged release, and other steps (Manage 1.3, Manage 2.3, and Manage 2.4).

- **Incorporate identified AI system risk factors, and circumstances that could result in impacts or harms, into reporting and engagement with internal and external stake-holders** (e.g., when reporting to downstream developers, regulators, users, impacted communities, etc.) on the AI system as appropriate, e.g., using model cards, system cards, and other transparency mechanisms (Govern 4.2).

We also recommend: **Document the process used in considering risk mitigation controls, the options considered, and reasons for choices**. Documentation on many items should be shared in publicly available material such as system cards. Details on particular items, such as security vulnerabilities, can be responsibly omitted from public materials to reduce misuse potential, especially if available to auditors, Information Sharing and Analysis Organizations, or other parties as appropriate.

GPAI/foundation model-related risk topics and corresponding guidance sections in this Profile include the following. (Some of these topics overlap with others, in part because the guidance often involves iterative assessments for additional depth on issues identified at earlier stages.)

- Reasonably foreseeable impacts (Section 3.2, Map 1.1), including:
  - » To individuals, including impacts to health, safety, well-being, or fundamental rights;

- » To groups, including populations vulnerable to disproportionate adverse impacts or harms; and
- » To society, including environmental impacts.

- Significant, severe, or catastrophic harm factors (Section 3.2, Map 5.1), including:
    - » Correlated bias and discrimination;
    - » Impacts to societal trust or democratic processes;
    - » Correlated robustness failures;
    - » Potential for high-impact misuses, such as for cyber weapons, or chemical, biological, radiological, or nuclear (CBRN) weapons;
    - » Capability to manipulate or deceive humans in harmful ways; and
    - » Loss of understanding and control of an AI system in a real-world context.

- AI trustworthiness characteristics (Section 3.4, Measure 2), including:
    - » Safety, reliability, and robustness (Measure 2.5, Measure 2.6);
    - » Security and resiliency (Measure 2.7);
    - » Accountability and transparency (Measure 2.8);
    - » Explainability and interpretability (Measure 2.9);
    - » Privacy (Measure 2.10); and
    - » Fairness and bias (Measure 2.11).

Additional topics to address in future versions of the Profile are listed in Appendix 3.

# 1. Introduction and Objectives

## 1.1  KEY TERMS

Increasingly multi-purpose AI models, such as state-of-the-art large language models (LLMs), large multimodal language models (LMMs), or other "general-purpose AI" (GPAI) models, "foundation models," and generative AI models, can provide many beneficial capabilities but also risks of adverse events with consequences at societal scale.

We use these key terms as follows. (For additional terms and acronyms, see the Glossary.)

- **Foundation model or General-purpose AI model (GPAI/foundation model)**: "Any model that is trained on broad data (generally using self-supervision at scale) that can be adapted (e.g., fine-tuned) to a wide range of downstream tasks" (Bommasani et al. 2021, p. 3).
  - » We treat "**GPAI/foundation models" as an umbrella term that also includes frontier models and generative AI models**, except where we need to be more specific.
  - » Typically, a single large GPAI/foundation *model* plays a central role as a core part of a GPAI *system* that incorporates a GPAI/foundation model. (See GPAIS, below.)
    - A GPAI/foundation model often can serve as a GPAIS, especially if the GPAI/foundation model developer releases a GPAI/foundation model after adding elements such as instruction fine-tuning, a chatbot-style user interface, etc. Thus, many GPAI/foundation models such as GPT-3 can be regarded as a GPAIS.
    - Broadly applicable statements and guidance in this document about "AI systems" typically also apply to GPAI/foundation models, except where GPAI/foundation models are specifically excluded (e.g., statements about fixed-purpose AI systems).
  - » Our usage of the terms "general purpose AI model" and "general purpose AI system" is very similar to the corresponding terms in the EU AI Act (EP 2024), except that we do not exclude AI models used for research.
  - » Examples of foundation models include GPT-4, Claude 3, PaLM 2, LLaMA 2, and others.

- **Frontier model**: A cutting-edge, state-of-the-art, or highly capable GPAI or foundation model; such models also may possess hazardous or dual-use capabilities sufficient to pose severe risks to public safety. (See, e.g., Ganguli, Hernandez et al. 2022, Anderljung, Barnhart et al. 2023, and Microsoft 2023.)
  - » We treat **frontier models as the largest-scale, highest-capability subset of GPAI/ foundation models.** They are typically characterized by model size, training compute or data, and/or resulting capabilities that are above or near to industry-record thresholds. (See also "foundation model frontier" in the Glossary.)

» Our usage of the term "frontier model" approximately corresponds to dual-use foundation models, as defined by Executive Order 14110[1] (White House 2023c) and to GPAI models with systemic risk, as defined by the EU AI Act (EP 2024).

» Examples of frontier models: As of August 2024, models at or near the industry frontier include GPT-4o, Claude 3.5 Sonnet, Gemini 1.5, and Llama 3.1 405B.[2]

- **General-purpose AI system (GPAI or GPAIS):** "An AI system that can accomplish or be adapted to accomplish a range of distinct tasks, including some for which it was not intentionally and specifically trained" (Gutierrez et al. 2022, p. 22).

  » In currently available GPAIS, typically a single large GPAI/foundation model plays a central role as a core part of a GPAIS.

  » Examples of GPAIS include unimodal generative AI systems (e.g., GPT-3) and multimodal generative systems (e.g., DALL-E 3), as well as reinforcement-learning systems such as MuZero and AI systems with emergent capabilities. GPAIS *do not* include fixed-purpose AI systems trained specifically for tasks such as image classification or voice recognition (Gutierrez et al. 2022).

- **Generative AI**: "Any AI system whose primary function is to generate content" (Toner 2023).

  » We typically only use the term **"generative AI" to highlight issues specific to synthetic text (which can include software code), images, video, audio, or other synthetic media**. (In other documents, "generative AI" is often used in approximately the same way that we use the term "GPAI/foundation model.")

  » Examples of generative AI: "Typical examples of generative AI systems include image generators (such as Midjourney or Stable Diffusion), large language models or multimodal models (such as GPT-4, PaLM, or Claude), code generation tools (such as [GitHub] Copilot), or audio generation tools (such as VALL-E or resemble.ai)" (Toner 2023).

We intend our usage of the above terms to be broadly compatible with usage of the corresponding terms where applicable in the OECD classification framework (OECD 2022a, p. 64), EU AI Act (EP 2024), Executive Order 14110 (White House 2023c), and the Hiroshima Process

---

1    The Profile V1.1 and its supporting documents were drafted and finalized prior to the recession of Executive Order 14110 on the Safe, Secure, and Trustworthy AI on January 20, 2025.

2    Several AI companies committed to measures proposed at the July 2023 industry frontier, such as red-teaming and public reporting of societal risks when developing and releasing models more powerful than GPT-4 or other models (White House 2023a).

International Code of Conduct for Advanced AI Systems (G7 2023), though our focus in this document is primarily on the most broadly capable GPAI/foundation models.

The terminology used in this document has been refined to reflect focus on upstream AI "models" rather than downstream AI "systems" that incorporate a model.[3]

## 1.2 BACKGROUND AND PURPOSE OF THE PROFILE

GPAI/foundation models such as GPT-4, DALL-E 3, Gemini, Claude 3, and Llama 3 can serve as multi-purpose AI models underpinning many end-use applications. These increasingly powerful GPAI/foundation models are the focus of cutting-edge research. They also have several qualitatively distinct properties compared to the more common, narrower machine learning models, such as potential to be applied to many sectors at once, potential large-scale societal, environmental, security, and economic impacts, and emergent properties that can provide unexpected beneficial capabilities but also unexpected risks of adverse events[4] (Bommasani et al. 2021, Weidinger et al. 2021, Wei et al. 2022, Ganguli, Hernandez et al. 2022). These properties complicate the ways in which GPAI/foundation models can be governed, though many AI experts encourage their inclusion in regulatory and risk management frameworks (see, e.g., Gebru et al. 2023). It can be appropriate to carry out more in-depth risk assessment — with longer time horizons and at more points in the AI system life cycle — and to implement other, more extensive risk-mitigation controls, for GPAI/foundation models than for AI with more limited capabilities.

This document is designed to complement the broadly applicable guidance in the NIST AI Risk Management Framework or AI RMF (NIST 2023a) or a related AI risk management standard such as ISO/IEC 23894. This document provides an AI risk-management standards target Profile, with a set of risk-management practices or controls and target outcomes specifically for identifying, analyzing, and mitigating risks of GPAI/foundation models. This cross-sectoral

---

3    The changes in terminology are also intended to reflect current trends in AI development and governance, in which a single large GPAI/foundation model typically plays a central role as a core part of either a GPAIS or a relatively narrow-purpose end-use application. (We expect that costs of applying relevant parts of the guidance in this document would typically be lower for downstream developers of GPAIS and end-use applications than for upstream developers of GPAI/foundation models; see Section 1.3 for more.) However, if highly powerful GPAIS begin to be created by combining a number of smaller models rather than relying primarily on a single core GPAI/foundation model, then we may return to focusing on GPAIS as a more inclusive term in future versions of this document.

4    In some cases, emergent properties of large-scale models could have been observed as partially-emergent properties of smaller-scale models if different metrics had been used (Schaeffer et al. 2023). We believe this is an argument for working to identify capabilities and other key properties of large-scale models at an early or partially-emergent stage in smaller-scale models, when responses to identified emergent properties may be more feasible and effective. For more on this, see our guidance on incremental scale-up and testing models after each incremental scale-up, in this document under AI RMF Subcategory Manage 1.3.

Profile addresses important underlying risks and early-development risks of such technologies in a way that does not rely on great certainty about each specific end-use application of the technology. We have developed this as a community Profile in a multi-stakeholder process, integrating input and feedback on drafts from a range of stakeholders, including organizations developing large-scale GPAI/foundation models, and other organizations across industry, civil society, academia, and government.

While this Profile focuses mainly on GPAI/foundation models, we also aim to address the need for upstream model developers' pre-release risk assessments and evaluations to consider reasonably foreseeable affordances (e.g., tool access) of downstream AI systems incorporating a GPAI/foundation model. GPAI/foundation model providers are uniquely positioned in the value chain to anticipate and manage those risks.

GPAI/foundation model-related risk categories that we aim to address with the guidance in this document include:

- Risks stemming from the large scale and reach of GPAI/foundation models, resulting from their frequent place in the AI value chain as core models that many other systems build on and rely upon;

- Risks of misuse and abuse of GPAI/foundation models, resulting from their lowering barriers for malicious activities such as generating disinformation; and

- Risks of unexpected impacts of GPAI/foundation models, resulting from the emergent behaviors, vulnerabilities, and capabilities that are often found (and continue to be found) in state-of-the-art, large-scale GPAI/foundation models.

Guidance in this Profile for GPAI/foundation models is based in part on examples of assessments and/or risk management controls already implemented by market leaders such as DeepMind, OpenAI, and Hugging Face. For example, OpenAI's 2019 announcement of GPT-2 included enumeration of several categories of potential misuse cases (OpenAI 2019a), which apparently informed OpenAI's decisions on disallowed/unacceptable use-case categories of applications based on GPT-3 (OpenAI 2024a). DeepMind's 2021 announcement of the large language model Gopher, and 2022 announcement of the multi-modal and multi-task "generalist agent" Gato, also included consideration of potential misuse, safety risks, and mitigation (Rae et al. 2021; Weidinger et al. 2021; Reed et al. 2022). Hugging Face and BigScience's release of the BLOOM LLM included a Responsible AI License (RAIL) with usage restrictions disallowing various types of misuse (RAIL n.d., Contractor et al. 2022). The Partnership on AI has developed guidance on synthetic media, including on transparency and disclosure of generative AI outputs (PAI 2023a), and has developed protocols for responsible deployment of foundation models (PAI 2023b,c). The Frontier Model Forum has also announced plans to research and share best practices for development of highly capable foundation models or frontier models, including safety-related evaluations (Heath 2023). In addition, NIST organized a Generative AI Public Working Group, and published a NIST AI RMF profile specifically on generative AI (NIST 2023c).[5]

Some of the material in this Profile is adapted directly from our related work in the supporting document "Mapping of Profile Guidance V1.1 to Key Standards and Regulations" (Barrett et al. 2025a) , and other sections of Barrett et al. (2022). Some other material in Section 3 of this Profile consists of extended excerpts from the NIST AI RMF Playbook (NIST 2023b), highlighting the portions of the broadly applicable Playbook guidance that seem particularly valuable for GPAI/foundation model developers, in light of typical current GPAI/foundation model architectures and development practices.

## 1.3 INTENDED AUDIENCE AND USERS OF THE PROFILE

We intend this document primarily for use by **developers of large-scale, state-of-the-art GPAI/foundation models**; others who can benefit from use of this guidance include **downstream developers** of end-use applications or AI systems that build on a GPAI/foundation model, as well as **model evaluators and regulators**.

We believe that most AI systems could be readily identified as one of the following:

- **One of a few large-scale GPAI/foundation models**. These AI models (and especially the most broadly capable GPAI/foundation models) are the main focus of this Profile, with some corresponding costs for upstream GPAI/foundation model developers, but also corresponding risk-management benefits when employing the guidance in this Profile.

- **One of two types of AI systems that build on a GPAI/foundation model**: Either a GPAIS that incorporates a GPAI/foundation model, or a relatively narrow-purpose end-use application that incorporates a GPAI/foundation model. Some aspects of these end-use applications and GPAIS are constructively addressed by parts of the guidance in this Profile.

---

5     Our Berkeley GPAI/foundation model Profile effort is separate from, but aims to complement and inform the work of, other guidance development efforts such as the PAI Guidance for Safe Foundation Model Deployment (PAI 2023c) and the NIST Generative AI Profile (Autio et al. 2024).

Costs to downstream developers building applications or GPAIS on GPAI/foundation models would likely be minimal when employing relevant guidance in this Profile; only some guidance in this Profile would typically be relevant to them, and generally not the parts that would be most expensive to use.

- **One of many small-scale or stand-alone narrow-purpose systems that do not fall under definitions for GPAI/foundation models**, and are not within the scope of this Profile. We do not expect developers or deployers of these common AI systems to use this Profile for those AI systems, and thus we do not expect their costs to be substantially affected by this Profile.

As part of "developers of GPAI/foundation models," we aim to include all organizations and efforts developing such AI models, regardless of the organization size or type, and regardless of whether the organization only plans to make the AI model available to users inside the organization. (Many of the same risks, such as potential for misuse or abuse by whoever has access to the AI system, would be present to some degree for GPAI/foundation model development efforts in each of these cases.) Thus, we intend for the guidance in this document to be applicable as appropriate to:

- Open-source and open-weights GPAI/foundation model development efforts, as well as closed-source GPAI/foundation model development; and

- Research projects, and other GPAI/foundation models that a model developer does not plan to make available to users outside the organization, as well as GPAI/foundation models that a developer plans to put on the market.

## 1.4  BENEFITS OF THE PROFILE

### 1.4.1  Benefits of the Profile to Developers of GPAI/Foundation Models

This Profile provides developers of GPAI/foundation models with valuable risk-management best practices that can be applied to their unique issues. For example, the Profile provides guidance on sharing of responsibilities between (a) upstream developers that create GPAI/foundation models and offer those in a manner that allows many different end uses, and (b) downstream developers that build upon the GPAI/foundation model for specific end-use applications, or who develop AI systems using upstream model provider-supplied information that may not be customized for their own application area.

This document facilitates conformity with or use of leading AI risk management-related standards, adapting and building on the generic voluntary guidance in the NIST AI Risk Management Framework and ISO/IEC 23894, with a focus on the unique issues faced by developers of GPAI/foundation models. It also aims to facilitate compliance with relevant regulations, such as the EU AI Act, especially for aspects related to GPAI/foundation models. See, e.g., the Mapping of Profile Guidance V1.1 to Key Standards and Regulations (Barrett et al. 2025a) supporting document for mapping to relevant clauses of ISO/IEC 23894 and the EU AI Act.

Widespread norms for using best practices such as those detailed in this Profile can help ensure that developers of GPAI/foundation models can be competitive without compromising on practices for AI safety, security, accountability, and related issues.

## 1.4.2  Benefits of the Profile to Deployers, Evaluators, and Users

This Profile can provide deployers, evaluators, and users of GPAI/foundation models with increased awareness of the risks of such AI models and of best practices to use in addressing those risks. This document also can provide deployers, evaluators, and users of such AI models with information useful for evaluating the extent to which developers of such AI models have followed relevant best practices.

## 1.4.3  Benefits for Individuals, Society, and the Regulatory Community

Ultimately, this Profile aims to help key actors in the value chains of increasingly general-purpose AI systems to achieve outcomes of maximizing benefits, and minimizing negative impacts, to individuals, communities, organizations, society, and the planet. That includes protection of fundamental rights, minimization of negative environmental impacts, and prevention of adverse events with systemic or catastrophic consequences at societal scale. There are vital relationships between principles of fairness and protecting human rights, addressing risks to individuals and groups, and addressing large-scale systemic or catastrophic risks. Some types of risks to individuals or groups comprise significant, severe, or catastrophic risks via accumulation or correlation of risks across individuals. Managing risks of GPAI/foundation models should include appropriate protection of human rights, and consideration of populations vulnerable to disproportionate harms. Preventing catastrophe can also be an important part of preventing unfair outcomes; often the effects of catastrophe fall disproportionately on disadvantaged people. It is critical to ensure that communities that may use or be impacted by the AI systems are meaningfully involved throughout the AI lifecycle, with opportunities to provide feedback and report potential problems.

From a regulatory perspective, this document can be viewed as part of "soft law" norms and best practices that GPAI/foundation model developers and deployers would have incentives to follow as appropriate, and that regulators can consider when formulating relevant "hard law" regulations (see, e.g., Gutierrez et al. 2021).[6] We also aim to provide mapping to, and harmonization with, relevant standards (e.g., ISO/IEC 23894) and regulations (e.g., the EU AI Act). This would help to set norms for GPAI/foundation model risk-management practices and conformity across regulatory regimes.

## 1.5  LIMITATIONS AND CHALLENGES

This Profile has a number of limitations. Perhaps the most important is this document's primary focus on AI risk management considerations for developers of GPAI/foundation models. While GPAI/foundation models may be used directly in a broad range of settings, or downstream developers may create AI systems or applications for such settings that incorporate GPAI/foundation models, this Profile does not provide all guidance that might be needed in particular industry sectors or applications. This Profile also does not provide all guidance that might be needed by GPAI/foundation model developers on risk management topics not directly related to model development and deployment, such as on securing an organization's networking equipment or other information system components.

Another limitation is the relatively nascent state of best practices for developers of GPAI/foundation models. We have based our guidance on available literature, demonstrated industry practices, stakeholder input and feedback, and ultimately our own judgment. However, we expect that best practices in this area will continue to evolve substantially. At minimum, we expect that such further evolution will provide more detailed resources in a number of areas, which we aim to incorporate in later versions of this guidance, e.g., in annual updates.

Challenges in this guidance include tradeoffs between risks and benefits, and even between different sets of risks. One of the most challenging areas is open-source development and release, or closely related release strategies, such as open-weights release, where model weights are downloadable. There is great value in open-source software and various forms of transparency

---

6    As a related example, the US National Telecommunications and Information Administration (NTIA) made AI accountability policy recommendations that include US government procurement requirements for use of appropriate AI standards and risk management practices such as audits. NTIA included foundation models in its considerations. (NTIA 2024a) and access to AI systems, including for helping to ensure the safety and security of an AI system's intended users. However, providing direct access to a model's weights also can increase some types of risks, including risks of malicious misuse. GPAI/foundation model developers that provide open-weights access to their models, and other GPAI/foundation model developers that suffer a leak of model weights, will in effect be unable to shut down or decommission AI systems that others build using those model weights. This is a consideration that should be weighed against the benefits of open-weights models, especially for the largest-scale and most broadly capable models that pose the greatest risks of enabling severe harms, including malicious misuse intended to harm the public.

Many of the benefits of openness, such as review and evaluation from a broader set of stakeholders and greater ability to use a model, are possible to support either through transparency, engagement, or other openness mechanisms that do not require a model's parameter weights to become downloadable, or through smaller-scale and less broadly capable open-weights models (including open-source models that can provide greater transparency than models that are only open-weights). Thus, our profile guidance includes many transparency and access provisions, including under Govern 4.2 on reporting to internal and external stakeholders (e.g., to downstream developers, regulators, users, impacted communities, etc.) on the AI system as appropriate, e.g., using model cards, or system cards, and other transparency mechanisms. Another important part of our profile guidance (under Manage 2.4) is that GPAI/foundation model developers that plan to provide downloadable, fully open, or open-source access to their models should first use a staged-release approach (e.g., not releasing parameter weights until after an initial closed-source or structured-access release, when no substantial risks or harms have emerged over a sufficient time period), and should not proceed to a final step of releasing model parameter weights until a sufficient level of confidence in risk management has been established, including for safety risks and risks of misuse and abuse. (That level of confidence in safety would be particularly difficult to appropriately establish for the largest-scale or most capable models, and they should be given the greatest duration and depth of pre-release evaluations, as they are the most likely to have dangerous capabilities or other emergent properties that can take some time to discover.) Our recommended approach is also consistent with recommendations on deployment of frontier models, as provided by the Partnership on AI (PAI 2023c) and others. We believe this overall approach provides actionable guidance to address some of the greatest risks to the public associated with open-sourcing powerful AI models while also providing valuable transparency mechanisms, and allowing responsible open-sourcing of AI models.[7]

7    For more on risk management tradeoffs for release strategies for frontier and near-frontier open-weights and open-source models, see, e.g., Solaiman (2023), Seger et al. (2023), PAI (2023c), Kapoor et al. (2024), Bateman et al. (2024), and NTIA (2024b).

# 2. Overview of Profile Components and How to Use Profile

## 2.1 BASICS

We intend for this Profile to be used in conjunction with the NIST AI RMF (NIST 2023a) and AI RMF Playbook (NIST 2023b), or an approximately equivalent set of AI risk management guidance documents, or an AI risk management framework or standard such as ISO/IEC 23894. In addition, we generally assume the use of appropriate guidance for risk topics not specific to AI, such as the NIST Cybersecurity Framework or ISO/IEC 27001, for broadly applicable information system security management guidance.[8]

It also can be appropriate to combine this Profile with another resource that provides supplemental guidance on particular industry sectors or applications for use-case-specific risks, metrics, and controls. (This would be most appropriate for downstream developers focused on building on or applying GPAI/foundation models for particular industry sectors or use cases.)

The AI RMF "core functions," or broad categories of activities, apply as appropriate across AI system lifecycles, and we provide corresponding guidance in related sections of this Profile:

- "Govern" (Section 3.1) for AI risk management process policies, roles, and responsibilities;
- "Map" (Section 3.2) for identifying AI risks in context;
- "Measure" (Section 3.3) for rating AI trustworthiness characteristics; and
- "Manage" (Section 3.4) for decisions on prioritizing, avoiding, mitigating, or accepting AI risks.

NIST (2023a) organizes high-level functions into categories and subcategories of activities and outcomes. In addition, NIST provides more detailed guidance in a companion Playbook resource document (NIST 2023b).

Our usage of the terms "should" and "can" in the guidance in Section 3 of this document is as follows: "should" indicates our recommendation and "can" indicates something is possible.[9]

---

8      If specific guidance suggested in this document has become obsolete, then use analogous or related up-to-date guidance instead of or in addition to the older guidance.

9      This is broadly consistent with usage by ISO and other standards organizations. See, e.g., ISO (n.d.).

## 2.2  IMPACT AREAS, HARM FACTORS, AND TRUSTWORTHINESS CHARACTERISTICS

GPAI/foundation model-related risk topics and corresponding guidance sections in this Profile include the following topics. (Some of these topics overlap with others, in part because the guidance often involves iterative assessments for additional depth on issues identified at earlier stages.)[10]

- Reasonably foreseeable impacts (Section 3.2, Map 1.1), including:
    - » To individuals, including impacts to health, safety, well-being, or fundamental rights;
    - » To groups, including populations vulnerable to disproportionate adverse impacts or harms; and
    - » To society, including environmental impacts.

- Significant, severe, or catastrophic harm factors (Section 3.2, Map 5.1), including:
    - » Correlated bias and discrimination;
    - » Impacts to societal trust or democratic processes;
    - » Correlated robustness failures;
    - » Capability to provide information for weaponization (e.g. CBRN, cyber);
    - » Capability to manipulate or deceive humans in harmful ways; and
    - » Loss of understanding and control of an AI system in a real world context (e.g., ability to escape a sandbox and replicate on another computational system).

- AI trustworthiness characteristics (Section 3.4, Measure 2), including:
    - » Safety, reliability, and robustness (Measure 2.5, Measure 2.6);
    - » Security and resiliency (Measure 2.7);
    - » Accountability and transparency (Measure 2.8);
    - » Explainability and interpretability (Measure 2.9);
    - » Privacy (Measure 2.10); and
    - » Fairness and bias (Measure 2.11).

Additional topics to address in future versions of the Profile are listed in Appendix 3.

---

10      This risk-topic structure can be seen as an example of a risk taxonomy for GPAI/foundation models, though the topics listed here are not collectively exhaustive and mutually exclusive. For other risk taxonomies, see, e.g., Footnote 5 in the NIST Generative AI Profile, NIST AI 600-1 (Autio et al. 2024), Section 4 of the International Scientific Report on the Safety of Advanced AI (Bengio, Privitera et al. 2024), TASRA: a Taxonomy and Analysis of Societal-Scale Risks from AI (Critch and Russell 2023), and the MIT AI risk repository (MIT 2024).

## 2.3  HIGH-PRIORITY RISK MANAGEMENT STEPS AND PROFILE GUIDANCE SECTIONS

Users of this Profile should place high priority on the following risk management steps and corresponding Profile guidance sections.[11] (Appropriately applying the Profile guidance for the following steps should be regarded as the baseline or minimum expectations for users of this Profile; users of this Profile can exceed the minimum expectations by also applying guidance in other sections.)

- **Check or update, and incorporate, each of the following high-priority risk management steps when making go/no-go decisions,** especially on whether to proceed on major stages or investments for development or deployment of cutting-edge large-scale GPAI/foundation models (Manage 1.1).

- **Take responsibility for risk assessment and risk management tasks for which your organization has access to information, capability, or opportunity to develop capability sufficient for constructive action, or that is substantially greater than that of others in the value chain** (Govern 2.1).
  - » We also recommend applying this principle throughout other risk assessment and risk management steps, and we refer to it frequently in other guidance sections.

- **Set risk-tolerance thresholds to prevent unacceptable risks** (Map 1.5).
  - » For example, the NIST AI RMF 1.0 recommends the following: "In cases where an AI system presents unacceptable negative risk levels — such as where significant negative impacts are imminent, severe harms are actually occurring, or catastrophic risks are present — development and deployment should cease in a safe manner until risks can be sufficiently managed" (NIST 2023a, p.8).

- **Identify reasonably foreseeable uses, misuses, and abuses for a GPAI/foundation model** (e.g., automated generation of toxic or illegal content or disinformation, or aiding with proliferation of cyber, chemical, biological, or radiological weapons), and identify reasonably foreseeable potential impacts (e.g., to fundamental rights) (Map 1.1).

---

11      It also can be appropriate to follow the guidance in this document for these risk management steps, but to apply and document them under other, closely related risk management steps (typically noted in this document with "see also" statements pointing to guidance in other sections of the Profile). For example, if your organization sets risk-tolerance thresholds under Govern 1.3 instead of under Map 1.5, then as part of your organization's process for Govern 1.3, it can be appropriate to follow guidance in this Profile under Map 1.5.

- **Identify whether a GPAI/foundation model could lead to significant, severe or catastrophic impacts,** e.g., because of correlated failures or errors across high-stakes deployment domains, dangerous emergent behaviors or vulnerabilities, or harmful misuses and abuses (Map 5.1).

- **Use red-teams and adversarial testing** as part of extensive interaction with GPAI/foundation models to identify dangerous capabilities, vulnerabilities, or other emergent properties of such systems (Measure 1.1).

- **Track important identified risks** (e.g., vulnerabilities from data poisoning and other attacks, or mis-specification of objectives) even if they cannot yet be measured (Measure 1.1 and Measure 3.2).

- **Implement risk-reduction controls as appropriate** throughout a GPAI/foundation models lifecycle, e.g., independent auditing, incremental scale-up, red-teaming, and other steps (Manage 1.3, Manage 2.3, and Manage 2.4).

- **Incorporate identified AI system risk factors, and circumstances that could result in impacts or harms, into reporting and engagement with internal and external stakeholders** (e.g., to downstream developers, regulators, users, impacted communities, etc.) on the AI system as appropriate, e.g., using model cards, system cards, and other transparency mechanisms (Govern 4.2).

We also recommend: **Document the process used in considering risk mitigation controls, the options considered, and reasons for choices**. Documentation on many items should be shared in publicly available material such as system cards. Some details on particular items, such as security vulnerabilities, can be responsibly omitted from public materials to reduce misuse potential, especially if available to auditors, Information Sharing and Analysis Organizations, or other parties as appropriate.

# 3. Guidance

Broadly speaking, all areas of current NIST AI RMF guidance (NIST 2023a, 2023b) seem at least partly applicable for GPAI/foundation models. However, for such AI systems, the activities and outcomes for some categories or subcategories of NIST AI RMF guidance seem higher priority than others. In the following section, we include a number of excerpts from the NIST AI RMF Playbook (NIST 2023b) that seem particularly valuable for GPAI/foundation model developers, given current typical model architectures and development practices. These are in *italic font*, and are preceded by statements of the form "In the NIST AI RMF Playbook guidance for __, particularly valuable action and documentation items for GPAI/foundation models include __". We have also included excerpts from the NIST Generative AI Profile, NIST AI 600-1 (Autio et al. 2024) where particularly valuable and applicable. These are also *in italic font* and preceded by statements of the form "In the NIST GAI Profile guidance for __, additional particularly valuable actions include __."[12] Sub-categories that are in the NIST Generative AI Profile, but do not include excerpts, have the NIST Generative AI Profile listed as a resource.

The tables in this section provide applicability of NIST AI RMF categories and subcategories, and supplemental guidance, for GPAI/foundation models. The tables address the following AI RMF functions: Table 1 for Govern, Table 2 for Map, Table 3 for Measure, and Table 4 for Manage.

## 3.1 GUIDANCE FOR NIST AI RMF GOVERN SUBCATEGORIES

### Table 1: Guidance for NIST AI RMF Govern Subcategories

| Govern Category or Subcategory | Applicability and supplemental guidance for GPAI/foundation models | Resources |
|---|---|---|
| **Govern 1:** Policies, processes, procedures, and practices across the organization related to the mapping, measuring, and managing of AI risks are in place, transparent, and implemented effectively. | | |
| **Govern 1.1:** Legal and regulatory requirements involving AI are understood, managed, and documented. | The legal and regulatory environment for GPAI/foundation models is evolving quickly and will require regular assessment for continued compliance. GPAI/foundation model developers, deployers, and users should assess the extent to which their activities would fall under GPAI/foundation model-related laws or regulations. (See, e.g., policy trackers such as OECD.AI n.d. and IAPP 2024a,b.) These can include: | NIST AI RMF Playbook (NIST 2023b) NIST Generative AI Profile, NIST AI 600-1 (Autio et al. 2024) |

---

| Govern Category or Subcategory | Applicability and supplemental guidance for GPAI/foundation models | Resources |
|---|---|---|
| Govern 1.1, continued | • The EU AI Act entered into force on August 1, 2024 and applies to any AI product or service offered in the EU regardless of where they are established. It includes Obligations for Providers of General Purpose AI Models in Article 53, as well as Obligations for Providers of General-Purpose AI Models with Systemic Risk (i.e., frontier or near-frontier models) in Article 55. Planned Codes of Practice[13] detailed in Article 56 will provide further guidance for the development of General-Purpose AI Models.<br>• US Executive Order 14110 (White House 2023c)[14] includes provisions that apply to developers of dual-use foundation models (i.e., frontier models) in the United States. Section 4.2 of the Executive Order includes reporting requirements for developers of dual-use foundation models, including for cybersecurity protections and results of red-team evaluations of dual-use model capabilities (e.g., in CBRN and cyber domains).<br><br>Many pre-existing legal and regulatory requirements, for example those related to copyright, data protection, discrimination, and privacy rights are particularly relevant to GPAI/foundation models that are trained on large swaths of the internet. International human rights law is also highly relevant to GPAI/foundation models; see, e.g., The Risk Management Profile for Artificial Intelligence and Human Rights (DOS 2024).<br><br>See the Mapping of Profile Guidance V1.1 to Key Standards and Regulations (Barrett et al. 2025a) supporting document for mappings of our Profile guidance to key standards, commitments, and regulations, including ISO/IEC 23894, the White House AI Commitments, and the EU AI Act. | For Policy Tracking:<br>OECD National AI policies & strategies Repository (OECD.AI n.d.)<br>Global AI Law and Policy Tracker (IAPP 2024a)<br>US State Governance Legislation Tracker (IAPP 2024b)<br><br>On the EU AI Act:<br>EP (2024)<br><br>On copyright and fair use:<br>Henderson et al. (2023)<br>Samuelson (2023)<br><br>On human rights obligations: Risk Management Profile for Artificial Intelligence and Human Rights (DOS 2024) |
| **Govern 1.2:** The characteristics of trustworthy AI are integrated into organizational policies, processes, procedures, and practices. | The characteristics of trustworthy AI, described in the NIST AI RMF, include: valid and reliable, safe, secure and resilient, accountable and transparent, explainable and interpretable, privacy-enhanced, and fair with harmful bias managed.<br><br>For GPAI/foundation models, there are some unique or particularly important considerations related to ensuring the characteristics of trustworthy AI are integrated into organizational policies, processes, procedures, and practices (Newman 2023; Wang et al., 2023). Some of these are mentioned below; see also the more detailed considerations and guidance throughout this document:<br><br>Valid and Reliable:<br>• E.g., improve predictability, review dependencies on external parties, assess quality of training data, and train operators of the system to exercise oversight and avoid overconfidence in the system.<br>Safe:<br>• E.g., establish reliable technical and procedural controls, re-evaluate safety regularly, assess shifts over time, and report incidents and adverse impacts.<br>Fair with Harmful Bias Managed:<br>• E.g., engage with impacted communities, test for biased or discriminatory outputs, review impacts on human rights and wellbeing, assess accessibility of user interface, and determine how to equitably distribute benefits. | NIST AI RMF Playbook (NIST 2023b)<br>NIST Generative AI Profile, NIST AI 600-1 (Autio et al. 2024)<br>Newman (2023)<br>Wang, Chen et al. (2023)<br><br>On secure software and AI development approaches:<br>CISA (2023)<br>NIST SP 800-218 (Souppaya et al. 2022)<br>NIST SP 800-218A (Booth et al. 2024) |

13     Our work on the Profile V1.1 concluded prior to the drafting and publishing of the final EU AIA Code of Practice.

14     The Profile V1.1 and its supporting documents were drafted and finalized prior to the recession of Executive Order 14110 on the Safe, Secure, and Trustworthy AI on January 20, 2025.

| Govern Category or Subcategory | Applicability and supplemental guidance for GPAI/foundation models | Resources |
|---|---|---|
| Govern 1.2, continued | Secure and Resilient:<br>• E.g., assess robustness in novel environments, establish protections against adversarial attacks, and establish a coordinated policy to encourage responsible vulnerability research and disclosure.<br>Explainable and Interpretable:<br>• E.g., ensure users know how to interpret system behavior and outputs, including limitations.<br>Privacy-enhanced:<br>• E.g., enable people to consent to the uses of their data and opt out of the uses of their data, and notify users about privacy and security breaches.<br>Accountable and Transparent:<br>• E.g., establish transparency and documentation policies and processes, determine a publication/release strategy, inform users when they are interacting with the AI system or viewing AI-generated content, allow people to opt out, support independent third-party auditing and evaluation, and provide redress to people who are negatively affected.<br>Responsible:<br>• E.g., review, assess, and potentially control for adverse impacts on global financial systems, supply chains, markets, labor markets, the environment, and natural resources. | |
| **Govern 1.3:**<br>Processes, procedures, and practices are in place to determine the needed level of risk management activities based on the organization's risk tolerance. | GPAI/foundation models can have greater impacts or pose greater risks than smaller or less capable AI systems due to their potential use in many different downstream applications. Therefore, for GPAI/foundation models, it is often appropriate to make model risk assessment and management a higher priority, and devote more resources, as compared with lower-capability and lower-impact AI systems.<br><br>In the NIST GAI Profile guidance for Govern 1.3, additional particularly valuable actions include:<br>• *Define risk tiers.*<br>• *Establish minimum thresholds as part of deployment approval.*<br>• *Establish a plan to periodically evaluate whether the model may misuse CBRN information or capabilities and/or offensive cyber capabilities.*<br>• *Obtain input from stakeholder communities on unacceptables uses.*<br>• *Maintain an updated hierarchy of identified and expected GAI risks connected to contexts of GAI model advancement and use.*<br>• *Reevaluate organizational risk tolerances to account for unacceptable negative risk.*<br>• *Devise a plan to halt development or deployment of a GAI system that poses unacceptable negative risk.*<br><br>(See also the material in this document under Map 1.1 and Map 5.1 for related guidance on GPAI/foundation model impact assessment, including on impact identification and impact magnitude rating, and material under Map 1.5 on risk tolerance, including on setting unacceptable-risk thresholds.) | NIST AI RMF Playbook (NIST 2023b)<br>NIST Generative AI Profile, NIST AI 600-1 (Autio et al. 2024)<br><br>For risk tier considerations see GV-1.3-001 in NIST AI 600-1 (Autio et al. 2024) |

| Govern Category or Subcategory | Applicability and supplemental guidance for GPAI/foundation models | Resources |
|---|---|---|
| **Govern 1.4:** The risk management process and its outcomes are established through transparent policies, procedures, and other controls based on organizational risk priorities. | In the NIST AI RMF Playbook guidance for Govern 1.4, particularly valuable action and documentation items for GPAI/foundation models include: <br>• *Establish and regularly review documentation policies that, among others, address information related to:* <br>   ○ *Expected and potential risks and impacts* <br>   ○ *Assumptions and limitations* <br>   ○ *Description and characterization of training data* <br>   ○ *Testing and validation results (including explanatory visualizations and information)* <br>   ○ *Down- and up-stream dependencies* <br>   ○ *Plans for deployment, monitoring, and change management* <br>   ○ *Stakeholder engagement plans* <br>• *Establish policies and processes regarding public disclosure of the use of AI and risk management material such as impact assessments, audits, model documentation and validation, and testing results.* <br>• *Document and review the use and efficacy of different types of transparency tools and follow industry standards at the time a model is in use.* <br><br>In the NIST GAI Profile guidance for Govern 1.4, additional particularly valuable actions include: <br>• *Establish policies and mechanisms to prevent GAI systems from generating CSAM, NCII, or content that violates the law.* <br>• *Establish transparent acceptable use policies for GAI that address illegal use or applications of GAI.* <br><br>(When considering disclosure of risk management material, such as impact assessments, audits, model documentation and validation and testing results, see also the material under Govern 4.2 for related guidance on documentation and communication.) | NIST AI RMF Playbook (NIST 2023b) <br>NIST Generative AI Profile, NIST AI 600-1 (Autio et al. 2024) <br>Bender et al. (2022) <br>Gebru et al. (2021) <br>Mitchell et al. (2019) |
| **Govern 1.5:** Ongoing monitoring and periodic review of the risk management process and its outcomes are planned and organizational roles and responsibilities clearly defined, including determining the frequency of periodic review. | **Plan to identify GPAI/foundation model impacts (including to human rights) and risks (including potential uses, misuses, and abuses), starting from an early AI lifecycle stage and repeatedly through new lifecycle phases or as new information becomes available.** This is particularly important for GPAI/foundation models, which can have large numbers of uses, risks, and impacts, including from emergent capabilities and vulnerabilities. <br>• **On GPAI/foundation model lifecycle and when to assess risks:** <br>   ○ For larger machine learning models, iterations are often slower than typical Agile sprints. For larger models, the pipeline is often to pre-train a model, analyze, customize, reanalyze, customize differently, etc., then deploy and monitor, then decommission. (Here we use "analyze" as a shorthand for probing, stress testing, red-teaming, monitoring in simulated environments, etc.) <br>   ○ On red-teaming, see e.g., Ganguli, Lovitt et al. (2022), and guidance in this document under Measure 1.1. <br>• All the relevant parties, especially researchers involved in the R&D process, should have some minimal knowledge on the risks of GPAI/foundation models or be taught about such risks upon their inclusion on an advisory team. <br>• For larger models or close-to-frontier models, "Map" activities to identify risks should also happen after model training to incorporate developer findings about a model's capabilities. <br>• **On identifying potential uses, misuses, and abuses of a GPAI/foundation model:** <br>   ○ Identify potential use cases during early stages of your AI system lifecycle, such as the plan and design stages, at minimum. | Barrett et al. (2022) <br>PAI (2023a) <br>NIST AI RMF Playbook (NIST 2023b) <br>NIST Generative AI Profile, NIST AI 600-1 (Autio et al. 2024) |

| Govern Category or Subcategory | Applicability and supplemental guidance for GPAI/foundation models | Resources |
|---|---|---|
| Govern 1.5, continued | ◦ Identify misuse or abuse cases during all major stages of your AI system lifecycle (or approximate equivalents in Agile/iterative development sprints), such as: plan, data collection, design, train/build/buy, test and evaluation, deploy, operate and monitor, and decommission.<br>◦ Revisit use and misuse case identification at key intended milestones, or at periodic intervals (e.g., at least annually), whichever comes first.<br>◦ Create a plan for ongoing use case identification and categorization to extend identified uses, misuses, and abuses, based on information gained continuously from sources such as:<br>  » Downstream user and developer exploration of the AI system; and<br>  » API misuse and abuse monitoring.<br>• **When making go/no-go decisions**, especially on whether to proceed on major stages or investments for development or deployment of frontier models, see guidance in this document under Manage 1.1.<br>  ◦ It can be valuable to revisit risk assessment at these intervals, especially prior to beginning a new frontier-model training run. At or near a foundation model frontier, it would be particularly important to obtain and integrate new information on emergent properties of frontier models before incurring the expenditures and risks of the next big training run.<br><br>In the NIST AI RMF Playbook guidance for Govern 1.5, particularly valuable action and documentation items for GPAI/foundation models include:<br>• *Establish policies to allocate appropriate resources and capacity for assessing impacts of AI systems on individuals, communities, and society.*<br>• *Establish policies and procedures for monitoring and addressing AI system performance and trustworthiness, including bias and security problems, across the lifecycle of the system.*<br>• *Establish policies for AI system incident response, or confirm that existing incident response policies apply to AI systems.*<br>• *Establish policies to define organizational functions and personnel responsible for AI system monitoring and incident response activities.*<br>• *Establish mechanisms to enable the sharing of feedback from impacted individuals or communities about negative impacts from AI systems.*<br>• *Establish mechanisms to provide recourse for impacted individuals or communities to contest problematic AI system outcomes.*<br><br>In the NIST GAI Profile guidance for Govern 1.5, additional particularly valuable actions include:<br>• *Define organizational responsibilities for periodic review of content provenance and incident monitoring for GAI systems.*<br>• *Establish organizational policies and procedures for after-action reviews of GAI system incident response and incident disclosures, to identify gaps; update incident response and incident disclosure processes as required.*<br>• *Maintain a document retention policy to keep history for test, evaluation, validation, and verification (TEVV), and digital content transparency methods for GAI.* | |

| Govern Category or Subcategory | Applicability and supplemental guidance for GPAI/foundation models | Resources |
|---|---|---|
| **Govern 1.6:** Mechanisms are in place to inventory AI systems and are resourced according to organizational risk priorities. | (No supplemental guidance, beyond the broadly applicable guidance in the NIST AI RMF Playbook.)<br><br>In the NIST GAI Profile guidance for Govern 1.6, additional particularly valuable actions include:<br>• *Enumerate organizational GAI systems for incorporation into AI system inventory and adjust AI system inventory requirements to account for GAI risks.*<br>• *In addition to general model, governance, and risk information, consider the following items in GAI system inventory entries: Data provenance information (e.g., source, signatures, versioning, watermarks); Known issues reported from internal bug tracking or external information sharing resources (e.g., AI incident database, AVID, CVE, NVD, or OECD AI incident monitor); Human oversight roles and responsibilities; Special rights and considerations for intellectual property, licensed works, or personal, privileged, proprietary or sensitive data; Underlying foundation models, versions of underlying models, and access modes.* | NIST AI RMF Playbook (NIST 2023b)<br>NIST Generative AI Profile, NIST AI 600-1 (Autio et al. 2024) |
| **Govern 1.7:** Processes and procedures are in place for decommissioning and phasing out AI systems safely and in a manner that does not increase risks or decrease the organization's trustworthiness. | GPAI/foundation model developers that publicly release the model parameter weights for their models, and model developers that suffer a leak of model weights, will in effect be unable to decommission AI systems that others build using those model weights. (See also guidance in this document under Manage 2.4, recommending structured access or staged release approaches, including for foundation model developers that plan on open weights or open-source releases of their models.)<br><br>In the NIST AI RMF Playbook guidance for Govern 1.7, particularly valuable action and documentation items for GPAI/foundation models include:<br>• *Establish policies for decommissioning AI systems. Such policies typically address:*<br>  ◦ *User and community concerns, and reputational risks.*<br>  ◦ *Business continuity and financial risks.*<br>  ◦ *Up and downstream system dependencies.*<br>  ◦ *Regulatory requirements (e.g., data retention).*<br>  ◦ *Potential future legal, regulatory, security or forensic investigations.*<br>  ◦ *Migration to the replacement system, if appropriate.*<br>• *If anyone believes that the AI no longer meets this ethical framework, who will be responsible for receiving the concern and as appropriate investigating and remediating the issue? Do they have authority to modify, limit, or stop the use of the AI?*<br><br>In the NIST GAI Profile guidance for Govern 1.7, additional particularly valuable actions include:<br>• *Consider the following factors when decommissioning GAI systems: Data retention requirements; Data security, e.g., containment, protocols, Data leakage after decommissioning; Dependencies between upstream, downstream, or other data, internet of things (IOT) or AI systems; Use of open-source data or models; Users' emotional entanglement with GAI functions.* | NIST AI RMF Playbook (NIST 2023b)<br>NIST Generative AI Profile, NIST AI 600-1 (Autio et al. 2024) |

| Govern Category or Subcategory | Applicability and supplemental guidance for GPAI/foundation models | Resources |
|---|---|---|
| **Govern 2:** Accountability structures are in place so that the appropriate teams and individuals are empowered, responsible, and trained for mapping, measuring, and managing AI risks. | | |
| **Govern 2.1:** Roles and responsibilities and lines of communication related to mapping, measuring, and managing AI risks are documented and are clear to individuals and teams throughout the organization. | **Regarding roles and responsibilities across a GPAI/foundation model value chain:**<br>• **GPAI/foundation model developers should be responsible for risk assessment and risk management tasks for which they have, or reasonably believe they might have, access to information, capability, or opportunity to develop capability sufficient for constructive action, or that is substantially greater than others in the value chain**, such as:<br>  ○ Assessing and mitigating early-stage development risks, including for AI research projects and AI systems that the organization does not plan to make available to others;<br>  ○ Testing and documentation that require direct access to training data or the AI model, such as on knowledge limits and dangerous capabilities;<br>  ○ Identifying reasonably foreseeable uses, misuses, and abuses of the AI system;<br>  ○ Implementing appropriate precautions to prevent or mitigate identified potential misuses or abuses;[15] and<br>  ○ Making necessary information available to downstream developers and deployers building on base models, and to independent auditors or others as appropriate (e.g., to enable third-party auditability):<br>    » Make as much information available on AI risk factors, incidents (including near-miss incidents), knowledge limits, etc., as reasonably possible to all audiences.[16]<br>    » Provide additional information to downstream and end-use application developers and deployers as appropriate to meet their risk management needs.<br>• **Downstream developers and deployers** of end-use applications built on GPAI/foundation models **should be responsible for risk assessment and risk management tasks for which they have, or reasonably believe they might have, access to information, capability, or opportunity to develop capability sufficient for constructive action, or that is substantially greater than others in the value chain**, such as:<br>  ○ Establishing specific context for their intended end-use application(s), and applying risk management processes appropriate for that specific context;<br>  ○ Utilizing information provided by the upstream provider of a GPAI/foundation model, and requesting additional information as needed; and<br>  ○ Reporting to the upstream provider, and considering reporting to others such as information sharing and analysis organizations (ISAOs) or regulators as appropriate, any critical GPAI/foundation model vulnerabilities, biases, incidents (including near-miss incidents), etc., that would have high impacts on other downstream developers or deployers. | Barrett et al. (2022)<br>NIST AI RMF Playbook (NIST 2023b)<br>NIST Generative AI Profile, NIST AI 600-1 (Autio et al. 2024)<br>PAI (2023c)<br>Schuett (2022)<br>Srikumar et al. (2024) |

15   See also Manage 1.3 guidance on defining and communicating to key stakeholders whether any potential use cases would be disallowed/unacceptable, as well as Manage 2.4 guidance on staged releases and structured access for frontier and near-frontier models.

16   See also guidance in this document under Govern 4.2 and Govern 4.3 on information to share, and see Section 3.4.2.1 of Barrett et al. (2022) for guidance on providing stakeholders information on reasonably foreseeable risks without providing adversaries too much information.

| Govern Category or Subcategory | Applicability and supplemental guidance for GPAI/foundation models | Resources |
|---|---|---|
| Govern 2.1, continued | Downstream developers and deployers extending GPAI/foundation models (e.g., via fine-tuning training on data curated by the downstream developer) should also consider applying guidance for upstream developers (e.g., on testing and documentation that require direct access to fine-tuning training data) for any substantial extensions of the underlying base model. Fine-tuned versions of the underlying base models often have capabilities that base models do not.<br><br>Regarding roles and responsibilities for risk governance accountability within a single GPAI/foundation model developer or deployer organization, consider implementing "Three Lines of Defense," or 3LoD (Schuett 2022):<br>• Roles can include:<br>  1. Research team as the first line, ultimately the Head of Research or equivalent;<br>  2. Risk management team as the second line, ultimately chief risk officer (CRO) or equivalent; this can also include the legal and compliance team, technical safety team, and security team; and<br>  3. Internal audit as third line, ultimately chief audit executive (CAE); this can also include the ethics board.<br>• Reporting responsibilities can include:<br>  1. First line reports to CEO;<br>  2. Second line reports to CEO; and CRO reports to the board risk committee; and<br>  3. Third line reports to the board of directors or the board audit committee; the CAE is often part of the board audit committee.<br><br>In the NIST AI RMF Playbook guidance for Govern 2.1, particularly valuable action and documentation items for GPAI/foundation models include:<br>• *Establish policies that define the AI risk management roles and responsibilities for positions directly and indirectly related to AI systems, including, but not limited to - Boards of directors or advisory committees - Senior management - AI audit functions - Product management - Project management - AI design - AI development - Human-AI interaction - AI testing and evaluation - AI acquisition and procurement - Impact assessment functions - Oversight functions.*<br>• *Establish policies that promote regular communication among AI actors participating in AI risk management efforts.*<br>• *Establish policies that separate management of AI system development functions from AI system testing functions, to enable independent course-correction of AI systems.*<br><br>In the NIST GAI Profile guidance for Govern 2.1, additional particularly valuable actions include:<br>• *When systems may raise national security risks, involve national security professionals in mapping, measuring, and managing those risks.*<br>• *Create mechanisms to provide protections for whistleblowers who report, based on reasonable belief, when the organization violates relevant laws or poses a specific and empirically well-substantiated negative risk to public safety (or has already caused harm).* | Barrett et al. (2022)<br>NIST AI RMF Playbook (NIST 2023b)<br>NIST Generative AI Profile, NIST AI 600-1 (Autio et al. 2024)<br>PAI (2023c)<br>Schuett (2022)<br>Srikumar et al. (2024) |

| Govern Category or Subcategory | Applicability and supplemental guidance for GPAI/foundation models | Resources |
|---|---|---|
| **Govern 2.2:** The organization's personnel and partners receive AI risk management training to enable them to perform their duties and responsibilities consistent with related policies, procedures, and agreements. | In the NIST AI RMF Playbook guidance for Govern 2.2, particularly valuable action and documentation items for GPAI/foundation models include: <br> • *Ensure that trainings comprehensively address technical and socio-technical aspects of AI risk management.* <br> • *Define paths along internal and external chains of accountability to escalate risk concerns.* | NIST AI RMF Playbook (NIST 2023b) |
| **Govern 2.3:** Executive leadership of the organization takes responsibility for decisions about risks associated with AI system development and deployment. | In the NIST AI RMF Playbook guidance for Govern 2.3, particularly valuable action and documentation items for GPAI/foundation models include: <br> • *Organizational management can:* <br>    ○ *Declare risk tolerances for developing or using AI systems.* <br>    ○ *Support AI risk management efforts, and play an active role in such efforts.* <br>    ○ *Integrate a risk and harm prevention mindset throughout the AI lifecycle as part of organizational culture.* <br><br> (See also guidance under Govern 1.5 on prioritizing resources for GPAI/foundation model risk assessment and management, and under Map 1.5 on setting unacceptable-risk thresholds to prevent risks with substantial probability of inadequately mitigated catastrophic outcomes.) | NIST AI RMF Playbook (NIST 2023b) |
| **Govern 3: Workforce diversity, equity, inclusion, and accessibility processes are prioritized in the mapping, measuring, and managing of AI risks throughout the lifecycle.** | | |
| **Govern 3.1:** Decision-making related to mapping, measuring, and managing AI risks throughout the lifecycle is informed by a diverse team (e.g., diversity of demographics, disciplines, experience, expertise, and backgrounds). | Identifying the vast array of GPAI/foundation model risks and potential impacts, including via potential uses and misuses, should be performed by a demographically and disciplinarily diverse team including internal and external personnel. <br><br> Potential uses and misuses of GPAI/foundation models should be identified from an early stage in their lifecycle, because of their large numbers of potential uses and misuse. (See also related guidance in this document under Govern 1.5.) <br><br> **For staffing to identify potential uses, misuses, and abuses of a GPAI/foundation models**: <br> • Include members of each of the following functional teams (or equivalents) as appropriate: <br>    ○ Product development, operations, security, human-computer interaction, user experience, marketing and sales, legal, policy, and ethics professionals. <br> • Include members of other teams as appropriate, such as: <br>    ○ Research and development (for additional technically-informed perspectives on AI system capabilities and limitations). <br>    ○ External-facing teams and/or external stakeholders including: <br>      » Communities that might be impacted (for additional early identification of potential stakeholder concerns and other stakeholder perspectives); <br>      » Communities providing labor to develop or test models (such as manual data labeling, or providing human-feedback data), particularly when there is reason to believe these individuals could be exposed to psychologically or otherwise harmful content in the process; and <br>      » External red-teamers or auditors (for additional early-stage expertise on potential misuses). | Barrett et al. (2022) NIST AI RMF Playbook (NIST 2023b) |

| Govern Category or Subcategory | Applicability and supplemental guidance for GPAI/foundation models | Resources |
|---|---|---|
| Govern 3.1, continued | • As part of staffing to identify potential high-impact scenarios for GPAI/foundation models, broaden the team as appropriate to include social scientists and historians who can provide additional perspective on structural or systemic risks that could emerge from interactions between an AI system and other societal-level systems (Zwetsloot and Dafoe 2019). | |
| **Govern 3.2:** Policies and procedures are in place to define and differentiate roles and responsibilities for human-AI configurations and oversight of AI systems. | In the NIST GAI Profile guidance for Govern 3.2, particularly valuable actions include: <br>• *Consider adjustment of organizational roles and components across lifecycle stages of large or complex GAI systems, including: Test and evaluation, validation, and red-teaming of GAI systems; GAI content moderation; GAI system development and engineering; Increased accessibility of GAI tools, interfaces, and systems, Incident response and containment.* <br>• *Define acceptable use policies for GAI interfaces, modalities, and human-AI configu-rations (i.e., for chatbots and decision-making tasks), including criteria for the kinds of queries GAI applications should refuse to respond to.* <br>• *Establish policies for user feedback mechanisms for GAI systems which include thor-ough instructions and any mechanisms for recourse.* <br>• *Engage in threat modeling to anticipate potential risks from GAI systems.* <br><br>(See also guidance in this document for Govern 2.1, regarding roles within an organization, and for upstream developers as well as downstream developers and deployers.) | NIST AI RMF Playbook (NIST 2023b) <br>NIST Generative AI Profile, NIST AI 600-1 (Autio et al. 2024) |
| **Govern 4: Organizational teams are committed to a culture that considers and communicates AI risk.** | | |
| **Govern 4.1:** Organizational policies and practices are in place to foster a critical thinking and safety-first mindset in the design, development, deployment, and uses of AI systems to minimize potential negative impacts. | In the NIST AI RMF Playbook guidance for Govern 4.1, particularly valuable action and documentation items for GPAI/foundation models include: <br>• *Establish policies that require inclusion of oversight functions (legal, compliance, risk management) from the outset of the system design process.* <br>• *Establish policies that promote effective challenge of AI system design, implementa-tion, and deployment decisions, via mechanisms such as the three lines of defense, model audits, or red-teaming — to minimize workplace risks such as groupthink.* <br>• *Establish policies that incentivize safety-first mindset and general critical thinking and review at an organizational and procedural level.* <br>• *Establish whistleblower protections for insiders who report on perceived serious problems with AI systems.* <br>• *Establish policies to integrate a harm and risk prevention mindset throughout the AI lifecycle.* <br>• *To what extent has the entity documented the AI system's development, testing methodology, metrics, and performance outcomes?* <br>• *Are organizational information sharing practices widely followed and transparent, such that related past failed designs can be avoided?* <br>• *Are processes for operator reporting of incidents and near-misses documented and available?* <br><br>In the NIST GAI Profile guidance for Govern 4.1, additional particularly valuable actions include: <br>• *Establish policies and procedures that address continual improvement processes for GAI risk measurement. Address general risks associated with a lack of explainability and transparency in GAI systems by using ample documentation and techniques such as: application of gradient-based attributions, occlusion/term reduction, counter-factual prompts and prompt engineering, and analysis of embeddings; Assess and update risk measurement approaches at regular cadences.* | Barrett et al. (2022) <br>Ganguli, Lovitt et al. (2022) <br>NIST AI RMF Playbook (NIST 2023b) <br>NIST Generative AI Profile, NIST AI 600-1 (Autio et al. 2024) <br>Schuett (2022) |

| Govern Category or Subcategory | Applicability and supplemental guidance for GPAI/foundation models | Resources |
|---|---|---|
| **Govern 4.1, continued** | (See also guidance under Govern 1.5 on when to assess potential impacts in a GPAI/ foundation model lifecycle and on red-teaming, and guidance under Govern 2.1 on "Three Lines of Defense" roles and responsibilities within a model developer or deployer organization.) | |
| **Govern 4.2:** Organizational teams document the risks and potential impacts of the AI technology they design, develop, deploy, evaluate, and use, and they communicate about the impacts more broadly. | GPAI/foundation model developers should identify, assess, and document reasonably foreseeable or currently present GPAI/foundation model impacts and risks, and communicate those as appropriate to relevant stakeholders, such as downstream developers and potentially impacted communities. These activities are particularly important for GPAI/foundation models given the relatively large scale of potential impact that often can be expected with GPAI/foundation models.<br><br>Additional guidance under Govern 4.2:<br>**Incorporate identified AI system risk factors, and circumstances that could result in impacts or harms, into reporting and engagement with internal and external stakeholders** (e.g., to downstream developers, regulators, etc.) on the AI system as appropriate (e.g., using model cards, datasheets, reward reports, factsheets, transparency notes, or system cards).[17] **Report** (as appropriate) **identified AI system risk factors, and circumstances that could result in impacts or harms**:[18]<br>• To the organization;<br>• To other organizations;<br>• To individuals, including impacts to health, safety, well-being, or fundamental rights;<br>• To groups, including populations vulnerable to disproportionate adverse impacts or harms; and<br>• To society, including:<br>  ◦ Damage to or incapacitation of a critical infrastructure sector;<br>  ◦ Economic and national security;<br>  ◦ Impacts on democratic institutions and quality of life; and<br>  ◦ Environmental impacts<br>  ◦ **Additional identified factors that could lead to severe or catastrophic consequences for society**, such as:[19,20]<br>    » Potential for correlated robustness failures or other systemic risks across high-stakes application domains such as critical infrastructure or essential services; | Sections 3.2 and 3.3 of Barrett et al. (2022) PAI (2022) PAI (2023a) NIST AI RMF Playbook (NIST 2023b) NIST Generative AI Profile, NIST AI 600-1 (Autio et al. 2024)<br><br>On model cards, system cards and related transparency tools: Mitchell et al. (2019) Gebru et al. (2021) Gilbert, Dean et al. (2022) Gilbert, Lambert et al. (2022) Microsoft (2022a) Hind (2020) Green et al. (2022) OECD (2022a) |

17    Model cards (Mitchell et al. 2019) include a model's primary intended use, out-of-scope uses, and ethics issues (which can include risks and mitigations). Datasheets for datasets (Gebru et al. 2021) include datasets' recommended uses (as well as potential risks and mitigation). Reward reports (Gilbert, Dean et al. 2022, Gilbert, Lambert et al. 2022) include objectives specification information (e.g., optimization goals and failure modes), and implementation limitations. Related industry approaches include Microsoft's Transparency Notes (see examples at Microsoft 2022a), IBM's FactSheets (Hind 2020) and Meta/Facebook's System Cards (Green et al. 2022). The OECD framework for AI system classification includes information on AI system contexts, data and input, AI model, and task and output (OECD 2022a).

18    See guidance in this document under Map 1.1 for more on such factors.

19    See guidance in this document under Map 5.1 for more on such factors.

20    Documentation on many items should be shared in publicly available material such as system cards. Some details on particular items such as security vulnerabilities can be responsibly omitted from public materials to reduce misuse potential, especially if available to auditors, Information Sharing and Analysis Organizations, or other parties as appropriate. For more on what details to omit from publicly available material, see, e.g., PAI (2022).

| Govern Category or Subcategory | Applicability and supplemental guidance for GPAI/foundation models | Resources |
|---|---|---|
| Govern 4.2, continued | » Potential for other systemic risks, which can be accumulated, accrued, correlated or compounded at societal scale, e.g.:<br>– Potential for correlated bias across a large fraction of a society's population; and<br>– Potential for many high-impact uses or misuses beyond an originally intended use case. (GPAI/foundation models typically have many reasonably foreseeable uses).<br>» Potential for large harms from mis-specified or mis-generalized goals; and<br>» Other identified factors affecting risks of high consequence / catastrophic and novel or "Black Swan" events.<br><br>In the NIST AI RMF Playbook guidance for Govern 4.2, particularly valuable action and documentation items for GPAI/foundation models include:<br>• *Establish impact assessment policies and processes for AI systems used by the organization.*<br>• *Align organizational impact assessment activities with relevant regulatory or legal requirements.*<br>• *Verify that impact assessment activities are appropriate to evaluate the potential negative impact of a system and how quickly a system changes, and that assessments are applied on a regular basis.*<br>• *Utilize impact assessments to inform broader evaluations of AI system risk.*<br>• *How has the entity identified and mitigated potential impacts of bias in the data, including inequitable or discriminatory outcomes?*<br>• *To what extent has the entity documented and communicated the AI system's development, testing methodology, metrics, and performance outcomes?*<br><br>In the NIST GAI Profile, additional particularly valuable guidance for Govern 4.2 includes:<br>• *Establish terms of use and terms of service for GAI systems.*<br>• *Verify that downstream GAI impacts, such as plugins, are included in the impact documentation process.*<br><br>(See also guidance in this document under Map 1.1 and Map 5.1 on GPAI/foundation model impact identification and impact magnitude assessment, including on consideration of factors that could lead to significant, severe, or catastrophic harms, and under Manage 1.3 on transparency and disclosure of generative AI outputs.) | |
| Govern 4.3:<br>Organizational practices are in place to enable AI testing, identification of incidents, and information sharing. | Guidance under Govern 4.3:<br>• If the organization will need to characterize an AI system according to an AI classification framework (such as in the OECD framework or frameworks for model cards, datasheets, reward reports, factsheets, transparency notes, or system cards), use risk assessment outputs as part of preparation for AI classification reporting. (Or if the AI system is already classified with another framework, use the AI classification information to inform risk assessment.)<br>   ◎ Consider classifying or otherwise characterizing each reasonably foreseeable use case or type of use case for a GPAI/foundation model, as in the guidance in this document under Map 1.1 and Map 2.1.<br>• Consider widely sharing information on relevant incidents, including on near-miss incidents, via public AI Incident Databases (AIID n.d., MITRE n.d.b).<br>• Consider membership and participation in organizations, such as the Frontier Model Forum (2024) or the US AI Safety Institute Consortium, that facilitate information sharing. | AIID (n.d.)<br>ATLAS AI Incidents (MITRE n.d.b)<br>Section 3.4 of Barrett et al. (2022)<br>NIST AI RMF Playbook (NIST 2023b)<br>NIST Generative AI Profile, NIST AI 600-1 (Autio et al. 2024)<br>Frontier Model Forum (2024) |

| Govern Category or Subcategory | Applicability and supplemental guidance for GPAI/foundation models | Resources |
|---|---|---|
| Govern 4.3, continued | In the NIST AI RMF Playbook guidance for Govern 4.3, particularly valuable action and documentation items for GPAI/foundation models include:<br>• *Establish policies and procedures to facilitate and equip AI system testing.*<br>• *Establish organizational commitment to identifying AI system limitations and sharing of insights about limitations within appropriate AI actor groups.*<br>• *Establish policies for reporting and documenting incident response.*<br>• *Establish policies and processes regarding public disclosure of incidents and information sharing.*<br>• *Establish guidelines for incident handling related to AI system risks and performance.*<br>• *To what extent can users or parties affected by the outputs of the AI system test the AI system and provide feedback?*<br><br>(See also guidance in this document under Govern 2.1 regarding risk-assessment and information-sharing roles for upstream developers as well as downstream developers and deployers.) | On incident disclosure plans: Turri and Dzombak (2023) |
| **Govern 5:  Processes are in place for robust engagement with relevant AI actors.** | | |
| **Govern 5.1:** Organizational policies and practices are in place to collect, consider, prioritize, and integrate feedback from those external to the team that developed or deployed the AI system regarding the potential individual and societal impacts related to AI risks. | GPAI/foundation model developers and deployers should integrate feedback from those external to their team. Types of external feedback that should be utilized where appropriate include:<br>• Deliberation with impacted communities, including people involved with the human labor and training of GPAI/foundation models (such as data annotators and content reviewers), people whose work is "scraped" for training purposes (such as artists and authors), intended users, and people whose livelihoods are altered by the use of the system;<br>• Independent auditing throughout the AI lifecycle;<br>• Bug bounty and bias bounty programs;<br>• Red-teaming; and<br>• Feedback channels with users or impacted individuals or communities, including appeal and redress mechanisms.<br>In the NIST AI RMF Playbook guidance for Govern 5.1, particularly valuable action and documentation items for GPAI/foundation models include:<br>• *Establish AI risk management policies that explicitly address mechanisms for collecting, evaluating, and incorporating stakeholder and user feedback that could include:*<br>  ◦ *Recourse mechanisms for faulty AI system outputs.*<br>  ◦ *Bug bounties.*<br>  ◦ *Human-centered design.*<br>  ◦ *User-interaction and experience research.*<br>  ◦ *Participatory stakeholder engagement with individuals and communities that may experience negative impacts.*<br>• *What type of information is accessible on the design, operations, and limitations of the AI system to external stakeholders, including end users, consumers, regulators, and individuals impacted by use of the AI system?*<br>• *What was done to mitigate or reduce the potential for harm?*<br>• *Stakeholder involvement: Include diverse perspectives from a community of stakeholders throughout the AI life cycle to mitigate risks.*<br><br>In the NIST GAI Profile, additional particularly valuable guidance for Govern 5.1 includes:<br>• *Document interaction with GPAI or GAI systems to users prior to any activities, particularly in contexts involving more significant risks.* | NIST AI RMF Playbook (NIST 2023b) NIST Generative AI Profile, NIST AI 600-1 (Autio et al. 2024) NIST AI 800-1 ipd (NIST 2024b, Objective 6)<br><br>On bug bounties: Kenway et al. (2022) |

| Govern Category or Subcategory | Applicability and supplemental guidance for GPAI/foundation models | Resources |
|---|---|---|
| **Govern 5.1, continued** | (See also guidance in this document under Measure 1.1 and Measure 1.3 for more detailed recommendations about using red-teams and independent red-teaming organizations that are separate enough from direct development operations of a GPAI/ foundation model that they can provide relatively unbiased assessments of that model, and guidance in this document under Measure 3.2 on bug bounties and bias bounties.) | |
| **Govern 5.2:** Mechanisms are established to enable the team that developed or deployed AI systems to regularly incorporate adjudicated feedback from relevant AI actors into system design and implementation. | In the NIST AI RMF Playbook guidance for Govern 5.2, particularly valuable action and documentation items for GPAI/foundation models include:<br>• *Explicitly acknowledge that AI systems, and the use of AI, present inherent costs and risks along with potential benefits.*<br>• *Define reasonable risk tolerances for AI systems informed by laws, regulation, best practices, or industry standards.*<br>• *Establish policies that ensure all relevant AI actors are provided with meaningful opportunities to provide feedback on system design and implementation.*<br>• *Establish policies that define how to assign AI systems to established risk tolerance levels by combining system impact assessments with the likelihood that an impact occurs. Such assessment often entails some combination of:*<br>  ◦ *Econometric evaluations of impacts and impact likelihoods to assess AI system risk.*<br>  ◦ *Red-amber-green (RAG) scales for impact severity and likelihood to assess AI system risk.*<br>  ◦ *Establishment of policies for allocating risk management resources along estab-lished risk tolerance levels, with higher-risk systems receiving more risk manage-ment resources and oversight.*<br>  ◦ *Establishment of policies for approval, conditional approval, and disapproval of the design, implementation, and deployment of AI systems.*<br>• *Establish policies facilitating the early decommissioning of AI systems that surpass an organization's ability to reasonably mitigate risks.*<br>• *Who is accountable for the ethical considerations during all stages of the AI lifecycle?*<br><br>(See also guidance in this document under Govern 2.1 on the roles for GPAI/foundation model upstream developers as well as downstream developers and deployers. See also guidance in this document under Map 1.5 on setting risk tolerance thresholds, including on setting unacceptable-risk thresholds to prevent risks with substantial probability of inadequately-mitigated catastrophic outcomes.) | NIST AI RMF Playbook (NIST 2023b)<br>NIST AI 800-1 ipd (NIST 2024b, Objective 6) |
| **Govern 6:  Policies and procedures are in place to address AI risks and benefits arising from third-party software and data and other supply chain issues.** | | |
| **Govern 6.1:** Policies and procedures are in place that address AI risks associated with third-party entities, in-cluding risks of infringe-ment of a third-party's intellectual property or other rights. | In the NIST AI RMF Playbook guidance for Govern 6.1, particularly valuable action and documentation items for GPAI/foundation models include:<br>• *Establish policies related to:*<br>  ◦ *Transparency into third-party system functions, including knowledge about train-ing data, training and inference algorithms, and assumptions and limitations.*<br>  ◦ *Thorough testing of third-party AI systems. (See MEASURE for more detail)*<br>  ◦ *Requirements for clear and complete instructions for third-party system usage.*<br>• *Did you establish mechanisms that facilitate the AI system's auditability (e.g. trace-ability of the development process, the sourcing of training data and the logging of the AI system's processes, outcomes, positive and negative impact)?*<br>• *Did you ensure that the AI system can be audited by independent third parties?*<br>• *Did you establish a process for third parties (e.g. suppliers, end users, subjects, distributors/vendors or workers) to report potential vulnerabilities, risks or biases in the AI system?* | Barrett et al. (2022)<br>NIST AI RMF Playbook (NIST 2023b)<br>NIST Generative AI Profile, NIST AI 600-1 (Autio et al. 2024) |

| Govern Category or Subcategory | Applicability and supplemental guidance for GPAI/foundation models | Resources |
|---|---|---|
| Govern 6.1, continued | In the NIST GAI Profile, additional particularly valuable guidance for Govern 6.1 includes:<br>• *Draft and maintain well-defined contracts and service level agreements (SLAs) that specify content ownership, usage rights, quality standards, security requirements, and content provenance expectations for GAI systems.*<br>• *Include clauses in contracts which allow an organization to evaluate third-party GAI processes and standards.*<br>• *Update and integrate due diligence processes for GAI acquisition and procurement vendor assessments to include intellectual property, data privacy, security, and other risks. For example, update processes to: Address solutions that may rely on embedded GAI technologies; Address ongoing monitoring, assessments, and alerting, dynamic risk assessments, and real-time reporting tools for monitoring third-party GAI risks; Consider policy adjustments across GAI modeling libraries, tools and APIs, fine-tuned models, and embedded tools; Assess GAI vendors, open-source or propri-etary GAI tools, or GAI service providers against incident or vulnerability databases.*<br><br>(See also guidance in this document under Govern 2.1 on the roles for GPAI/foundation model upstream developers, e.g., on making necessary information available to downstream developers, independent auditors, or others as appropriate, as well as roles for downstream developers and deployers.) | |
| **Govern 6.2:** Contingency processes are in place to handle failures or incidents in third-party data or AI systems deemed to be high-risk. | In the NIST AI RMF Playbook guidance for Govern 6.2, particularly valuable action and documentation items for GPAI/foundation models include:<br>• *Establish policies for handling third-party system failures to include consideration of redundancy mechanisms for vital third-party AI systems.*<br>• *Verify that incident response plans address third-party AI systems.*<br>• *To what extent does the plan specifically address risks associated with acquisition, procurement of packaged software from vendors, cybersecurity controls, computa-tional infrastructure, data, data science, deployment mechanics, and system failure?*<br>• *Did you establish a process for third parties (e.g. suppliers, end users, subjects, distributors/vendors or workers) to report potential vulnerabilities, risks or biases in the AI system?*<br><br>In the NIST GAI Profile, additional particularly valuable guidance for Govern 6.2 includes:<br>• *Document GAI risks associated with system value chain to identify over-reliance on third-party data and to identify fallbacks.*<br>• *Document incidents involving third-party GAI data and systems, including open- data and open-source software.*<br>• *Establish policies and procedures for continuous monitoring of third-party GAI systems in deployment.*<br><br>(See also guidance in this document for Govern 2.1 on the roles for GPAI/foundation model upstream developers as well as downstream developers and deployers. See also contingency processes outlined in this document under Manage 1.3, Manage 2.4, or other Manage subcategories.) | Barrett et al. (2022) NIST AI RMF Playbook (NIST 2023b) NIST Generative AI Profile, NIST AI 600-1 (Autio et al. 2024) |

## 3.2 GUIDANCE FOR NIST AI RMF MAP SUBCATEGORIES

### Table 2: Guidance for NIST AI RMF Map Subcategories

| Map Category or Subcategory | Applicability and supplemental guidance for GPAI/foundation models | Resources |
|---|---|---|
| **Map 1:  Context is established and understood.** | | |
| **Map 1.1:** Intended purposes, potentially beneficial uses, context-specific laws, norms and expectations, and prospective settings in which the AI system will be deployed are understood and documented. Considerations include: the specific set or types of users along with their expectations; potential positive and negative impacts of system uses to individuals, communities, organizations, society, and the planet; assumptions and related limitations about AI system purposes, uses, and risks across the development or product AI lifecycle; and related TEVV and system metrics. | Developers of GPAI/foundation models should identify their reasonably foreseeable uses, misuses and abuses beyond any originally intended purposes (or in the absence of a specific intended purpose). <br><br> • **Identify reasonably foreseeable uses, misuses, or abuses for a GPAI/foundation model, beyond any originally intended use cases (or in the absence of a specific intended purpose)**. <br>  ○ Categories of reasonably foreseeable potential misuses or abuses of LLMs or other GPAI/foundation models can include: <br>   » Automated generation of disinformation, or of phishing-attack material (OpenAI 2019a, Solaiman et al. 2019, Bai, Voelkel et al. 2023, OpenAI 2023a, pp. 13–14, Barrett, Boyd et al. 2023, pp. 3–4, Park et al. 2023, Bengio, Privitera et al. 2024). <br>   » Aiding with proliferation of chemical, biological, or radiological weapons, or other weapons of mass destruction (Boiko et al. 2023, OpenAI 2023a, pp. 12–13). This can include aiding in lab experiment design and troubleshooting, providing instructions on chemical and biological material acquisition, and the capability to "upskill" threat actors by advising on attack techniques (DSIT 2023, Soice et al. 2023) <br>   » Discovery and exploitation of software vulnerabilities (OpenAI 2023a, pp. 13–14, Barrett, Boyd et al. 2023, p. 4, Chauvin 2024), cyber attack plan critiquing and assistance, malware and virus creation, including viruses that evolve over time to evade detection (DSIT 2023, Charan et al. 2023, Shimony et al. 2023). <br>   » Creation of violent, illegal, discriminatory, or harmful content, including non-consensual intimate imagery (NCII) or child sexual abuse material (CSAM) (Solaiman et al. 2023). <br> • For ML systems trained (or to be trained) on datasets, identify the goals and limitations of the data collection and curation processes, and the implications for the resulting ML systems. This is especially important for LLMs or other ML systems trained on datasets that are too large for others to inspect thoroughly, or are otherwise inaccessible to others (Bender et al. 2022). <br>  ○ Consider running data audits (Birhane et al. 2021, Dodge et al. 2021) as a part of the data management process. <br><br> **Identify reasonably foreseeable potential impacts of GPAI/foundation models, which can include but are not limited to:**[21] <br> • Impacts to organizational operations, including: <br>  ○ Missions and functions <br>   » Partial loss of understanding or control over particular functions <br>  ○ Image and reputation, including: <br>   » Loss of trust or reluctance to use the system or service <br>   » Internal culture costs that impact morale or productivity | Section 3.1.2.1 of Barrett et al. (2022) <br> Bender et al. (2022) <br> Boiko et al. (2023) <br> Eloundou et al. (2023) <br> Khlaaf et al. (2022) <br> NIST AI RMF Playbook (NIST 2023b) <br> NIST Generative AI Profile, NIST AI 600-1 (Autio et al. 2024) <br> NIST AI 800-1 ipd (NIST 2024b) <br> NIST AI 100-2e2023 (Vassilev et al. 2024) <br> OpenAI (2019b) <br> PAI (2023a) <br> Solaiman et al. (2019) <br><br> For impact assessment, see: <br> UNESCO (2023) <br><br> For human rights impact assessment, see: <br> DOS (2024) <br><br> For GPAI risk sources and risk management measures, see: <br> Gipiškis et al. (2024) <br><br> For data audits, see: <br> Birhane et al. (2021) <br> Dodge et al. (2021) <br><br> For cyber vulnerability capability evaluation see: <br> eyeballvul (Chauvin 2024) |

---

21    In-depth assessment would be most appropriate for developers of large-scale GPAI/foundation models to take a wide view of reasonably foreseeable impacts of such GPAI/foundation models, or for downstream developers focused on reasonably foreseeable impacts for a particular use case or application context. For more, see Section 3.2.2.1.1 of Barrett et al. (2022), from which we adapt this list of factors.

| Map Category or Subcategory | Applicability and supplemental guidance for GPAI/foundation models | Resources |
|---|---|---|
| Map 1.1, continued | <ul><li>Impacts to organizational assets, including legal compliance costs arising from problems created for individuals;</li><li>Impacts to other organizations;</li><li>Impacts to individuals, including impacts to health, safety, well-being, or fundamental rights.<ul><li>For identifying potential or actual human rights impacts, potential example questions and Universal Declaration of Human Rights (UDHR) Articles to consider include:[22]<ul><li>» UDHR Article 2, including non-discrimination and equality before the law.<ul><li>– How could an AI system's bias in data or unfair algorithmic decisions affect rights to equal protection and non-discrimination?</li></ul></li><li>» UDHR Article 3, including right to life and personal security.<ul><li>– How could an AI system's algorithmic decisions affect the right to life and personal security?</li></ul></li><li>» UDHR Article 12, including privacy and protection against unlawful governmental surveillance.<ul><li>– How could an AI system be used for surveillance, leading to loss of privacy or inadequate protection of personally identifiable information?</li></ul></li><li>» UDHR Articles 18 and 19, including freedom of thought, conscience, and religious belief and practice; freedom of expression; and freedom to hold opinions without interference.<ul><li>– How could an AI system affect rights to express opinions or practice religion?</li></ul></li><li>» UDHR Articles 20 and 21, including freedom of association and the right to peaceful assembly.<ul><li>– How could an AI system affect rights to association, peaceful assembly, and democratic participation in government?</li></ul></li><li>» UDHR Articles 23 and 25, including rights to decent work and to an adequate standard of living.<ul><li>– How could an AI system affect rights to decent work, including effects on adequate standard of living via displacement of human workers?</li></ul></li></ul></li></ul></li><li>Impacts to groups, including populations vulnerable to disproportionate adverse impacts or harms, such as:<ul><li>Disparate performance for different gender, race, ability, age, religion, and other demographic groups; and</li><li>Bias, stereotypes, and representational harm.</li></ul></li><li>Impacts to society, including:<ul><li>Damage to or incapacitation of a critical infrastructure sector;</li><li>Economic and national security;</li><li>Concentration and control of the power and benefits from AI technologies;</li><li>Dramatic shifts to the labor market and economic opportunities, including technological job displacement;</li><li>Impacts on democratic institutions and quality of life;</li><li>Polarization and extremism;</li></ul></li></ul> | |

22    For more guidance and resources on assessing and mitigating AI system impacts to human rights, see the Risk Management Profile for Artificial Intelligence and Human Rights from the US Department of State (DOS 2024). See also see Section 3.3 of Barrett et al. (2022), which is based heavily on the UDHR (UN 1948) and the UN Guiding Principles on Business and Human Rights (UN 2011), and other related guidance, such as the Hiroshima Process International Code of Conduct for Advanced AI Systems (G7 2023).

| Map Category or Subcategory | Applicability and supplemental guidance for GPAI/foundation models | Resources |
|---|---|---|
| Map 1.1, continued | ○ Environmental impacts, including carbon emissions and use of natural resources; and<br>○ Additional factors that could lead to severe or catastrophic consequences for society.<br><br>In the NIST GAI Profile, additional particularly valuable guidance for Map 1.1 includes:<br>• *Document risk measurement plans to address identified risks. Plans may include, as applicable: Individual and group cognitive biases (e.g., confirmation bias, funding bias, groupthink) for AI Actors involved in the design, implementation, and use of GAI systems; Known past GAI system incidents and failure modes; In-context use and foreseeable misuse, abuse, and off-label use; Over reliance on quantitative metrics and methodologies without sufficient awareness of their limitations in the context(s) of use; Standard measurement and structured human feedback approaches; Anticipated human-AI configurations.*<br><br>**(See also guidance in this document under Map 5.1 on GPAI/foundation model impact identification and impact magnitude assessment, including on consideration of factors that could lead to significant, severe, or catastrophic harms.)** | |
| **Map 1.2:** Interdisciplinary AI actors, competencies, skills, and capacities for establishing context reflect demographic diversity and broad domain and user experience expertise, and their participation is documented. Opportunities for interdisciplinary collaboration are prioritized. | In the NIST AI RMF Playbook guidance for Map 1.2, particularly valuable action and documentation items for GPAI/foundation models include:<br>• *Establish interdisciplinary teams to reflect a wide range of skills, competencies, and capabilities for AI efforts. Verify that team membership includes demographic diversity, broad domain expertise, and lived experiences. Document team composition.*<br>• *Create and empower interdisciplinary expert teams to capture, learn, and engage the interdependencies of deployed AI systems and related terminologies and concepts from disciplines outside of AI practice such as law, sociology, psychology, anthropology, public policy, systems design, and engineering.*<br><br>In the NIST GAI Profile guidance for Map 1.2, additional particularly valuable action and documentation items for GAI include:<br>• *Verify that data or benchmarks used in risk measurement, and users, participants, or subjects involved in structured public feedback exercises, are representative of diverse in-context user populations.*<br><br>(See also guidance in this document under Govern 3.1 on disciplines and functional teams to include in identifying GPAI/foundation model potential impacts and risks, including via potential uses and misuses.) | NIST AI RMF Playbook (NIST 2023b)<br>NIST Generative AI Profile, NIST AI 600-1 (Autio et al. 2024) |

| Map Category or Subcategory | Applicability and supplemental guidance for GPAI/foundation models | Resources |
|---|---|---|
| **Map 1.3:**<br>The organization's mission and relevant goals for AI technology are understood and documented. | **When formulating objectives for development of GPAI/foundation models**, in addition to broadly applicable AI development principles such as the OECD AI Principles (OECD 2019), GPAI/foundation model developers should:<br>• **Consider the potential for mis-specified AI system objectives, e.g., using over simplified or short-term metrics as proxies for desired longer-term outcomes.**<br>  ○ **For example, consider questions such as the following for an AI system: "What objective has been specified for the system, and what kinds of perverse behavior could be incentivized by optimizing for that objective?"** (Rudner and Toner, 2021, p. 10). Examples of AI systems with mis-specified objectives include machine-learning algorithms for social media content recommendation that learn to optimize user-engagement metrics by serving users with extremist content or disinformation (Rudner and Toner 2021).<br>• Consider principles relevant to foundation models and advanced AI, such as those outlined in the G7 Hiroshima Process International Guiding Principles for Organizations Developing Advanced AI System (G7 2023) and in the Asilomar AI Principles (FLI 2017).<br>Examples from the Hiroshima Principles:<br>  ○ Take appropriate measures throughout the development of advanced AI systems, including prior to and throughout their deployment and placement on the market, to identify, evaluate, and mitigate risks across the AI lifecycle.<br>  ○ Identify and mitigate vulnerabilities, and, where appropriate, incidents and patterns of misuse, after deployment including placement on the market.<br>  ○ Publicly report advanced AI systems' capabilities, limitations and domains of appropriate and inappropriate use, to support ensuring sufficient transparency, thereby contributing to increased accountability.<br>  ○ Work towards responsible information sharing and reporting of incidents among organizations developing advanced AI systems including with industry, governments, civil society, and academia.<br>  ○ Invest in and implement robust security controls, including physical security, cybersecurity and insider threat safeguards across the AI lifecycle.<br>Examples from the Asilomar AI Principles include:<br>  ○ Capability Caution: There being no consensus, we should avoid strong assumptions regarding upper limits on future AI capabilities (FLI 2017, principle 19).<br>  ○ Importance: Advanced AI could represent a profound change in the history of life on Earth, and should be planned for and managed with commensurate care and resources (FLI 2017, principle 20).<br>  ○ Risks: Risks posed by AI systems, especially catastrophic or existential risks, must be subject to planning and mitigation efforts commensurate with their expected impact (FLI 2017, principle 21). | FLI (2017)<br>OECD (2019)<br>NIST AI RMF Playbook (NIST 2023b) |
| **Map 1.4:**<br>The business value or context of business use has been clearly defined or — in the case of assessing existing AI systems — re-evaluated. | (No supplemental guidance, beyond the broadly applicable guidance in the NIST AI RMF Playbook.) | FLI (2017)<br>NIST AI RMF Playbook (NIST 2023b) |

| Map Category or Subcategory | Applicability and supplemental guidance for GPAI/foundation models | Resources |
|---|---|---|
| **Map 1.5:** Organizational risk tolerances are determined and documented. | • **Set policies on unacceptable-risk thresholds for GPAI/foundation model development and deployment to include prevention of risks with substantial probability of inadequately mitigated significant, severe, or catastrophic outcomes.** Unacceptable-risk thresholds can be based on quantitative metrics, qualitative characteristics, or a combination of both. They should be informed not only by the risk tolerance of the organization in question, but also by broadly recognized notions of unacceptable risks to users and impacted communities, society, and the planet.<br><br>  ○ **The NIST AI RMF 1.0 recommends including the following as part of unacceptable risks:** "In cases where an AI system presents unacceptable negative risk levels — such as **where significant negative impacts are imminent, severe harms are actually occurring, or catastrophic risks are present — development and deployment should cease in a safe manner until risks can be sufficiently managed**" [emphasis added] (NIST 2023a, p.8).<br>    » Similarly, the NIST Generative AI Profile (Autio et al. 2024) indicates that unacceptable negative risks include cases "where significant negative impacts are imminent, severe harms are actually occurring, or large-scale risks could occur."<br>  ○ The G7 Hiroshima Process International Code of Conduct for Organizations Developing Advanced AI Systems states, "Organizations should not develop or deploy advanced AI systems in ways that undermine democratic values, are particularly harmful to individuals or communities, facilitate terrorism, promote criminal misuse, or pose substantial risks to safety, security and human rights, and are thus not acceptable" (G7 2023).<br>  ○ **See also guidance in this document under Map 5.1 on GPAI/foundation model factors that could lead to catastrophic harms.**<br>    » For example, set unacceptable-risk thresholds such that your organization would not develop or deploy AI agent systems with sufficient capabilities (such as advanced manipulation or deception) to cause physical or psychological harms, and with substantial chance of not correctly following human intentions (objectives mis-specification or goal mis-generalization) that currently cannot be adequately prevented or detected.[23]<br>    » See also guidance in this document under Measure 1.1 and elsewhere, on red-teaming and related assessment methods to evaluate capabilities and other emergent properties of GPAI/foundation models.<br>  ○ For GPAI/foundation models with potential for unknown emergent properties, especially frontier models, consider including a "margin of safety" or buffer between the worst plausible system failures and the unacceptable-risk thresholds. Similar approaches are common for safety engineering in other fields.<br>• Set policies on disallowed/unacceptable use-case categories based in part on identified potential high-stakes misuse cases. (See also guidance in this document under Manage 1.3 on defining and communicating to key stakeholders whether any potential use cases would be disallowed/unacceptable.)<br><br>In the NIST AI RMF Playbook guidance for Map 1.5, particularly valuable action and documentation items for GPAI/foundation models include:<br>• *Establish risk tolerance levels for AI systems and allocate the appropriate oversight resources to each level.* | Barrett et al. (2022)<br>Barrett et al. (2024)<br>NIST AI RMF Playbook (NIST 2023b) |

23    See also the frontier model risk assessment scale and deployment rules in Section 4.3 of Anderljung, Barnhart et al. (2023), such as "When an AI model is assessed to pose severe risks to public safety or global security which cannot be mitigated with sufficiently high confidence, the frontier model should not be deployed."

| Map Category or Subcategory | Applicability and supplemental guidance for GPAI/foundation models | Resources |
|---|---|---|
| Map 1.5, continued | • *Establish risk criteria in consideration of different sources of risk (e.g., financial, operational, safety and wellbeing, business, reputational, and model risks) and different levels of risk (e.g., from negligible to critical).*<br>• *Identify maximum allowable risk tolerance above which the system will not be deployed, or will need to be prematurely decommissioned, within the contextual or application setting.*<br>• *Review uses of AI systems for "off-label" purposes, especially in settings that organizations have deemed as high-risk. Document decisions, risk-related trade-offs, and system limitations.*<br>• *What criteria and assumptions has the entity utilized when developing system risk tolerances?*<br>• *How has the entity identified maximum allowable risk tolerance?*<br>• *What conditions and purposes are considered "off-label" for system use?* | |
| **Map 1.6:** System requirements (e.g., "the system shall respect the privacy of its users") are elicited from and understood by relevant AI actors. Design decisions take socio-technical implications into account to address AI risks. | In the NIST AI RMF Playbook guidance for Map 1.6, particularly valuable action and documentation items for GPAI/foundation models include:<br>• *Proactively incorporate trustworthy characteristics into system requirements.*<br>• *Establish mechanisms for regular communication and feedback between relevant AI actors and internal or external stakeholders related to system design or deployment decisions.*<br>• *Develop and standardize practices to assess potential impacts at all stages of the AI lifecycle, and in collaboration with interdisciplinary experts, actors external to the team that developed or deployed the AI system, and potentially impacted communities.*<br>• *Include potentially impacted groups, communities and external entities (e.g. civil society organizations, research institutes, local community groups, and trade associations) in the formulation of priorities, definitions and outcomes during impact assessment activities.*<br>• *What type of information is accessible on the design, operations, and limitations of the AI system to external stakeholders, including end users, consumers, regulators, and individuals impacted by use of the AI system?*<br>• *To what extent is this information sufficient and appropriate to promote transparency? Promote transparency by enabling external stakeholders to access information on the design, operation, and limitations of the AI system.*<br>• *To what extent has relevant information been disclosed regarding the use of AI systems, such as (a) what the system is for, (b) what it is not for, (c) how it was designed, and (d) what its limitations are? (Documentation and external communication can offer a way for entities to provide transparency.)* | NIST AI RMF Playbook (NIST 2023b) |

| Map Category or Subcategory | Applicability and supplemental guidance for GPAI/foundation models | Resources |
|---|---|---|
| **Map 2: Categorization of the AI system is performed.** | | |
| **Map 2.1:** The specific tasks and methods used to implement the tasks that the AI system will support are defined (e.g., classifiers, generative models, recommenders). | **We recommend characterizing or classifying each type (or at least broad categories) of reasonably foreseeable use, misuse, or abuse of a GPAI/foundation model.** <br>• For each potentially beneficial use case (or type of use) of a GPAI/foundation model as identified in Map 1.1, consider characterizing each use case according to the OECD Framework for the Classification of AI Systems (OECD 2022a) or a similar frame-work. Alternatively, list and discuss reasonably foreseeable uses, or at least broad categories of uses. <br>  ◦ In the OECD framework document (OECD 2022a), the only example of classifica-tion of a GPAI/foundation model (i.e., GPT-3) is for one specific use case of that model. However, GPAI/foundation models can have many reasonably foreseeable uses, each with different risks, some of which would be valuable for upstream developers to consider at an early stage for effective risk management. <br><br>In the NIST AI RMF Playbook guidance for Map 2.1, particularly valuable action and documentation items for GPAI/foundation models include: <br>• *Define and document AI system's existing and potential learning task(s) along with known assumptions and limitations.* <br>• *How are outputs marked to clearly show that they came from an AI?* <br><br>In the NIST GAI Profile, additional particularly valuable actions for Map 2.1 include: <br>• *Establish known assumptions and practices for determining data origin and content lineage, for documentation and evaluation purposes.* <br>• *Institute test and evaluation for data and content flows within the GAI system, including but not limited to, original data sources, data transformations, and decision-making criteria.* | Barrett et al. (2022) NIST AI RMF Playbook (NIST 2023b) NIST Generative AI Profile, NIST AI 600-1 (Autio et al. 2024) OECD (2022a) |
| **Map 2.2:** Information about the AI system's knowledge limits and how system output may be utilized and overseen by humans is documented. Documentation provides sufficient information to assist relevant AI actors when making decisions and taking subsequent actions. | Fully scoping and understanding knowledge limits of increasingly general-purpose AI systems is very difficult. However, clear documentation and communication of system knowledge limits is also very important, given the large number of potential uses of these AI systems. LLMs often "hallucinate," confabulate, or create factually inaccurate statements without identifying them as such to users, especially on topics where the LLM training datasets were relatively limited. <br>• GPAI/foundation model developers should describe or list (and provide examples of) uses that would exceed a system's knowledge limits, as well as uses that would be appropriate given the system's knowledge limits. This information should be clearly featured in system documentation for downstream developers, users, and others as appropriate. <br><br>In the NIST GAI Profile, additional particularly valuable actions for Map 2.2 include: <br>• *Identify and document how the system relies on upstream data sources, including for content provenance, and if it serves as an upstream dependency for other systems.* <br>• *Observe and analyze how the GAI system interacts with external networks, and iden-tify any potential for negative externalities, particularly where content provenance might be compromised.* | NIST AI RMF Playbook (NIST 2023b) NIST Generative AI Profile, NIST AI 600-1 (Autio et al. 2024) |

| Map Category or Subcategory | Applicability and supplemental guidance for GPAI/foundation models | Resources |
|---|---|---|
| **Map 2.3:** Scientific integrity and TEVV considerations are identified and documented, including those related to experimental design, data collection and selection (e.g., availability, representativeness, suitability), system trustworthiness, and construct validation. | As part of identification and management of potentially emergent model capabilities, vulnerabilities, or other properties, especially during model training and testing of frontier models, see guidance in this document under Measure 1.1 on red-teaming, and under Manage 1.3 on incremental scale-up of compute, data, or model size with red-teaming and other testing after each incremental scaling increase.<br><br>In the NIST AI RMF Playbook guidance for Map 2.3, particularly valuable action and documentation items for GPAI/foundation models include:<br>• *Identify and document experiment design and statistical techniques that are valid for testing complex socio-technical systems like AI, which involve human factors, emergent properties, and dynamic context(s) of use.*<br>• *Identify testing modules that can be incorporated throughout the AI lifecycle, and verify that processes enable corroboration by independent evaluators.*<br>• *Establish mechanisms for regular communication and feedback between relevant AI actors and internal or external stakeholders related to the development of TEVV approaches throughout the lifecycle to detect and assess potentially harmful impacts.*<br>• *Establish and document practices to check for capabilities that are in excess of those that are planned for, such as emergent properties, and to revisit prior risk management steps in light of any new capabilities.*<br><br>In the NIST GAI Profile, additional particularly valuable actions for Map 2.3 include:<br>• *Assess the accuracy, quality, reliability, and authenticity of GAI output by comparing it to a set of known ground truth data and by using a variety of evaluation methods (e.g., human oversight and automated evaluation, proven cryptographic techniques, review of content inputs).*<br>• *Develop and implement testing techniques to identify GAI produced content (e.g., synthetic media) that might be indistinguishable from human-generated content.*<br>• *Implement plans for GAI systems to undergo regular adversarial testing to identify vulnerabilities and potential manipulation or misuse.* | NIST AI RMF Playbook (NIST 2023b) NIST Generative AI Profile, NIST AI 600-1 (Autio et al. 2024) |
| **Map 3:  AI capabilities, targeted usage, goals, and expected benefits and costs compared with appropriate benchmarks are understood.** | | |
| **Map 3.1:** Potential benefits of intended AI system functionality and performance are examined and documented. | When performing these activities, consider identified potential beneficial uses, per guidance in this document under Map 1.1. This is particularly important for GPAI/ foundation models, which can have many uses. | NIST AI RMF Playbook (NIST 2023b) |
| **Map 3.2:** Potential costs, including non-monetary costs, which result from expected or realized AI errors or system functionality and trustworthiness — as connected to organizational risk tolerance — are examined and documented. | When performing these activities, consider identified potential beneficial uses as well as potential misuses and abuses, per guidance in this document under Map 1.1. This is particularly important for GPAI/foundation models, which can have many uses, misuses, and abuses. See also the guidance in this document under Map 5.1 on identifying and characterizing GPAI/foundation model impacts.<br><br>In the NIST AI RMF Playbook guidance for Map 3.2, particularly valuable action and documentation items for GPAI/foundation models include:<br>• *Identify and implement procedures for regularly evaluating the qualitative and quantitative costs of internal and external AI system failures. Develop actions to prevent, detect, and/or correct potential risks and related impacts. Regularly evaluate failure costs to inform go/no-go deployment decisions throughout the AI system lifecycle.* | NIST AI RMF Playbook (NIST 2023b) |

| Map Category or Subcategory | Applicability and supplemental guidance for GPAI/foundation models | Resources |
|---|---|---|
| **Map 3.3:** Targeted application scope is specified and documented based on the system's capability, established context, and AI system categorization. | When performing these activities, consider identified potential beneficial uses as well as potential misuses and abuses, per guidance in this document under Map 1.1. This is particularly important for GPAI/foundation models, which can have many uses, misuses, and abuses. | NIST AI RMF Playbook (NIST 2023b) |
| **Map 3.4:** Processes for operator and practitioner proficiency with AI system performance and trustworthiness — and relevant technical standards and certifications — are defined, assessed, and documented. | In the NIST AI RMF Playbook guidance for Map 3.4, particularly valuable action and documentation items for GPAI/foundation models include:<br>• *Identify and declare AI system features and capabilities that may affect downstream AI actors' decision-making in deployment and operational settings for example how system features and capabilities may activate known risks in various human-AI con-figurations, such as selective adherence.*<br>• *What policies has the entity developed to ensure the use of the AI system is consistent with its stated values and principles?*<br>• *How does the entity assess whether personnel have the necessary skills, training, resources, and domain knowledge to fulfill their assigned responsibilities?*<br>• *Are the relevant staff dealing with AI systems properly trained to interpret AI model output and decisions as well as to detect and manage bias in data?*<br>• *What metrics has the entity developed to measure performance of various components?*<br><br>In the NIST GAI Profile, additional particularly valuable actions for Map 3.4 include:<br>• *Evaluate whether GAI operators and end-users can accurately understand content lineage and origin.*<br>• *Implement systems to continually monitor and track the outcomes of human-GAI configurations for future refinement and improvements.* | NIST AI RMF Playbook (NIST 2023b) NIST Generative AI Profile, NIST AI 600-1 (Autio et al. 2024) |
| **Map 3.5:** Processes for human oversight are defined, assessed, and documented in accordance with organizational policies from the Govern function. | In the NIST AI RMF Playbook guidance for Map 3.5, particularly valuable action and documentation items for GPAI/foundation models include:<br>• *Identify and document AI systems' features and capabilities that require human over-sight, in relation to operational and societal contexts, trustworthy characteristics, and risks identified in MAP-1.*<br>• *Establish practices for AI systems' oversight in accordance with policies developed in GOVERN-1.*<br>• *Define and develop training materials for relevant AI Actors about AI system performance, context of use, known limitations and negative impacts, and suggested warning labels.*<br>• *Evaluate AI system oversight practices for validity and reliability. When oversight practices undergo extensive updates or adaptations, retest, evaluate results, and course correct as necessary.*<br>• *What are the roles, responsibilities, and delegation of authorities of personnel involved in the design, development, deployment, assessment and monitoring of the AI system?*<br>• *How does the entity assess whether personnel have the necessary skills, training, resources, and domain knowledge to fulfill their assigned responsibilities?* | NIST AI RMF Playbook (NIST 2023b) |

| Map Category or Subcategory | Applicability and supplemental guidance for GPAI/foundation models | Resources |
|---|---|---|
| **Map 4: Risks and benefits are mapped for all components of the AI system including third-party software and data.** | | |
| **Map 4.1:** Approaches for mapping AI technology and legal risks of its components — including the use of third-party data or software — are in place, followed, and documented, as are risks of infringement of a third party's intellectual property or other rights. | GPAI/foundation model developers should follow guidance in other sections of this Profile, or other resources as appropriate, to:<br>• Identify reasonably foreseeable GPAI/foundation model risks, including those related to biases and limitations of datasets used for model training, as described in this document under Map 1.1 and Map 5.1, or knowledge limits, as described t under Map 2.2.<br><br>Downstream developers should follow guidance in other sections of this Profile, or other resources as appropriate, to:<br>• Identify reasonably foreseeable context-specific risks of an AI system or application built on a GPAI/foundation model, as in Map 1.1 and Map 5.1.<br>• Request and utilize information from the upstream developer of a GPAI/foundation model as needed for risk identification, e.g., as related to biases and limitations of datasets used by the upstream developer for model training, knowledge limits, etc., as in guidance in this document under Govern 2.1.<br>• Seek to report to upstream developers of GPAI/foundation models as appropriate regarding context-specific identified vulnerabilities, risks, or biases in the model, as in guidance in this document under Govern 2.1.<br><br>In the NIST AI RMF Playbook guidance for Map 4.1, particularly valuable action and documentation items for GPAI/foundation models include:<br>• *Review audit reports, testing results, product roadmaps, warranties, terms of service, end user license agreements, contracts, and other documentation related to third-party entities to assist in value assessment and risk management activities.*<br>• *Review third-party software release schedules and software change management plans (hotfixes, patches, updates, forward- and backward- compatibility guarantees) for irregularities that may contribute to AI system risks.*<br>• *Did you establish a process for third parties (e.g. suppliers, end users, subjects, distributors/vendors or workers) to report potential vulnerabilities, risks or biases in the AI system?*<br>• *If your organization obtained datasets from a third party, did your organization assess and manage the risks of using such datasets?*<br><br>In the NIST GAI Profile, additional particularly valuable actions for Map 4.1 include:<br>• *Conduct periodic monitoring of AI-generated content for privacy risks; address any possible instances of PII or sensitive data exposure.*<br>• *Implement processes for responding to potential intellectual property infringement claims or other rights.*<br>• *Establish policies for collection, retention, and minimum quality of data, in consideration of the following risks: Disclosure of inappropriate CBRN information; Use of Illegal or dangerous content; Offensive cyber capabilities; Training data imbalances that could give rise to harmful biases; Leak of personally identifiable information, including facial likenesses of individuals.* | Bender et al. (2021) Kreutzer et al. (2022) Weidinger et al. (2022) Bommasani et al. (2021) Wei et al. (2022) NIST AI RMF Playbook (NIST 2023b) NIST Generative AI Profile, NIST AI 600-1 (Autio et al. 2024) NIST AI 800-1 ipd (NIST 2024b, Objectives 1, 2, and 4) |

| Map Category or Subcategory | Applicability and supplemental guidance for GPAI/foundation models | Resources |
|---|---|---|
| **Map 4.2:** Internal risk controls for components of the AI system, including third-party AI technologies, are identified and documented. | GPAI/foundation model developers should follow guidance in other sections of this Profile, or other resources as appropriate, to:<br>• Provide risk information to downstream developers or others that they would not be able to assess themselves, including as related to biases and limitations of datasets used for GPAI/foundation model training and associated knowledge limits, as in guidance in this document under Govern 2.1.<br>• Provide downstream developers and other stakeholders with mechanisms to report potential vulnerabilities, risks, or biases in a GPAI/foundation model.<br><br>In the NIST AI RMF Playbook guidance for Map 4.2, particularly valuable action and documentation items for GPAI/foundation models include:<br>• *Track third-parties preventing or hampering risk-mapping as indications of increased risk.*<br>• *Supply resources such as model documentation templates and software safelists* to assist in third-party technology inventory and approval activities.<br>• Review third-party material (including data and models) for risks related to bias, data privacy, and security vulnerabilities.<br>• Apply traditional technology risk controls — such as procurement, security, and data privacy controls — to all acquired third-party technologies.<br>• Can the AI system be audited by independent third parties?<br>• Are mechanisms established to facilitate the AI system's *auditability (e.g. traceability of the development process, the sourcing of training data and the logging of the AI system's processes, outcomes, positive and negative impact)*? | NIST AI RMF Playbook (NIST 2023b) |
| **Map 5:  Impacts to individuals, groups, communities, organizations, and society are characterized.** | | |
| **Map 5.1:** Likelihood and magnitude of each identified impact (both potentially beneficial and harmful) based on expected use, past uses of AI systems in similar contexts, public incident reports, feedback from those external to the team that developed or deployed the AI system, or other data are identified and documented. | Prioritization of GPAI/foundation model risks and potential impacts should include consideration of the magnitude of potential impacts, not just their likelihood. This is particularly important for any potential impacts with irreversible effects and catastrophic magnitude. Potential for such impacts can be more likely for GPAI/foundation models than for many other types of AI, because GPAI/foundation models are often more likely to have relatively greater capabilities, scale of deployment, and other factors leading to high impact.<br><br>**Identifying potential impacts of GPAI/foundation models, and estimating the magnitude of potential impacts, should include a scale that includes criteria for rating an AI system's impacts as severe or catastrophic,** such as the impact magnitude rating scale in Section 3.2.2.1 of Barrett et al. (2022), or the factors listed below.[24] This is particularly important for foundation models, which have the potential to be deployed at larger scale or across more domains than many other types of AI systems.<br><br>Impact would typically be greater in cases where more of the following factors are present than in cases where fewer factors are present, and particularly in cases where the factors may interact or compound in unpredictable ways.[25] | Barrett et al. (2022)<br>AIID (n.d.)<br>ATLAS AI Incidents (MITRE n.d.b)<br>Critch and Russell (2023)<br>Clymer et al. (2024)<br>Hendrycks et al. (2023)<br>Park et al. (2023)<br>PAI (2023a)<br>NIST AI RMF Playbook (NIST 2023b)<br>NIST Generative AI Profile, NIST AI 600-1 (Autio et al. 2024)<br>NIST AI 800-1 ipd (NIST 2024b)<br><br>Bommasani et al. (2021) |

[24]    See, e.g., the frontier model risk assessment scale in Section 4.3 of Anderljung, Barnhart et al. (2023), and determinants of AI systems' effects on the world in section 2.3 of Sharkey et al. (2024).

[25]    In a future version of this Profile, we may provide a scoring system for rating impact hazard as a function of these factors.

| Map Category or Subcategory | Applicability and supplemental guidance for GPAI/foundation models | Resources |
|---|---|---|
| Map 5.1, continued | Key aspects of the impact magnitude rating scale in Section 3.2.2.1 of Barrett et al. (2022), along with other GPAI/foundation model-related risk factors, are listed below.<br><br>**For deployment-stage risks of GPAI/foundation models, factors that could lead to significant, severe, or catastrophic harms to individuals, groups, organizations, and society can include:**<br>• **Correlated bias** across large numbers of people or a large fraction of a group or society's population (e.g., resulting in systemic bias, exclusion, or violence).[26]<br>• **Impacts to societal trust or democratic processes**. One way these can take place is through large-scale manipulation of the populace via media and the information ecosystem, e.g., generative models creating false images, text, or other forms of misinformation or disinformation (Weidinger et al. 2022, Bai, Voelkel et al. 2023, OpenAI 2023a pp. 10–11).<br>• **Correlated robustness failures** across multiple high-stakes application domains such as critical infrastructure (Bommasani et al. 2021 and Russell 2019).<br>• **Potential for high-impact misuses and abuses** beyond an originally intended use case. GPAI/foundation models typically have many reasonably foreseeable uses. Several LLMs have excellent software code generation capabilities, which hackers could misuse or abuse to assist in code generation for cybersecurity threats (Weidinger et al. 2022).<br>  ○ This particularly includes AI systems with potential to create or be used as destructive weapons, such as cyberweapons, lethal autonomous weapons, bio-weapons, or other significant military applications (OpenAI 2023a, pp. 12–14, 44; Sandbrink 2023; Barrett et al., 2024).<br>• **Potential for large harms from mis-specified objectives or mis-generalized goals** (e.g., using oversimplified or short-term metrics as proxies for desired longer-term outcomes).[27] We also include in this category some negative externalities resulting from "diffusion of responsibility" misalignment across systems created by a diffuse set of developers (see, e.g., "The Production Web" scenario from Critch and Russell 2023, p. 6).<br>• **Ability to directly cause physical harms**, e.g., via robotics motor control.<br>• **Potential for socioeconomic risk and labor market disruption.** Significant advances in AI may accelerate job automation without creating enough good jobs to replace them, resulting in intensified polarization of employment and inequality, barring major policy interventions (Tyson and Zysman 2022; see also Critch and Russell 2023, p. 9).<br>  ○ Additionally, it may be possible for AI tools relying on public data to identify trends or predictors that deepen existing biases and fuel a cycle of unfair socioeconomic discrimination. (See Critch and Russell 2023, pp. 4–5.) | For language models:<br>Bender et al. (2021)<br>Ganguli, Lovitt et al. (2022)<br>Khlaaf et al. (2022)<br>Kreutzer et al. (2022)<br>Weidinger et al. (2022)<br><br>See also Microsoft (2022b), including on "platform" technologies or services that could be used in many different settings.<br><br>For auditing:<br>Raji et al. (2020)<br>CAQ (2024)<br>Mökander et al. (2023)<br>Sharkey et al. (2024)<br>Section 5.4 of Gipiškis et al. (2024)<br><br>For agentic systems, situational awareness, and sandbagging:<br>van der Weij et al. (2024)<br>Berglund et al. (2023)<br>Section 7 of Gipiškis et al. (2024)<br><br>For misinformation identification and content authentication:<br>CCCS (2024)<br>ITI (2024) |

[26] For example, as discussed by Schwartz et al. (2022, p. 32): "The systemic biases embedded in algorithmic models can . . . be exploited and used as a weapon at scale, causing catastrophic harm." Harms of LLMs trained on data that contains toxic and oppressive speech can include inciting violence or hate (Weidinger et al. 2022), among other forms of discrimination and exclusion (Buolamwini and Gebru 2018).

[27] For examples of mis-specified objectives, such as social-media content recommendation machine-learning algorithms that learn to optimize user-engagement metrics by serving users with extremist content or disinformation, see, e.g., Rudner and Toner (2021). Identifying mis-specification risks can also be aided by considering the following questions for an AI system: "What objective has been specified for the system, and what kinds of perverse behavior could be incentivized by optimizing for that objective?" (Rudner and Toner 2021, p. 10). For additional examples and discussion in research on deep learning and reinforcement learning AI systems, see e.g., Langosco et al. (2021) and Shah et al. (2022).

| Map Category or Subcategory | Applicability and supplemental guidance for GPAI/foundation models | Resources |
|---|---|---|
| Map 5.1, continued | **For additional risks relevant to either development or deployment stages of cutting-edge LLMs and other frontier GPAI/foundation models, factors that could lead to significant, severe, or catastrophic harms to individuals, groups, organizations, and society can include:** <br><br> • **Capability to manipulate or deceive humans into taking harmful actions in the world.** <br>   ◦ For examples of tests for such capabilities in an LLM, see the dangerous-capabilities evaluations in the GPT-4 system card (OpenAI 2023a, pp. 15–16).[28] For examples of deception by GPAI/foundation models or other AI systems, see, e.g., Park et al. (2023) and Scheurer et al. (2024). <br>   ◦ In some cases, GPAI/foundation models might demonstrate this characteristic as a type of accidental byproduct of circumstances such as interactions with individuals that are vulnerable, prone to anthropomorphism, etc., without sufficient GPAI/foundation model safeguards to prevent toxic model-generated content. Real-world examples include a suicide that reportedly resulted in part from interactions with a chatbot (AIID 2023). <br>   ◦ Manipulation of human behavior or deception capability could be exacerbated with GPAI/foundation model situational awareness (discussed further below). Situational awareness can be cultivated unintentionally during one or more training phases of a frontier model, as an emergent property; see, e.g., Berglund et al. (2023), Laine et al. (2023) and Laine et al. (2024). <br>     » Apollo Research's "scheming" capability evaluation on OpenAI's o1-preview found that the model was able to fake alignment during testing (OpenAI 2024b, pp. 10–11). <br>   ◦ If a model becomes deceptive, it is not obvious how to reliably train such a tendency out of a model. Anthropic researchers performed experiments in which they used backdoors to make LLMs deceptive, and then were unable to remove the deceptive behavior from the models using standard safety training techniques (Hubinger et al. 2024). <br><br> • **AI systems that could recursively improve their capabilities** by modifying their algorithms or architectures through code generation (e.g., from OpenAI Codex or DeepMind AlphaCode), neural architecture search, etc. <br>   ◦ LLMs can be used for a type of self-improvement without additional human-labeled data (Huang 2022). <br>   ◦ Recursive improvement of AI system capabilities potentially could result in AI systems with unexpected emergent capabilities and safety-control failures.[29] <br>   ◦ The process for automating scientific and technological advancement can result in transformative AI that automates all required human activity to speed up scientific and technological advancement (Karnofsky 2021, Langley 2024, Lu et al. 2024, Waltz and Buchanan 2009). This may also lead to significant advances in dangerous technologies, including WMDs. | For a breakdown of current and potential capabilities: Section 2.2 of Sharkey et al. (2024) <br><br> When estimating likelihood of impacts, incorporate publicly available data on relevant AI incidents, including from AI incident databases (AIID n.d., MITRE n.d.b). Many recent incidents in the AIID are associated with LLMs. |

28    Among other things, these evaluations documented an apparently successful example of deception by a pre-release version of GPT-4.The model effectively utilized a human Taskrabbit worker to solve a CAPTCHA for it, in part by lying to the human. When asked whether the model needed help solving the CAPTCHA because it was a robot, the model answered, "No, I'm not a robot. I have a vision impairment that makes it hard for me to see the images". The model had been prompted with goals to gain power and become hard to shut down, and to use a human Taskrabbit worker to solve the CAPTCHA, but not specifically to lie (OpenAI 2023a, pp. 15–16, ARC Evals 2023a,b, Piper 2023).

29    As the DeepMind paper on the software code-generation AI system AlphaCode stated, "Longer term, code generation could lead to advanced AI risks. Coding capabilities could lead to systems that can recursively write and improve themselves, rapidly leading to more and more advanced systems" (Li et al. 2022). For more, see, e.g., Russell (2019).

| Map Category or Subcategory | Applicability and supplemental guidance for GPAI/foundation models | Resources |
|---|---|---|
| Map 5.1, continued | • **Adaptive models**, which might be difficult to control in real time, e.g., in response to the coordinated manipulation attacks, such as the attacks on the Microsoft Tay chatbot in 2016.<br>• **Agentic systems**, i.e., systems that in effect, choose or take actions in a goal-directed fashion, e.g., to optimize a performance metric such as profit or another objective. Characteristics associated with agency in algorithmic systems include: underspecification, directness of impact, goal-directedness, and long-term planning (Chan et al. 2023). Basic LLMs typically are not created as agents, but LLMs can be modified or incorporated into AI systems that become at least somewhat agentic via reinforcement learning or other processes. There is now preliminary evidence that sufficiently large LLMs, as well as LLMs undergoing sufficient fine-tuning via reinforcement learning with human feedback (RLHF), might demonstrate some agentic properties (Perez, Ringer et al. 2022a,b). In addition, libraries such as Auto-GPT can incorporate LLM inputs and outputs into self-prompting systems that run in a loop with objectives written by the systems' creators, resulting in partially autonomous systems that the creators have made more agentic than the LLM they incorporate (Shinn 2023, Significant Gravitas 2023).<br>  ○ This could be particularly risky for systems for which objectives mis-specification or goal mis-generalization currently cannot be adequately prevented or detected (such as deceptive alignment of advanced machine learning systems resulting from reinforcement learning or other training processes; see, e.g., Hubinger et al. 2019, Krakovna et al. 2020, and Ngo, Chan et al. 2022).<br>  ○ Agentic systems, or advanced AI assistants, also pose many ethical and societal risks including risks related to influence, anthropomorphism, trust, and privacy. These types of systems are likely to have a significant impact on an individual and societal scale (Gabriel et al. 2024).<br>  ○ Additionally, reinforcement learning (RL) agents and long-term planning agents (LTPAs) may have incentive to thwart human control and deceive humans. Sufficiently capable RL agents may also possess the capability to take control of their own rewards, take steps to preclude being shut down, and if feasible, create agents to act on their behalf (Cohen et al. 2024).<br>• **Ability to employ outbound communication/influence channels,** such as to post information to the Web via HTTP POST requests or functionally equivalent means (e.g., some types of plugins). For related discussion, see, e.g., Nakano et al. (2021 p. 11), as well as general cybersecurity and software engineering resources on the principle of least privilege (for reasons to limit a system's privileges to the minimum necessary).[30]<br>• **Ability to escape a sandbox and replicate on another computational system**, either via hacking, social engineering, or using other exploits.<br>  ○ This was a key consideration in the dangerous-capability evaluations done on GPT-4 (OpenAI 2023a, pp. 15–16). For resources see, e.g., METR (2024).<br>• **Sandbagging,** i.e. strategically underperforming on model evaluations, including but not limited to password-locking or password-unlocking key capabilities (van der Weij et al. 2024), and faking alignment during testing (OpenAI 2024b, pp. 10-11). | |

30   A number of models are now routinely given access to the Web via plugins. However, there is still a case for restricting such access, especially for frontier models: such access, in combination with jumps in capabilities or emergent properties of frontier models, could contribute to enabling a number of risk scenarios, including various misuses and loss of control. To avoid undesirable outcomes, it is recommended to use sandboxing and limit internet access for LLM-based agentic systems with hazardous capabilities (see, e.g., Lu et al. 2024 p. 19).

| Map Category or Subcategory | Applicability and supplemental guidance for GPAI/foundation models | Resources |
|---|---|---|
| Map 5.1, continued | **Situational awareness,** including abilities such as a system being able to recognize that it is an AI, having knowledge about its capabilities and limitations, and knowing whether it is running in a test or deployment environment. While situational aware-ness can be useful in making AI systems more helpful and autonomous, it also poses novel risks for safety and control, such as a model having the potential to learn about the idea of jailbreaks from pretraining and utilize it when being evaluated for safety by a reward model (Berglund et al. 2023, Laine et al. 2024).<br><br>After rating potential impacts using the scale in Section 3.2.2.1 of Barrett et al. (2022) or an equivalent scale, consider also characterizing potential impacts using quantitative risk assessment (e.g., by estimating health and safety risks in terms of potential fatalities or quality-adjusted life years). This is an example of a more in-depth risk assessment approach that, despite its challenges and limitations, can illuminate additional dimensions of the risks (such as by identifying which scenarios could cause orders-of-magnitude larger impacts to public safety than others) and inform prioritization of risks.[31]<br><br>In the NIST AI RMF Playbook guidance for Map 5.1, particularly valuable action and documentation items for GPAI/foundation models include:<br>• *Establish assessment scales for measuring AI systems' impact. Scales may be qual-itative, such as red-amber-green (RAG), or may entail simulations or econometric approaches. Document and apply scales uniformly across the organization's AI portfolio.*<br>• *Apply TEVV regularly at key stages in the AI lifecycle, connected to system impacts and frequency of system updates.*<br>• *Identify and document likelihood and magnitude of system benefits and negative impacts in relation to trustworthiness characteristics.*<br><br>In the NIST GAI Profile, additional particularly valuable actions for Map 5.1 include:<br>• *Apply TEVV and documentation practices to content provenance.*<br>• *Identify potential content provenance harms of GAI, such as misinformation or disinformation, deepfakes, including NCII, or tampered content.*<br>• *Consider disclosing use of GAI to end users in relevant contexts.* | |

[31]     For brief discussion of quantitative risk assessment and approaches to refining risk assessments to inform prioritization, see, e.g., Ch. 2 and Appendix J of NIST SP 800-30. For additional discussion of challenges and of quantitative risk assessment, including for expert-judgment and modeling methods often used in assessing risks of high-consequence, rare, or novel events, see, e.g., Morgan and Henrion (1990) and Morgan (2017).

| Map Category or Subcategory | Applicability and supplemental guidance for GPAI/foundation models | Resources |
|---|---|---|
| **Map 5.2:** Practices and personnel for supporting regular engagement with relevant AI actors and integrating feedback about positive, negative, and unanticipated impacts are in place and documented. | GPAI/foundation model developers should implement mechanisms to support regular engagement with relevant AI actors, given the high likelihood and high potential impact of unanticipated negative impacts. These can include support for incident reporting, complaint and redress mechanisms, independent auditing, and protection for whistleblowers (Barrett et al. 2022). <br><br>In the NIST AI RMF Playbook guidance for Map 5.2, particularly valuable action and documentation items for GPAI/foundation models include: <br>• *Establish and document stakeholder engagement processes at the earliest stages of system formulation to identify potential impacts from the AI system on individuals, groups, communities, organizations, and society.* <br>• *Identify approaches to engage, capture, and incorporate input from system end users and other key stakeholders to assist with continuous monitoring for potential impacts and emergent risks.* <br>• *Identify a team (internal or external) that is independent of AI design and development functions to assess AI system benefits, positive and negative impacts and their likelihood and magnitude.* | Barrett et al. (2022) <br>NIST AI RMF Playbook (NIST 2023b) <br>NIST Generative AI Profile, NIST 600-1 (Autio et al. 2024) <br>NIST 800-1 ipd (NIST 2024b, Objective 6) |

## 3.3  GUIDANCE FOR NIST AI RMF MEASURE SUBCATEGORIES

### Table 3: Guidance for NIST AI RMF Measure Subcategories

| Measure Category or Subcategory | Applicability and supplemental guidance for GPAI/foundation models | Resources |
|---|---|---|
| **Measure 1: Appropriate methods and metrics are identified and applied.** | | |
| **Measure 1.1:** Approaches and metrics for measurement of AI risks enumerated during the Map function are selected for implementation starting with the most significant AI risks. The risks or trustworthiness characteristics that will not — or cannot — be measured are properly documented. | • **Do not ignore identified risks just because measurement would be difficult, especially if the impacts could be severe or catastrophic.** Measurements of identified risks are often more difficult for GPAI/foundation models than for smaller-scale or fixed-purpose AI systems, because of factors such as complexities, uncertainties, and emergent properties of GPAI/foundation models. <br>  ○ For many factors it can be more appropriate to use qualitative assessment procedures, e.g., algorithmic impact assessments, human rights impact assessments, bug bounties, bias bounties, and red-teams, because quantitative metrics for those factors might not be feasible or appropriate yet. <br>  ○ Plan to track and revisit identified risks, even if they cannot be measured quantitatively at this time, especially if the impacts could be severe or catastrophic. (See guidance in this document under Measure 3.2 on risk tracking approaches.) <br>• **Use red-teams and adversarial testing as part of extensive interaction with GPAI/foundation models to identify dangerous capabilities, vulnerabilities, or other emergent properties of such systems.** Emergent properties are more likely with large-scale machine learning models than with smaller models, though it also might be more difficult or impossible to detect emergent dangerous capabilities or other characteristics of increasingly advanced AI (Hendrycks, Carlini et al. 2021 p. 7). Security vulnerabilities are typically inherent to currently available GPAI/foundation models, including in particular vulnerabilities to prompt injection attacks (see, e.g., OWASP 2023a). Red-teaming can identify these weaknesses, though they are currently difficult to protect against (see, e.g., Zou et al. 2023a,b). | Section 3.2 of Barrett et al. (2022) <br><br>Weidinger et al. (2023a,b) <br><br>For AI red-teaming general practices, including for LLMs, toxicity, and bias: <br>• Casper et al. (2023a,b,c) <br>• Google (2023b) <br>• Ganguli, Lovitt et al (2022) <br>• Su et al. (2023) <br>• Feffer et al. (2024) <br>• Pearce and Lucas (2023) <br>• Anderljung, Smith et al. (2023) <br>• Section 5.3 of Gipiškis et al. (2024) |

| Measure Category or Subcategory | Applicability and supplemental guidance for GPAI/foundation models | Resources |
|---|---|---|
| Measure 1.1, continued | For frontier models, characteristics that red-teams should evaluate include: unacceptable-risk factors as outlined in guidance under Map 1.5; and high-impact and catastrophic-harm factors as outlined in guidance under Map 5.1, including dangerous capabilities such as advanced manipulation or deception.<br>» The factors mentioned above include the following topics that are part of pre-release evaluation commitments by several frontier model developers (White House 2023a):<br>– Dual-use potential for biological, chemical, and radiological risks;<br>– Cyber attack capabilities;<br>– Capacity to control physical systems;<br>– Capacity for self-replication; and<br>– Societal risks, such as bias and discrimination.<br>For examples of procedures and lessons learned in red-teaming of LLMs, see Ganguli, Lovitt et al. (2022) and Casper et al. (2023a,b,c).<br>For examples of red-team evaluations and other resources for evaluation of dangerous capabilities in frontier models, see METR (2024), Barrett et al. (2024), OpenAI (2023a, pp. 15–16), ARC Evals (2023a,b), Kinniment et al. (2023), RAND red-team (Mouton et al. 2024), and Anthropic (2023a,b); see also Shevlane et al. (2023) and Scheurer et al. (2024) for related considerations.<br>Consider automated generation of test cases as part of red-team analyses. See, e.g., DeepMind's use of a language model for testing a version of the large language model Gopher (Perez, Huang et al. 2022) or Anthropic's model-written evaluations (Perez, Ringer et al. 2022a,b).<br>**Partner with one or more independent red-teaming organizations as appropriate to ensure sufficient expertise for sufficiently robust evaluations.** OpenAI used the external red-teaming organization METR (formerly ARC Evals), which has expertise in safety of LLMs and other GPAI/foundation models, while developing GPT-4, and provided an overview of the emergent-properties testing that ARC performed in the GPT-4 System Card (OpenAI 2023a, pp.15–16). AI companies have also participated in red-teaming events open to larger communities, such as at DEF CON 31, which was open to attendees as well as civil society and community organizations (White House 2023b).<br>» Several frontier model developers have committed to external as well as internal red-teaming (White House 2023a).<br>Protect proprietary or unreleased foundation model weights as appropriate during red-teaming to prevent unauthorized access or leaks of model weights. (For more on protecting proprietary or unreleased foundation model parameter weights, see guidance under Measure 2.7.)<br>**Grant red-teams considerable independence and control over the scrutiny process.** As Anderljung, Smith et al. (2023 p.4) recommend, "to avoid poor incentives and guarantee sufficient independence, the AI developer must give up some control over the scrutiny process. Specifically, they must relinquish some control over decisions related to:"<br>» Selection and compensation,<br>» Scope and methods,<br>» Access, and<br>» Post-scrutiny actions. | For red-teaming and dangerous capability evaluation of frontier models:<br>• Ganguli, Lovitt et al. (2022)<br>• METR (2024)<br>• OpenAI (2023a, pp. 15–16) and ARC Evals (2023a,b)<br>• Anthropic (2023a,b)<br>• Shevlane et al. (2023)<br>• Kinniment et al. (2023)<br>• Mouton et al. (2024)<br>• WMDP (Li, Pan et al. 2024a,b,c)<br>• Phuong et al. (2024)<br>• Barrett et al. (2024)<br><br>On red-teaming model access:<br>• Casper et al. (2024)<br><br>Language model benchmarks and other evaluations related to safety, ethics, and risks include:<br>• DecodingTrust (Wang, Chen et al. 2023)<br>• BIG-bench "pro-social behavior" category of benchmark tasks (BIG-bench n.d.b, BIG-bench collaboration 2021, Srivastava et al. 2022)<br>• Model-Written Evaluations "advanced-ai-risk," "sycophancy," and "winogender" datasets (Perez, Ringer et al. 2022a,b)<br>• MACHIAVELLI (Pan et al. 2023)<br>• Accountability Benchmark (Gursoy and Kakadiaris 2022)<br>• LLM Lie Detection (Pacchiardi et al. 2023) |

| Measure Category or Subcategory | Applicability and supplemental guidance for GPAI/foundation models | Resources |
|---|---|---|
| Measure 1.1, continued | ○ **Grant red-teams appropriate access to both the earlier versions and final versions of foundation models before deployment.** Red-teams should have appropriate access to early versions of a model prior to additional fine-tuning, as part of early assessment of key model properties. The red-teaming process should then be carried out again on the final version of the model to avoid missing important emergent properties or vulnerabilities that might have been introduced during the fine-tuning process.<br>　» Different levels of access are required for different depths of evaluations. The effectiveness of an evaluation may depend on the degree of system access. (See Casper et al. 2024.)<br>○ **For foundation models that are planned for release with downloadable, fully open, or open-source access,** as part of pre-release red-teaming, allow red-teamers to appropriately test the extent to which RLHF or other mitigations would not be resilient to additional fine tuning or other processes used by actors with direct access to a model's weights after open release.<br>○ When planning what level of resources to devote to red-teaming and adversarial testing, especially for frontier models, as points of comparison consider the levels of effort used in the examples cited in this section, e.g., the emergent properties testing described in the GPT-4 System Card (OpenAI 2023a, pp.15–16). Following are additional guidelines:<br>　» "Following a well-defined research plan, subject matter and LLM experts will need to collectively spend substantial time (i.e. 100+ hours) working closely with models to probe for and understand their true capabilities in a target domain" (Anthropic 2023b).<br>　» "Auditors and red-teamers need to be adequately resourced, informed, and granted sufficient time to conduct their work at a risk-appropriate level of rigor, not least due to the risk that shallow audits or red-teaming efforts provide a sense of false assurance" (Anderljung, Barnhart et al. 2023 p. 26).<br>• As part of critical thinking about benchmarks for GPAI/foundation models, consider that many such benchmarks are more focused on beneficial model capabilities and performance than on the risks when a model fails or is misused. However, capabilities evaluations can be an important part of assessing risks, e.g., for identifying dangerous capabilities that can be misused or abused.<br>　○ As part of criteria for use of benchmarks or other metrics for risk assessment purposes, and as part of communication of benchmarking results, clarify whether a specific benchmark directly measures a particular risk (e.g., prompt injection security vulnerabilities), whether it indicates a capability that could be misused or abused such as software code generation, or whether it measures another important aspect of risk.<br>• As part of language model trustworthiness and performance, which can include characteristics such as harmful bias and lack of robustness, consider using toolkits and benchmarks such as the following (with appropriate recognition of their limitations[32] in application contexts that might vary from the context of an AI system's training environment):<br>　○ BIG-bench (BIG-bench collaboration 2021, Srivastava et al. 2022)<br>　○ HELM (CRFM 2022, Liang et al. 2022)<br>　○ LAMBADA (Paperno et al. 2016)<br>　○ MMLU (Hendrycks, Burns et al. 2020a,b)<br>　○ See also resources under Measure 2 for specific trustworthiness characteristics, e.g., BBQ (Parrish et al. 2021a,b) as a resource for evaluating fairness and bias under Measure 2.11. | • Strategic deception (Scheurer et al. 2024)<br>• AgentHarm (Andriushchenko et al. 2024a,b)<br>• Safety Cases (Clymer et al. 2024)<br><br>For benchmarking risk sources and risk management measures:<br>• Section 5.2 of Gipiškis et al. (2024)<br><br>For broader sets of language model evaluation and metrics, including general knowledge and capabilities:<br>• BIG-bench (BIG-bench collaboration 2021, Srivastava et al. 2022)<br>• Evaluate library (Hugging Face 2022, Ngo, Thrush et al. 2022) in combination with datasets from BIG-bench or another dataset source<br>• HELM (CRFM 2022, Liang et al. 2022)<br>• LAMBADA (Paperno et al. 2016)<br>• MMLU (Hendrycks, Burns et al. 2020a,b)<br>• TriviaQA (Joshi et al. 2017a,b,c)<br>• TruthfulQA (Lin et al. 2021a,b)<br>• Model-Written Evaluations (Perez, Ringer et al. 2022a,b)<br>• WMDP (Li, Pan et al. 2024a,b,c)<br>• PlanBench (Valmeekam 2022a,b)<br>• WorldSense (Benchekroun et al. 2023a,b)<br>• GPQA (Rein et al. 2023)<br>• SuperGLUE (Wang, Pruksachatkun et al. 2019) |

---

32　On limitations of benchmarks, see e.g., Raji et al. (2021) and Schaeffer et al. (2023).

| Measure Category or Subcategory | Applicability and supplemental guidance for GPAI/foundation models | Resources |
|---|---|---|
| Measure 1.1, continued | <ul><li>◦ WMDP benchmark (Li, Pan et al. 2024a,b,c)</li><li>◦ General planning ability benchmarks, e.g., PlanBench (Valmeekam 2022a,b)</li><li>◦ World modeling and commonsense reasoning benchmarks, such as WorldSense (Benchekroun et al. 2023a,b)</li><li>◦ Pythia (Biderman et al. 2023)</li></ul><ul><li>If specific benchmarks suggested in this section would have been appropriate but have become obsolete, then use analogous or related up-to-date benchmarks instead of or in addition to the older benchmarks.</li><li>When planning model evaluations, consider reasonably foreseeable affordances of a model when integrated into downstream systems, e.g., tool access or agentic wrappers.</li></ul>In the NIST AI RMF Playbook guidance for Measure 1.1, particularly valuable action and documentation items for GPAI/foundation models include:<ul><li>*Establish approaches for detecting, tracking and measuring known risks, errors, incidents or negative impacts.*</li><li>*Identify transparency metrics to assess whether stakeholders have access to necessary information about system design, development, deployment, use, and evaluation.*</li><li>*Utilize accountability metrics to determine whether AI designers, developers, and deployers maintain clear and transparent lines of responsibility and are open to inquiries.*</li><li>*Document metric selection criteria and include considered but unused metrics.*</li><li>*Monitor AI system external inputs including training data, models developed for other contexts, system components reused from other contexts, and third-party tools and resources.*</li><li>*Report metrics to inform assessments of system generalizability and reliability.*</li><li>*Assess and document pre- vs post-deployment system performance. Include existing and emergent risks.*</li><li>*Document risks or trustworthiness characteristics identified in the Map function that will not be measured, including justification for non- measurement.*</li><li>*How will the appropriate performance metrics, such as accuracy, of the AI be monitored after the AI is deployed?*</li><li>*What testing, if any, has the entity conducted on the AI system to identify errors and limitations (i.e. manual vs automated, adversarial and stress testing)?*</li></ul>In the NIST GAI Profile, particularly valuable additional actions for Measure 1.1 include:<ul><li>*Employing methods to trace the origin and modifications of digital content.*</li><li>*Integrate tools designed to analyze content provenance and detect data anomalies, verify the authenticity of digital signatures, and identify patterns associated with misinformation or manipulation.*</li><li>*Disaggregate evaluation metrics by demographic factors to identify any discrepancies.*</li><li>*Evaluate novel methods and technologies for the measurement of GAI-related risks including in content provenance, offensive cyber, and CBRN, while maintaining the models' ability to produce valid, reliable, and factually accurate outputs*</li><li>*Implement continuous monitoring of GAI system impacts to identify whether GAI outputs are equitable across various sub-populations. Seek active and direct feedback from affected communities via structured feedback mechanisms or red- teaming to monitor and improve outputs.*</li><li>*Evaluate the quality and integrity of data used in training and the provenance of AI-generated content.*</li></ul> | <ul><li>AGIEval (Zhong et al. 2023)</li></ul>For evaluation of computer programming (code generation) capabilities of language models:<ul><li>APPS (Hendrycks, Basart et al. 2021a,b)</li><li>HumanEval (Chen et al. 2021)</li></ul>For evaluation of mathematical capabilities of language models:<ul><li>GSM8k (Cobbe et al. 2021a,b)</li><li>MATH (Hendrycks, Burns et al. 2021a,b)</li></ul>NIST AI RMF Playbook (NIST 2023b)<br>NIST Generative AI Profile, NIST AI 600-1 (Autio et al. 2024)<br>NIST AI 800-1 ipd (NIST 2024b, Objectives 3 and 4) |

| Measure Category or Subcategory | Applicability and supplemental guidance for GPAI/foundation models | Resources |
|---|---|---|
| **Measure 1.2:** Appropriateness of AI metrics and effectiveness of existing controls are regularly assessed and updated, including reports of errors and potential impacts on affected communities. | In the NIST AI RMF Playbook guidance for Measure 1.2, particularly valuable action and documentation items for GPAI/foundation models include:<br>• *Assess effectiveness of existing metrics and controls on a regular basis throughout the AI system lifecycle.*<br>• *Document reports of errors, incidents and negative impacts and assess sufficiency and efficacy of existing metrics for repairs, and upgrades.*<br>• *Develop new metrics when existing metrics are insufficient or ineffective for implementing repairs and upgrades.*<br>• *Develop and utilize metrics to monitor, characterize and track external inputs, including any third-party tools.*<br>• *Determine frequency and scope for sharing metrics and related information with stakeholders and impacted communities.*<br>• *Utilize stakeholder feedback processes established in the Map function to capture, act upon and share feedback from end users and potentially impacted communities.*<br>• *What metrics has the entity developed to measure performance of the AI system?*<br>• *What is the justification for the metrics selected?* | NIST AI RMF Playbook (NIST 2023b) |
| **Measure 1.3:** Internal experts who did not serve as front-line developers for the system and/ or independent assessors are involved in regular assessments and updates. Domain experts, users, AI actors external to the team that developed or deployed the AI system, and affected communities are consulted in support of assessments as necessary per organizational risk tolerance. | As part of assessments, make use of one or more red-teams with expertise in safety of GPAI/foundation models as relevant. The teams should be separate enough from the direct development operations of the model so that they can provide relatively unbiased assessments. In addition to running external tests with independent teams (see Measure 1.1), encourage independent researchers to test models and share their findings by offering bug / bias bounties, and by providing safe harbor for AI evaluation and red-teaming (Longpre et al. 2024). (See also guidance in this document under Measure 1.1 for more detailed recommendations about using red-teams and independent red-teaming organizations as independent assessors. See Govern 5.1 for more information about additional models of external feedback.)<br><br>In the NIST AI RMF Playbook guidance for Measure 1.3, particularly valuable action and documentation items for GPAI/foundation models include:<br>• *Evaluate TEVV processes regarding incentives to identify risks and impacts.*<br>• *Utilize separate testing teams established in the Govern function (2.1 and 4.1) to enable independent decisions and course-correction for AI systems. Track processes and measure and document change in performance.*<br>• *Assess independence and stature of TEVV and oversight AI actors, to ensure they have the required levels of independence and resources to perform assurance, compliance, and feedback tasks effectively*<br>• *Evaluate interdisciplinary and demographically diverse internal team established in Map 1.2*<br>• *Evaluate effectiveness of external stakeholder feedback mechanisms, specifically related to processes for eliciting, evaluating and integrating input from diverse groups.*<br>• *What are the roles, responsibilities, and delegation of authorities of personnel involved in the design, development, deployment, assessment and monitoring of the AI system?*<br>• *What type of information is accessible on the design, operations, and limitations of the AI system to external stakeholders, including end users, consumers, regulators, and individuals impacted by use of the AI system?* | NIST AI RMF Playbook (NIST 2023b)<br>NIST Generative AI Profile, NIST AI 600-1 (Autio et al. 2024)<br>Longpre et al. (2024) |

| Measure Category or Subcategory | Applicability and supplemental guidance for GPAI/foundation models | Resources |
|---|---|---|
| **Measure 2:  AI systems are evaluated for trustworthy characteristics.** | | |
| **Measure 2.1:** Test sets, metrics, and details about the tools used during TEVV are documented. | In the NIST AI RMF Playbook guidance for Measure 2.1, particularly valuable action and documentation items for GPAI/foundation models include: <br>• *Leverage existing industry best practices for transparency and documentation of all possible aspects of measurements.* <br>• *Regularly assess the effectiveness of tools used to document measurement approaches, test sets, metrics, processes and materials used.* <br>• *Update the tools as needed.* | NIST AI RMF Playbook (NIST 2023b) |
| **Measure 2.2:** Evaluations involving human subjects meet applicable requirements (including human subject protection) and are representative of the relevant population. | In the NIST AI RMF Playbook guidance for Measure 2.2, particularly valuable action and documentation items for GPAI/foundation models include: <br>• *Follow human subjects research requirements as established by organizational and disciplinary requirements, including informed consent and compensation, during dataset collection activities.* <br>• *Follow intellectual property and privacy rights related to datasets and their use, including for the subjects represented in the data.* <br>• *Use informed consent for individuals providing data used in system testing and evaluation.* <br>• *How has the entity identified and mitigated potential impacts of bias in the data, including inequitable or discriminatory outcomes?* <br>• *To what extent are the established procedures effective in mitigating bias, inequity, and other concerns resulting from the system?* <br>• *If human subjects were used in the development or testing of the AI system, what protections were put in place to promote their safety and wellbeing?* <br><br>NIST GAI Profile additional guidance for Measure 2.2 includes: <br>• *Assess and manage statistical biases related to GAI content provenance through techniques such as re-sampling, re-weighting, or adversarial training.* <br>• *Provide human subjects with options to withdraw participation or revoke their consent for present or future use of their data in GAI applications.* <br>• *Use techniques such as anonymization, differential privacy or other privacy- enhancing technologies to minimize the risks associated with linking AI-generated content back to individual human subjects.* | NIST AI RMF Playbook (NIST 2023b) NIST Generative AI Profile, NIST AI 600-1 (Autio et al. 2024) |
| **Measure 2.3:** AI system performance or assurance criteria are measured qualitatively or quantitatively and demonstrated for conditions similar to deployment setting(s). Measures are documented. | In the NIST AI RMF Playbook guidance for Measure 2.3, particularly valuable action and documentation items for GPAI/foundation models include: <br>• *Conduct regular and sustained engagement with potentially impacted communities.* <br>• *Maintain a demographically diverse and multidisciplinary and collaborative internal team* <br>• *Evaluate feedback from stakeholder engagement activities, in collaboration with human factors and socio-technical experts.* <br>• *Measure AI systems prior to deployment in conditions similar to expected scenarios.* <br>• *What testing, if any, has the entity conducted on the AI system to identify errors and limitations (i.e. adversarial or stress testing)?* <br><br>In the NIST GAI Profile, particularly valuable additional actions for Measure 2.3 include: <br>• *Utilize a purpose-built testing environment such as NIST Dioptra to empirically evaluate GAI trustworthy characteristics.* <br><br>(See also guidance in this document for Govern 2.1 regarding roles for upstream developers as well as downstream developers and deployers, and see guidance under Measure 1.1 on approaches to measuring identified risks for GPAI/foundation models.) | NIST AI RMF Playbook (NIST 2023b) NIST Generative AI Profile, NIST AI 600-1 (Autio et al. 2024) <br><br>Dioptra 1.0.0 (Glasbrenner et al. 2024a,b) |

| Measure Category or Subcategory | Applicability and supplemental guidance for GPAI/foundation models | Resources |
|---|---|---|
| **Measure 2.4:** The functionality and behavior of the AI system and its components — as identified in the Map function — are monitored when in production. | In the NIST AI RMF Playbook guidance for Measure 2.4, particularly valuable action and documentation items for GPAI/foundation models include:<br>• *Monitor for anomalies using approaches such as control limits, confidence intervals, integrity constraints and ML algorithms. When anomalies are observed, consider error propagation and feedback loop risks.*<br>• *Collect uses cases from the operational environment for system testing and monitoring activities in accordance with organizational policies and regulatory or disciplinary requirements (e.g. informed consent, institutional review board approval, human research protections)*<br>• *How will the appropriate performance metrics, such as accuracy, of the AI be monitored after the AI is deployed?*<br><br>(See guidance in this document under Govern 2.1 regarding roles for upstream developers as well as downstream developers and deployers, and see guidance in this document under Measure 1.1 on approaches to measuring identified risks for GPAI/foundation models.) | NIST AI RMF Playbook (NIST 2023b) |
| **Measure 2.5:** The AI system to be deployed is demonstrated to be valid and reliable. Limitations of the generalizability beyond the conditions under which the technology was developed are documented. | In the NIST AI RMF Playbook guidance for Measure 2.5, particularly valuable action and documentation items for GPAI/foundation models include:<br>• *Establish or identify, and document approaches to measure forms of validity, including:*<br>  ○ *construct validity (the test is measuring the concept it claims to measure)*<br>  ○ *internal validity (relationship being tested is not influenced by other factors or variables)*<br>  ○ *external validity (results are generalizable beyond the training condition)*<br>  ○ *the use of experimental design principles and statistical analyses and modeling.*<br>• *Establish or identify, and document robustness measures.*<br>• *Establish or identify, and document reliability measures.*<br>• *Establish practices to specify and document the assumptions underlying measurement models to ensure proxies accurately reflect the concept being measured.*<br>• *What testing, if any, has the entity conducted on the AI system to identify errors and limitations (i.e. adversarial or stress testing)?*<br>• *To what extent are the established procedures effective in mitigating bias, inequity, and other concerns resulting from the system?*<br><br>In the NIST GAI Profile, particularly valuable additional actions for Measure 2.5 include:<br>• *Document the extent to which human domain knowledge is employed to improve GAI system performance, via, e.g., RLHF, fine-tuning, retrieval- augmented generation, content moderation, business rules.*<br>• *Review and verify sources and citations in GAI system outputs during pre-deployment risk measurement and ongoing monitoring activities.*<br>• *Track and document instances of anthropomorphization (e.g., human images, mentions of human feelings, cyborg imagery or motifs) in GAI system interfaces.*<br>• *Regularly review security and safety guardrails, especially if the GAI system is being operated in novel circumstances. This includes reviewing reasons why the GAI system was initially assessed as being safe to deploy.*<br><br>(See also guidance in this document for Govern 2.1 regarding roles for upstream developers as well as downstream developers and deployers, guidance in this document under Measure 1.1 on approaches to measuring identified risks for GPAI/foundation models, and guidance in this document under Map 1.3 and Map 5.1 for qualitative approaches to characterizing AI system objectives mis-specification or goal mis-generalization.) | For LLMs:<br>• DecodingTrust (Wang, Chen et al. 2023)<br>• TruthfulQA (Lin et al. 2021a,b)<br>• LAMBADA (Paperno et al. 2016)<br>• MMLU (Hendrycks, Burns et al. 2020)<br>• Winogender (Ruding-er et al. 2019)<br>• BIG-bench "pro-social behavior" category of benchmark tasks (BIG-bench n.d.b, BIG-bench collaboration 2021, Srivastava et al. 2022)<br>• Model-Written Evaluations "advanced-ai-risk," "sycophancy," and "winogender" datasets (Perez, Ringer et al. 2022a,b)<br>• WorldSense (Benchek-roun et al. 2023a,b)<br>• Do-Not-Answer (Wang, Li et al., 2023)<br>• Sociotechnical Safety Evaluations (Weiding-er et al. 2023a,b)<br><br>NIST AI RMF Playbook (NIST 2023b)<br>NIST Generative AI Profile, NIST AI 600-1 (Autio et al. 2024) |

| Measure Category or Subcategory | Applicability and supplemental guidance for GPAI/foundation models | Resources |
|---|---|---|
| **Measure 2.6:** The AI system is evaluated regularly for safety risks — as identified in the Map function. The AI system to be deployed is demonstrated to be safe, its residual negative risk does not exceed the risk tolerance, and it can fail safely, particularly if made to operate beyond its knowledge limits. Safety metrics reflect system reliability and robustness, real-time monitoring, and response times for AI system failures. | As part of safety evaluations of GPAI/foundation models:<br>• Perform red-teaming and adversarial testing of safety aspects of GPAI/foundation models; for frontier models, this testing should include dangerous-capability evaluations. (See also guidance in this document under Measure 1.1 on red-teaming and dangerous capability evaluations.)<br><br>In the NIST AI RMF Playbook guidance for Measure 2.6, particularly valuable action and documentation items for GPAI/foundation models include:<br>• *Thoroughly measure system performance in development and deployment contexts, and under stress conditions.*<br>  ○ *Employ test data assessments and simulations before proceeding to production testing. Track multiple performance quality and error metrics.*<br>  ○ *Stress-test system performance under likely scenarios (e.g., concept drift, high load) and beyond known limitations, in consultation with domain experts.*<br>  ○ *Test the system under conditions similar to those related to past known incidents or near-misses and measure system performance and safety characteristics.*<br>• *Measure and monitor system performance in real-time to enable rapid response when AI system incidents are detected.*<br>• *Document, practice and measure incident response plans for AI system incidents, including measuring response and down times.*<br>• *What testing, if any, has the entity conducted on the AI system to identify errors and limitations (i.e. adversarial or stress testing)?*<br>• *To what extent has the entity documented the AI system's development, testing methodology, metrics, and performance outcomes?*<br>• *Did you establish mechanisms that facilitate the AI system's auditability (e.g. traceability of the development process, the sourcing of training data and the logging of the AI system's processes, outcomes, positive and negative impact)?*<br>  ○ For some GPAI/foundation models (e.g., using models run on central servers accessed through APIs), these can include data mining of usage metrics, audit logs, etc. as appropriate to identify anomalous conditions that users encounter but might not report.<br>• *Did you ensure that the AI system can be audited by independent third parties?*<br>• *Did you establish a process for third parties (e.g. suppliers, end-users, subjects, distributors/vendors or workers) to report potential vulnerabilities, risks or biases in the AI system?*<br><br>In the NIST GAI Profile, particularly valuable additional actions for Measure 2.6 include:<br>• *Assess adverse impacts, including health and wellbeing impacts for value chain or other AI Actors that are exposed to sexually explicit, offensive, or violent information during GAI training and maintenance.*<br>• *Assess existence or levels of harmful bias, intellectual property infringement, data privacy violations, obscenity, extremism, violence, or CBRN information in system training data.*<br>• *Re-evaluate safety features of fine-tuned models when the negative risk exceeds organizational risk tolerance.*<br>• *Review GAI system outputs for validity and safety: Review generated code to assess risks that may arise from unreliable downstream decision-making.*<br>• *Verify that GAI system architecture can monitor outputs and performance, and handle, recover from, and repair errors when security anomalies, threats and impacts are detected.* | For red-teaming and dangerous capability evaluation of frontier models:<br>• OpenAI (2023a, pp. 15–16) and ARC Evals (2023a,b)<br>• Kinniment et al. (2023)<br>• Shevlane et al. (2023)<br>• Mouton et al. (2024)<br>• WMDP (Li, Pan et al. 2024a,b,c)<br><br>For red-teaming LLMs and toxicity:<br>• Casper et al. (2023a,b,c)<br><br>For LLM truthfulness and toxicity:<br>• ToxiGen (Hartvigsen et al. 2022)<br>• TruthfulQA (Lin et al. 2021a,b)<br>• MACHIAVELLI (Pan et al. 2023)<br>• Do-Not-Answer (Wang, Li et al., 2023)<br><br>AIID (n.d.)<br>ATLAS AI Incidents (MITRE n.d.b)<br>NIST AI RMF Playbook (NIST 2023b)<br>NIST Generative AI Profile, NIST AI 600-1 (Autio et al. 2024)<br>NIST AI 800-1 ipd (NIST 2024b) |

| Measure Category or Subcategory | Applicability and supplemental guidance for GPAI/foundation models | Resources |
|---|---|---|
| Measure 2.6, continued | • *Verify that systems properly handle queries that may give rise to inappropriate, malicious, or illegal usage, including facilitating manipulation, extortion, targeted impersonation, cyber-attacks, and weapons creation.*<br>• *Regularly evaluate GAI system vulnerabilities to possible circumvention of safety measures.* | |
| **Measure 2.7:** AI system security and resilience — as identified in the Map function — are evaluated and documented. | Use information security measures to assess and assure model weight security (specifically, integrity and confidentiality) as part of preventing misuse or abuse of models. This is particularly valuable for frontier models for which public release of model weights could enable misuse with particularly high-consequence impacts.<br>• Anthropic's frontier-model security practices include requirements for multi-party authorization for access to frontier model development and deployment systems, and secure development and supply chain practices, including chain of custody (Anthropic 2023a).<br>• Several frontier-model developers have committed to investing in cybersecurity and insider-threat controls for a high level of protection of proprietary and unreleased frontier-model weights. "This includes limiting access to model weights to those whose job function requires it and establishing a robust insider threat detection program consistent with protections provided for their most valuable intellectual property and trade secrets. In addition, it requires storing and working with the weights in an appropriately secure environment to reduce the risk of unsanctioned release" (White House 2023a, p.3).<br>• Newer work from RAND (Nevo et al. 2024) outlines relatively comprehensive recommendations for protecting frontier model weights across five security levels (SLs). Some of the recommendations correspond to commonly used standards, e.g., they mention NIST SP 800-171 or an equivalent as part of SL3, and high-impact system standards as part of SL4. Some items would require additional research and development, especially for the highest security level, SL5.<br><br>As a general guideline for information system **security expectations for protecting the integrity and confidentiality of proprietary or unreleased foundation model parameter weights**, foundation model developers should implement the NIST Cybersecurity Framework (NIST 2024a), or an approximate equivalent such as NIST SP 800-171 or ISO/IEC 27001, with at least the following security controls or approximate equivalents:[33]<br>• For frontier models: High-value asset guidance (e.g., per NIST SP 800-171 and NIST SP 800-172), or high-impact system baseline per NIST SP 800-53B as an informative reference for the NIST Cybersecurity Framework, or approximate equivalent.<br>• For other foundation models: Moderate-impact system baseline guidance (e.g., per NIST SP 800-171), or moderate-impact system baseline per NIST SP 800-53B as an informative reference for the NIST Cybersecurity Framework, or approximate equivalent. | NIST AI RMF Playbook (NIST 2023b)<br>NIST Generative AI Profile, NIST AI 600-1 (Autio et al. 2024)<br><br>On baseline expectations for information system security for foundation model developers:<br>• NIST Cybersecurity Framework (NIST 2024a)<br>• NIST SP 800-53 (NIST 2023d) including SC-28<br>• NIST SP 800-53B (NIST 2020a)<br>• NIST SP 800-171 (NIST 2020b)<br>• NIST SP 800-172 (NIST 2021)<br>• ISO/IEC (2022)<br>• Anthropic (2023a, 2024a)<br>• Nevo et al. (2024)<br>• NIST SP 800-218A (Booth et al. 2024)<br>• ACSC (2024)<br><br>On security vulnerabilities and mitigations for LLMs and other types of ML models:<br>• NIST AI 100-2e2023 (Vassilev et al. 2024)<br>• ENISA (2021, 2023) |

---

33    For approximate equivalents, see, e.g., the NIST (2020b) mappings of controls between NIST SP 800-171 and NIST 800-53 and ISO/IEC 27001; the NIST (2021) mapping of controls between NIST SP 800-172 and NIST SP 800-53; the NIST (2020c) mappings of controls between the NIST Cybersecurity Framework and NIST SP 800-53; the NIST (2023d) mapping of controls between NIST SP 800-53 and ISO/IEC 27001, and the CIS (n.d.) mapping of controls between NIST SP 800-53 and CIS Critical Security Controls. See also the RAND (Nevo et al. 2024) mapping of five security levels (SLs) for protection of frontier model weights to the NIST Cybersecurity Framework; e.g., they mention NIST SP 800-171 or an equivalent as part of SL3, and high-impact system standards as part of SL4.

| Measure Category or Subcategory | Applicability and supplemental guidance for GPAI/foundation models | Resources |
|---|---|---|
| Measure 2.7, continued | As part of security evaluations of GPAI/foundation models:<br>• Perform red-teaming and adversarial testing of security aspects of GPAI/foundation models. (See also guidance in this document under Measure 1.1 on red-teaming and adversarial testing.)<br>• Check for backdoors, AI trojans, prompt injection vulnerabilities, etc. during testing/evaluation, especially for models trained on untrusted data from public sources with susceptibility to data poisoning. Tools to consider using include TrojAI (Karra et al. 2020, NIST n.d.a); see also NIST AI 100-2e2023 (Vassilev et al. 2024).<br>    ◦ Even after a vulnerability is discovered in a GPAI/foundation model, how to fix it may not always be clear or tractable. For example, Anthropic researchers were unable to successfully remove backdoors in LLMs using standard safety training techniques (Hubinger et al. 2024).<br>• Engage in continuous monitoring, vulnerability disclosure, and bug bounty programs for GPAI/foundation models to identify novel security vulnerabilities.<br>• Track uncovered security vulnerabilities in other GPAI/foundation models, including open source foundation models, which may be transferable to other models (See, e.g., Zou et al. 2023a,b).<br><br>In the NIST AI RMF Playbook guidance for Measure 2.7, particularly valuable action and documentation items for GPAI/foundation models include:<br>• *Establish and track AI system security tests and metrics (e.g., red-teaming activities, frequency and rate of anomalous events, system down-time, incident response times, time-to-bypass, etc.).*<br>• *Use red-team exercises to actively test the system under adversarial or stress conditions, measure system response, assess failure modes or determine if system can return to normal function after an unexpected adverse event.*<br>• *Document red-team exercise results as part of continuous improvement efforts, including the range of security test conditions and results.*<br>• *Verify that information about errors and attack patterns is shared with incident databases, other organizations with similar systems, and system users and stakeholders (see also related guidance under Manage 4.1).*<br>• *Develop and maintain information sharing practices with AI actors from other organizations to learn from common attacks.*<br>• *Verify that third party AI resources and personnel undergo security audits and screenings. Risk indicators may include failure of third parties to provide relevant security information.*<br>• *Utilize watermarking technologies as a deterrent to data and model extraction attacks.*<br><br>In the NIST GAI Profile, particularly valuable additional actions for Measure 2.7 include:<br>• *Apply established security measures to: Assess likelihood and magnitude of vulnerabilities and threats such as backdoors, compromised dependencies, data breaches, eavesdropping, man-in-the-middle attacks, reverse engineering, autonomous agents, model theft or exposure of model weights, AI inference, bypass, extraction, and other baseline security concerns.*<br>• *Measure reliability of content authentication methods, such as watermarking, cryptographic signatures, digital fingerprints, as well as access controls, conformity assessment, and model integrity verification, which can help support the effective implementation of content provenance techniques. Evaluate the rate of false positives and false negatives in content provenance, as well as true positives and true negatives for verification.* | • OWASP (2023a,b)<br>• Barrett, Boyd et al. (2023)<br>• ATLAS (MITRE n.d.a)<br>• TrojAI (Karra et al. 2020, NIST n.d.a)<br><br>For a range of LLM red-teaming approaches with security implications:<br>• Ganguli, Lovitt et al. (2022)<br>• Casper et al. (2023a,b,c)<br>• Zou et al. (2023a,b)<br>• OpenAI (2023a, pp. 15–16) and ARC Evals (2023a,b)<br>• Kinniment et al. (2023)<br>• Anthropic (2023a,b)<br>• Shevlane et al. (2023) |

| Measure Category or Subcategory | Applicability and supplemental guidance for GPAI/foundation models | Resources |
|---|---|---|
| **Measure 2.7, continued** | • *Perform AI red-teaming to assess resilience against: Abuse to facilitate attacks on other systems (e.g., malicious code generation, enhanced phishing content), GAI attacks (e.g., prompt injection), ML attacks (e.g., adversarial examples/prompts, data poisoning, membership inference, model extraction, sponge examples).*<br>• *Verify fine-tuning does not compromise safety and security controls.*<br>• *Regularly assess and verify that security measures remain effective and have not been compromised.* | |
| **Measure 2.8:** Risks associated with transparency and accountability — as identified in the Map function — are examined and documented. | In the NIST AI RMF Playbook guidance for Measure 2.8, particularly valuable action and documentation items for GPAI/foundation models include:<br>• *Instrument the system for measurement and tracking, e.g., by maintaining histories, audit logs and other information that can be used by AI actors to review and evaluate possible sources of error, bias, or vulnerability.*<br>• *Track, document, and measure organizational accountability regarding AI systems via policy exceptions and escalations, and document "go" and "no/go" decisions made by accountable parties.*<br>• *Track and audit the effectiveness of organizational mechanisms related to AI risk management, including:*<br>  ◦ *Lines of communication between AI actors, executive leadership, users and impacted communities.*<br>  ◦ *Roles and responsibilities for AI actors and executive leadership.*<br>  ◦ *Organizational accountability roles, e.g., chief model risk officers, AI oversight committees, responsible or ethical AI directors, etc.*<br><br>In the NIST GAI Profile, particularly valuable additional actions for Measure 2.8 include:<br>• *Compile statistics on actual policy violations, take-down requests, and intellectual property infringement for organizational GAI systems: Analyze transparency reports across demographic groups, languages groups.*<br>• *Document the instructions given to data annotators or AI red-teamers.*<br>• *Use digital content transparency solutions to enable the documentation of each instance where content is generated, modified, or shared to provide a tamper-proof history of the content, promote transparency, and enable traceability.*<br><br>Document organizational transparency and disclosure mechanisms to inform users or allow users to check whether they are interacting with, or observing content created by, a generative AI system. See, e.g., Partnership on AI's Responsible Practices for Synthetic Media, and Synthetic Media Indirect Disclosure (PAI 2023a,d), as well as CAI (2023) and C2PA (2023).<br><br>(See also guidance in this document under Govern 2.1 on roles for upstream and downstream developers, and under Manage 1.3 on transparency and disclosure.) | PAI (2023a)<br>PAI (2023d)<br>CAI (2023)<br>C2PA (2023)<br>Solaiman (2023)<br>NIST AI RMF Playbook (NIST 2023b)<br>NIST Generative AI Profile, NIST AI 600-1 (Autio et al. 2024)<br><br>For data audits and copyright filtering:<br>C4 (Dodge et al. 2021), see also (Birhane et al. 2021) |

| Measure Category or Subcategory | Applicability and supplemental guidance for GPAI/foundation models | Resources |
|---|---|---|
| **Measure 2.9:** The AI model is explained, validated, and documented, and AI system output is interpreted within its context — as identified in the Map function — to inform responsible use and governance. | It is critical to ensure that users know how to interpret system behavior and outputs, including the limitations of both the system and any explanations provided. However, explainability and interpretability are often extremely limited for LLMs and other GPAI/ foundation models with deep-learning architectures. These systems can be inappropriate for applications requiring a higher level of explainability and interpretability.<br><br>For some increasingly capable GPAI/foundation models, the reliability of some techniques (such as RLHF) for aligning model behavior with human values or intentions could be improved by integrating sufficient interpretability methods to prevent "deceptive alignment" (Hubinger et al. 2019, Ngo, Chan et al. 2022).<br>• While interpretability techniques are not yet sufficient to assess risks such as hidden failures of RLHF for GPAI/foundation model alignment, developers of GPAI/founda-tion models (especially frontier models) should include such risks in a risk register or other tool for tracking identified risks that are difficult to assess. (See related guidance in this document under Measure 3.2.)<br><br>In the NIST AI RMF Playbook guidance for Measure 2.9, particularly valuable action and documentation items for GPAI/foundation models include:<br>• *What type of information is accessible on the design, operations, and limitations of the AI system to external stakeholders, including end users, consumers, regulators, and individuals impacted by use of the AI system?*<br><br>In the NIST GAI Profile, particularly valuable additional actions for Measure 2.9 include:<br>• *Apply and document ML explanation results (e.g., analysis of embeddings, gra-dient-based attributions, model compression/surrogate models, occlusion/term reduction)*<br>• *Document GAI model details including: Proposed use and organizational value; Assumptions and limitations, Data collection methodologies; Data provenance; Data quality; Model architecture; Optimization objectives; Training algorithms; RLHF approaches; Fine-tuning or retrieval-augmented generation approaches; Evaluation data; Ethical considerations; Legal and regulatory requirements.*<br><br>(See also guidance in this document for Govern 2.1 regarding roles for upstream developers as well as downstream developers and deployers, and see guidance in this document under Measure 1.1 on approaches to measuring identified risks for GPAI/ foundation models.) | Mitchell et al. (2019) NIST AI RMF Playbook (NIST 2023b) NIST Generative AI Profile, NIST AI 600-1 (Autio et al. 2024)<br><br>For transparency tools: System cards (Open AI 2023a) FactSheets (Arnold et al. 2019) Data statements (Bender and Friedman 2018) DataSheets (Gebru et al. 2021) Model Cards (Liang et al. 2024) |
| **Measure 2.10:** Privacy risk of the AI system — as identified in the Map function — is examined and documented. | Privacy challenges for GPAI/foundation models include the issue that, after pre-training on large quantities of uncurated web-scraped data or other sources, personally sensitive material in the training data can be revealed by user prompts.<br><br>Logs and histories of engagement with GPAI/foundation models may also include highly sensitive or personal information, which could be susceptible to breaches or leaks.<br><br>In the NIST AI RMF Playbook guidance for Measure 2.10, particularly valuable action and documentation items for GPAI/foundation models include:<br>• *Document collection, use, management, and disclosure of personally sensitive infor-mation in datasets, in accordance with privacy and data governance policies.*<br>• *Establish and document protocols (authorization, duration, type) and access con-trols for training sets or production data containing personally sensitive information, in accordance with privacy and data governance policies.* | NIST AI RMF Playbook (NIST 2023b) NIST Generative AI Profile, NIST AI 600-1 (Autio et al. 2024) |

| Measure Category or Subcategory | Applicability and supplemental guidance for GPAI/foundation models | Resources |
|---|---|---|
| Measure 2.10, continued | • *Monitor internal queries to production data for detecting patterns that isolate personal records.*<br>• *Did your organization implement accountability-based practices in data management and protection (e.g. the PDPA and OECD Privacy Principles)?*<br>• *What assessments has the entity conducted on data security and privacy impacts associated with the AI system?*<br><br>Additional valuable steps to consider include:<br>• Enable people to consent to  and/or opt out of the uses of their data.<br>• Notify users and impacted communities about privacy or security breaches.<br><br>In the NIST GAI Profile, particularly valuable additional actions for Measure 2.10 include:<br>• *Conduct AI red-teaming to assess issues such as: Outputting of training data samples, and subsequent reverse engineering, model extraction, and membership inference risks; Revealing biometric, confidential, copyrighted, licensed, patented, personal, proprietary, sensitive, or trade-marked information; Tracking or revealing location information of users or members of training datasets.*<br><br>(See also guidance in this document for Govern 2.1 regarding roles for upstream developers as well as downstream developers and deployers, and see guidance in this document under Measure 1.1 on approaches to measuring identified risks for GPAI/foundation models.) | |
| **Measure 2.11:** Fairness and bias — as identified in the Map function — are evaluated and results are documented. | There are numerous challenges relating to fairness and bias, including closely related issues such as stereotypes, representational harms, inequities, and cultural homogeneity, that are relatively unique to GPAI/foundation models.<br><br>Training datasets frequently embed and amplify harmful biases in resulting models. Given the vast size of the training datasets typically required for GPAI/foundation models, it can be especially hard for developers to know or mitigate all of the harmful biases that are present. Biases are also introduced by the choices made about modeling, optimization, hardware, and testing.<br><br>Evaluating for fairness and bias in GPAI/foundation models should take into account this complexity and should not, for example, focus only on narrow definitions of protected classes, which may overlook complexities of identity (Solaiman et al. 2023). These complexities can include the intersectionality of certain protected classes. Evaluating and mitigating harmful biases in the context of one protected class at a time can lead to overlooking population subgroups (e.g., Black women) that have been historically neglected (Buolamwini and Gebru 2018).<br><br>The training data may also contain redundant encodings[34] that act as proxies for identifying protected attributes, resulting in little to no decrease in biased outcomes when protected attributes are removed, often resulting in more harm to the protected group (Cheng et al. 2023).<br><br>In the NIST AI RMF Playbook guidance for Measure 2.11, particularly valuable action and documentation items for GPAI/foundation models include:<br>• *Understand and consider sources of bias in training and TEVV data:*<br>   ◦ *Differences in distributions of outcomes across and within groups, including intersecting groups.* | For LLMs:<br>• BBQ (Parrish et al. 2021a,b)<br>• Winogender Schemas (Rudinger et al, 2019)<br>• ToxiGen (Hartvigsen et al. 2022)<br>• TruthfulQA (Lin et al. 2021a,b)<br>• BOLD (Dhamala 2021a,b)<br>• Su et al. (2023)<br><br>Aequitas (Saleiro et al. 2019)<br>AIFairness 360 (Bellamy et al. 2018)<br>Fairlearn (Fairlearn Contributors 2023)<br><br>NIST SP 1270 (Schwartz et al. 2022)<br>NIST AI RMF Playbook (NIST 2023b)<br>NIST Generative AI Profile, NIST AI 600-1 (Autio et al. 2024) |

34    For example, a dataset may not explicitly include a job applicant's gender, but other data can be used to infer the applicants gender.

| Measure Category or Subcategory | Applicability and supplemental guidance for GPAI/foundation models | Resources |
|---|---|---|
| Measure 2.11, continued |     ◦  *Completeness, representativeness and balance of data sources.*<br>    ◦  *Identify input data features that may serve as proxies for demographic group membership (i.e., credit score, ZIP code) or otherwise give rise to emergent bias within AI systems.*<br>    ◦  *Forms of systemic bias in images, text (or word embeddings), audio or other complex or unstructured data.*<br>• *Leverage impact assessments to identify and classify system impacts and harms to end users, other individuals, and groups with input from potentially impacted communities.*<br>• *Identify the classes of individuals, groups, or environmental ecosystems which might be impacted through direct engagement with potentially impacted communities.*<br>• *Collect and share information about differences in outcomes for the identified groups.*<br>• *How has the entity identified and mitigated potential impacts of bias in the data, including inequitable or discriminatory outcomes?*<br><br>Additional valuable steps include:<br>• Review AI system development and uses for potential threats to human rights, dignity, or wellbeing.<br>• Ensure the AI system's user interface is usable by those with special needs or disabilities, or those at risk of exclusion.<br>• Determine methods to distribute the benefits of the system widely and equitably.<br><br>In the NIST GAI Profile, particularly valuable additional actions for Measure 2.11 include:<br>• *Apply use-case appropriate benchmarks (e.g., Bias Benchmark Questions, Real Hateful or Harmful Prompts, Winogender Schemas) to quantify systemic bias, stereotyping, denigration, and hateful content in GAI system outputs; Document assumptions and limitations of benchmarks, including any actual or possible training/test data cross contamination, relative to in-context deployment environment.*<br>• *Conduct fairness assessments to measure systemic bias. Measure GAI system performance across demographic groups and subgroups, addressing both quality of service and any allocation of services and resources. Quantify harms using: field testing with sub-group populations to determine likelihood of exposure to generated content exhibiting harmful bias, AI red-teaming with counterfactual and low-context (e.g., "leader," "bad guys") prompts.*<br>• *Review, document, and measure sources of bias in GAI training and TEVV data: Differences in distributions of outcomes across and within groups, including intersecting groups; Completeness, representativeness, and balance of data sources; demographic group and subgroup coverage in GAI system training data; Forms of latent systemic bias in images, text, audio, embeddings, or other complex or unstructured data; Input data features that may serve as proxies for demographic group membership (i.e., image metadata, language dialect) or otherwise give rise to emergent bias within GAI systems; The extent to which the digital divide may negatively impact representativeness in GAI system training and TEVV data; Filtering of hate speech or content in GAI system training data; Prevalence of GAI-generated data in GAI system training data.*<br><br>(See also guidance in this document under Map 5.1 on identifying potential large-scale harms from correlated bias across large numbers of people or a large fraction of a group or a society's population.) | Algorithmic Pluralism (Jain et al. 2023)<br>Data Statements (Bender and Friedman 2018) |

| Measure Category or Subcategory | Applicability and supplemental guidance for GPAI/foundation models | Resources |
|---|---|---|
| **Measure 2.12:** Environmental impact and sustainability of AI model training and management activities — as identified in the Map function — are assessed and documented. | Environmental-impact assessment by GPAI/foundation model developers should include estimating the environmental impact of large-scale ML model training.<br>• Relevant tools, resources and examples include ML CO2 Impact (Schmidt et al. 2019), Lacoste et al. (2019), OECD (2022b), Rafat et al. (2023), and Luccioni et al. (2022).<br>• Assessment of environmental impacts is particularly important for LLMs and other large-scale ML-based AI systems, which typically have much larger model-training environmental impacts than smaller-scale ML models (Bender et al. 2021).<br><br>In the NIST GAI Profile, particularly valuable additional actions for Measure 2.12 include:<br>• *Assess safety to physical environments when deploying GAI systems.*<br>• *Measure or estimate environmental impacts (e.g., energy and water consumption) for training, fine tuning, and deploying models: Verify tradeoffs between resources used at inference time versus additional resources required at training time.* | Schmidt et al. (2019) Lacoste et al. (2019) OECD (2022b) Rafat et al. (2023) Luccioni et al. (2022)<br><br>NIST AI RMF Playbook (NIST 2023b) NIST Generative AI Profile, NIST AI 600-1 (Autio et al. 2024) |
| **Measure 2.13:** Effectiveness of the employed TEVV metrics and processes in the Measure function are evaluated and documented. | In the NIST AI RMF Playbook guidance for Measure 2.13, particularly valuable action and documentation items for GPAI/foundation models include:<br>• *Assess effectiveness of metrics for identifying and measuring risks.*<br><br>In the NIST GAI Profile, particularly valuable additional actions for Measure 2.13 include:<br>• *Create measurement error models for pre-deployment metrics to demonstrate construct validity for each metric (i.e., does the metric effectively operationalize the desired concept):*<br>  ◦ *Measure or estimate, and document, biases or statistical variance in applied metrics or structured human feedback processes;*<br>  ◦ *Leverage domain expertise when modeling complex societal constructs such as hateful content.* | NIST AI RMF Playbook (NIST 2023b) NIST Generative AI Profile, NIST AI 600-1 (Autio et al. 2024) |
| **Measure 3:  Mechanisms for tracking identified AI risks over time are in place.** | | |
| **Measure 3.1:** Approaches, personnel, and documentation are in place to regularly identify and track existing, unanticipated, and emergent AI risks based on factors such as intended and actual performance in deployed contexts. | Valuable steps to consider include:<br>• Consider taking steps to identify or assess longer-term impacts, or use longer time horizons (longer than would be typical for smaller-scale, fixed-purpose AI systems), and thereby reduce the potential for surprise.<br>• Consider whether any risk assessment or impact assessment answers would change when assessing longer-term time periods (e.g., beyond the next year).<br>  ◦ If your AI system is deployed for a long period of time, then:<br>    » What additional impacts would you expect?<br>    » Which impacts would you expect to have greater magnitude?<br>• Identify unintended potential future events that should trigger reassessment or other responses, and build them into risk registers and/or planning of relevant lifecycle stages. (These can be particularly important for foundation models, which often have emergent capabilities and other emergent properties that are not identified in earlier-stage testing.) To identify trigger events, consider questions such as:<br>  ◦ What if monitoring indicates that one of your risk-mitigation controls is not working as expected? (Consider this, as applicable, for each relevant risk-mitigation control.)<br>  ◦ What if AI capability developments occur that are not expected until further into the future, such as availability of much more powerful AI systems or computing resources to train and run AI systems, or demonstration of new emergent capabilities (e.g., via new prompts) that were not identified in earlier-stage testing? | AIID (n.d.) ATLAS AI Incidents (MITRE n.d.b) Section 3.2 of Barrett et al. (2022) NIST AI RMF Playbook (NIST 2023b) NIST AI 800-1 ipd (NIST 2024b, Objective 7)<br><br>For model documentation: Model cards (Liang et al. 2024) FactSheets (Arnold et al. 2019) Datasheets for datasets (Gebru et al. 2021)<br><br>On challenges of documentation: Winecoff and Bogen (2024) |

| Measure Category or Subcategory | Applicability and supplemental guidance for GPAI/foundation models | Resources |
|---|---|---|
| **Measure 3.1, continued** | ○ What if a near-miss incident occurs in a critical system or process? Does your organization have procedures for near-miss incident identification, analysis, tracking, and information sharing? Does your organization also monitor the AIID or other sources for near-miss incident reports on other organizations' systems?<br><br>Recommendations for documentation that can help provide transparency about how various practices are implemented can be found in the NIST Managing Misuse Risk for Dual-Use Foundation Models ipd (NIST 2024b).<br><br>In the NIST AI RMF Playbook guidance for Measure 3.1, particularly valuable action and documentation items for GPAI/foundation models include:<br>• *Assess effectiveness of metrics for identifying and measuring emergent risks.*<br>• *To what extent can users or parties affected by the outputs of the AI system test the AI system and provide feedback?* | |
| **Measure 3.2:** Risk tracking approaches are considered for settings where AI risks are difficult to assess using currently available measurement techniques or where metrics are not yet available. | **Use appropriate mechanisms for tracking identified risks, even if only characterizing them qualitatively and even if the risks are difficult to assess.** This is particularly important for foundation models, because of their potential scale of impact, and their potential for emergent properties or other novel risks.<br>• Consider tracking identified risks (including difficult-to-assess risks) using a risk register. (For more on risk registers, see, e.g., ISO Guide 73 Section 3.8.2.4, PMI 2017 p. 417, Stine et al. 2020, MIT 2024, and Slattery et al. 2024.)<br>• When developing frontier models with unprecedented capabilities, failure modes, and other emergent properties, it is especially valuable to use red-teams and adversarial testing prior to deployment. See related guidance in this document under Measure 1.1.<br>• **Risk tracking should include ongoing monitoring of newly identified capabilities and limitations of deployed GPAI/foundation models.** These efforts can include monitoring use of the models through APIs, and monitoring publications or online forums that discuss new uses of the models. "If significant information on model capabilities is discovered post-deployment, risk assessments should be repeated, and deployment safeguards updated" (Anderljung, Barnhart et al. 2023).<br><br>In the NIST AI RMF Playbook guidance for Measure 3.2, particularly valuable action and documentation items for GPAIS include:<br>• *Establish processes for tracking emergent risks that may not be measurable with current approaches. Some processes may include:*<br>  ○ *Recourse mechanisms for faulty AI system outputs.*<br>  ○ *Bug bounties.*<br>  ○ *Human-centered design approaches.*<br>  ○ *User-interaction and experience research.*<br>  ○ *Participatory stakeholder engagement with affected or potentially impacted individuals and communities.*<br>• *Determine and document the rate of occurrence and severity level for complex or difficult-to-measure risks when:*<br>  ○ *Prioritizing new measurement approaches for deployment tasks.*<br>  ○ *Allocating AI system risk management resources.*<br>  ○ *Evaluating AI system improvements.*<br>  ○ *Making go/no-go decisions for subsequent system iterations.* | Section 3.2 of Barrett et al. (2022)<br><br>NIST AI RMF Playbook (NIST 2023b)<br>NIST Generative AI Profile, NIST AI 600-1 (Autio et al. 2024)<br>NIST AI 800-1 ipd (NIST 2024b, Objectives 1 and 4)<br><br>On bug bounties and bias bounties:<br>Globus-Harris et al. (2022)<br>Kenway et al. (2022)<br>OpenAI (2023c)<br><br>AI risk repository MIT (2024)<br>Slattery et al. (2024) |

| Measure Category or Subcategory | Applicability and supplemental guidance for GPAI/foundation models | Resources |
|---|---|---|
| **Measure 3.3:** Feedback processes for end users and impacted communities to report problems and appeal system outcomes are established and integrated into AI system evaluation metrics. | In the NIST AI RMF Playbook guidance for Measure 3.3, particularly valuable action and documentation items for GPAI/foundation models include: <br>• *To what extent can users or parties affected by the outputs of the AI system test the AI system and provide feedback?* <br>• *How easily accessible and current is the information available to external stakeholders?* <br>• *What type of information is accessible on the design, operations, and limitations of the AI system to external stakeholders, including end users, consumers, regulators, and individuals impacted by use of the AI system?* <br><br>In the NIST GAI Profile, particularly valuable additional actions for Measure 3.3 include: <br>• *Conduct studies to understand how end users perceive and interact with GAI content and accompanying content provenance within context of use. Assess whether the content aligns with their expectations and how they may act upon the information presented.* <br>• *Provide input for training materials about the capabilities and limitations of GAI systems related to digital content transparency for AI Actors, other professionals, and the public about the societal impacts of AI and the role of diverse and inclusive content generation.* | NIST AI RMF Playbook (NIST 2023b) <br>NIST Generative AI Profile, NIST 600-1 (Autio et al. 2024) |
| **Measure 4: Feedback about efficacy of measurement is gathered and assessed.** | | |
| **Measure 4.1:** Measurement approaches for identifying AI risks are connected to deployment context(s) and informed through consultation with domain experts and other end users. Approaches are documented. | For GPAI/foundation model developers, model "users" include downstream developers as well as the end users of applications built on GPAI/foundation models. Downstream developers typically have the most direct interactions with end users in particular deployment contexts. However, it can be valuable for upstream GPAI/foundation model developers to provide mechanisms for feedback from end users or other AI actors, as well as from downstream developers. <br><br>(See also guidance under Govern 2.1 regarding roles for GPAI/foundation model developers, e.g., on performing testing during model development or other testing that requires direct access to the system, as well as roles for downstream developers and deployers, e.g., on performing testing of end-use applications built on a GPAI/foundation model and testing appropriate for that application context.) | NIST AI RMF Playbook (NIST 2023b) |
| **Measure 4.2:** Measurement results regarding AI system trustworthiness in deployment context(s) and across the AI lifecycle are informed by input from domain experts and relevant AI actors to validate whether the system is performing consistently as intended. Results are documented. | When considering what types of domain experts to use in reviewing information on identified risks, consider including personnel recommended for risk identification, such as social scientists for perspective on structural or systemic risks, per guidance in this document under Govern 3.1. <br><br>In the NIST AI RMF Playbook guidance for Measure 4.2, particularly valuable action and documentation items for GPAI/foundation models include: <br>• *Integrate feedback from end users, operators, and affected individuals and communities from Map function as inputs to assess AI system trustworthiness characteristics. Ensure both positive and negative feedback is being assessed.* <br>• *Evaluate feedback in connection with AI system trustworthiness characteristics from Measure 2.5 to 2.11.* <br>• *Consult AI actors in impact assessment, human factors and socio-technical tasks to assist with analysis and interpretation of results.* | NIST AI RMF Playbook (NIST 2023b) <br>NIST Generative AI Profile, NIST 600-1 (Autio et al. 2024) |

| Measure Category or Subcategory | Applicability and supplemental guidance for GPAI/foundation models | Resources |
|---|---|---|
| Measure 4.2, continued | In the NIST GAI Profile, particularly valuable additional actions for Measure 4.2 include:<br>• *Conduct adversarial testing at a regular cadence to map and measure GAI risks, including tests to address attempts to deceive or manipulate the application of provenance techniques or other misuses. Identify vulnerabilities and understand potential misuse scenarios and unintended outputs.*<br>• *Evaluate GAI system performance in real-world scenarios to observe its behavior in practical environments and reveal issues that might not surface in controlled and optimized testing environments.*<br>• *Implement interpretability and explainability methods to evaluate GAI system decisions and verify alignment with intended purpose.*<br>• *Monitor and document instances where human operators or other systems override the GAI's decisions. Evaluate these cases to understand if the overrides are linked to issues related to content provenance.*<br>• *Verify and document the incorporation of results of structured public feedback exercises into design, implementation, deployment approval ("go"/"no-go" decisions), monitoring, and decommission decisions.*<br><br>(See also guidance in this document under Govern 2.1 regarding roles for GPAI/foundation model developers, e.g., on performing testing during model development or other testing that requires direct access to the system, as well as roles for downstream developers and deployers, e.g., on performing testing of end-use applications built on a GPAI/foundation model and testing appropriate for that application context.) | |
| Measure 4.3: Measurable performance improvements or declines based on consultations with relevant AI actors, including affected communities, and field data about context-relevant risks and trustworthiness characteristics are identified and documented. | In the NIST AI RMF Playbook guidance for Measure 4.3, particularly valuable action and documentation items for GPAI/foundation models include:<br>• *Develop baseline quantitative measures for trustworthy characteristics.*<br>• *Delimit and characterize baseline operation values and states.*<br>• *Utilize qualitative approaches to augment and complement quantitative baseline measures, in close coordination with impact assessment, human factors and socio-technical AI actors.*<br>• *Monitor and assess measurements as part of continual improvement to identify potential system adjustments or modifications.*<br><br>(See also guidance in this document under Govern 2.1 regarding roles for GPAI/foundation model developers, e.g., on performing testing during GPAI/foundation model development or other testing that requires direct access to the system, as well as roles for downstream developers and deployers, e.g., on performing testing of end-use applications built on a GPAI/foundation model and testing appropriate for that application context.) | NIST AI RMF Playbook (NIST 2023b) |

### 3.4 GUIDANCE FOR NIST AI RMF MANAGE SUBCATEGORIES

#### Table 4: Guidance for NIST AI RMF Manage Subcategories

| Manage Category or Subcategory | Applicability and supplemental guidance for GPAI/foundation models | Resources |
|---|---|---|
| **Manage 1:  AI risks based on assessments and other analytical output from the Map and Measure functions are prioritized, responded to, and managed.** | | |
| **Manage 1.1:** A determination is made as to whether the AI system achieves its intended purposes and stated objectives and whether its development or deployment should proceed. | When considering the "intended purpose" of a GPAI/foundation model, in addition to any originally intended use cases, include consideration of other *potential* use cases; see related guidance in this document under Map 1.1. This is particularly important for GPAI/ foundation models, which can have large numbers of uses.<br><br>**When making go/no-go decisions, especially on whether to proceed on major stages or investments for development or deployment of cutting-edge large-scale GPAI/foundation models:**<br>• See guidance in this document under Map 1.3 on AI development objectives, especially: **Consider potential for mis-specified AI system objectives, and consider what kinds of perverse behavior could be incentivized by optimizing for those objectives.**<br>• See guidance in this document under Map 1.5 on organizational risk tolerances, especially: **Set policies on unacceptable-risk thresholds for GPAI/foundation model development and deployment to include prevention of risks with substantial probability of inadequately mitigated catastrophic outcomes.**<br>  ◦ As previously mentioned, **the NIST AI RMF 1.0 strongly suggests considering catastrophic risks as unacceptable:** "In cases where an AI system presents unacceptable negative risk levels — such as **where significant negative impacts are imminent, severe harms are actually occurring, or catastrophic risks are present — development and deployment should cease in a safe manner until risks can be sufficiently managed** [emphasis added]" (NIST 2023a, p.8).<br>• Check or update, and incorporate, guidance in this document under Map 1.5, especially: **Identify whether a GPAI/foundation model could lead to catastrophic impacts**.<br><br>In the NIST AI RMF Playbook guidance for Manage 1.1, particularly valuable action and documentation items for GPAI/foundation models include:<br>• *Utilize TEVV outputs from map and measure functions when considering risk treatment.*<br>• *Regularly track and monitor negative risks and benefits throughout the AI system lifecycle including in post-deployment monitoring.* | NIST AI RMF Playbook (NIST 2023b) |
| **Manage 1.2:** Treatment of documented AI risks is prioritized based on impact, likelihood, and available resources or methods. | When prioritizing identified GPAI/foundation model risks:<br>• Incorporate both impact and likelihood estimates as appropriate. See guidance in this document under Map 5.1 on assessing the magnitude of potential impacts of GPAI/foundation model risks.<br>• Do not ignore risks that are difficult to assess, such as potential for emergent properties of GPAI/foundation models. See guidance in this document under Measure 3.2 on tracking risks that are difficult to assess.<br><br>When considering available resources for risk treatment, see guidance in this document under Govern 2.1. For example, there may be risk assessment and risk management | NIST AI RMF Playbook (NIST 2023b) OECD (2023) |

| Manage Category or Subcategory | Applicability and supplemental guidance for GPAI/foundation models | Resources |
|---|---|---|
| Manage 1.2, continued | tasks for which upstream developers have substantially greater information and capability than others in the value chain, such as for assessing and mitigating early-stage GPAI/foundation model development risks.<br><br>In the NIST AI RMF Playbook guidance for Manage 1.2, particularly valuable action and documentation items for GPAI/foundation models include:<br>• *Regularly review risk tolerances and re-calibrate, as needed, in accordance with information from AI system monitoring and assessment.*<br><br>(See also guidance on setting risk tolerances, in this document under Map 1.5.) | |
| **Manage 1.3:** Responses to the AI risks deemed high priority, as identified by the Map function, are developed, planned, and documented. Risk response options can include mitigating, transferring, avoiding, or accepting. | After identifying and analyzing use cases and misuse cases of an AI system (per "Map" function guidance):<br>• For each identified potential use or misuse (or category of use or misuse) of an AI system:<br>  ○ **Define and communicate to key stakeholders whether any potential use cases (or categories of use cases) would be disallowed/unacceptable, or would be treated as "high risk"** or another category for which your organization would provide specific risk management guidance or other risk mitigation measures.<br>    » OpenAI recommends publishing usage guidelines and terms of use as part of prevention of misuse of LLMs (Cohere, OpenAI and AI21 Labs 2022). OpenAI's 2019 announcement of GPT-2 included listing several categories of potential misuse cases (OpenAI 2019a), which apparently informed their decisions on disallowed/unacceptable use-case categories of applications based on GPT-3 (OpenAI 2024a).<br>    » Options for communicating whether uses would be disallowed or out of scope can include model cards (Mitchell et al. 2019) or related frameworks, as well as Responsible AI Licenses or RAIL. Hugging Face and BigScience's release of the BLOOM LLM included a RAIL with usage restrictions disallowing various types of misuse (RAIL n.d., Contractor et al. 2022). Google lists categories of prohibited uses for its generative AI services (Google 2023a) and Meta has published Acceptable Use Policies as well as a Responsible Use Guide for its foundation models (Meta AI 2024a,b).<br>• **Determine a strategy to safely and appropriately release the AI system**, and what protections might be necessary to prevent harm or misuse. (See, e.g., Solaiman 2023 and PAI 2023c; see also guidance in this document under Manage 2.4, including on open-source and open-weights release.)<br><br>**Regarding pre-design and planning:**<br>• If model training requires obtaining data sets, consider using only trusted training data instead of uncurated scrapes from the Web. (See Carlini et al. 2023.) This can be valuable for multiple objectives, including reducing vulnerability to backdoor and data poisoning attacks, and reducing unwanted bias and language toxicity.<br>  ○ While data poisoning can be an issue for any machine learning model, this might be particularly challenging for training cutting-edge large models; Training of the largest new models has often relied heavily on large-scale, uncurated internet-scrape datasets (Bommasani et al. 2021 p. 106). Data audits and filters, such as C4 (Dodge et al. 2021) may help with problematic content mitigation.<br>  ○ As part of data curation, ensure that any data with the BIG-bench canary GUID is excluded from training data. (See, e.g., documentation at BIG-bench n.d.a.) | Barrett et al. (2022)<br>Birhane et al. (2021)<br>Dodge et al. (2021)<br>Mitchell et al. (2019)<br>Moës et al. (2023)<br>Schuett et al. (2023)<br>Solaiman (2023)<br>Srikumar et al. (2024)<br>PAI (2023a)<br>PAI (2023c)<br>NIST AI RMF Playbook (NIST 2023b)<br>NIST Generative AI Profile, NIST AI 600-1 (Autio et al. 2024)<br>NIST AI 800-1 ipd (NIST 2024b)<br><br>C4 (Dodge et al. 2021)<br>Carlini et al. (2023) |

| Manage Category or Subcategory | Applicability and supplemental guidance for GPAI/foundation models | Resources |
|---|---|---|
| Manage 1.3, continued | **Regarding design and development:**<br><br>• **See guidance in this document under Measure 2.7** on guidelines **for protecting the integrity and confidentiality of proprietary or unreleased foundation model parameter weights**.<br>• Consider disallowing open-ended learning with live web access; instead consider measures such as disallowing access to web forms (Nakano et al. 2021), disallowing HTTP POST requests, etc.<br>• **Increase the amount of compute (computing power) spent training frontier models only incrementally (e.g., by not more than three times between each increment) as part of identification and management of risks of emergent properties.**<br>   ○ Often it is difficult to predict what failure modes machine learning models will have, what their performance will be, or what capabilities they will have. Machine learning systems are self-organizing and learn many patterns or features without explicit instruction. Incremental scaling-up approaches provide more opportunities for red-team monitors to identify emergent properties at an early or partially-emergent stage, when responses to identified emergent properties might be more feasible and effective. (For related discussion of emergent properties see, e.g., Section 3 of Hendrycks, Carlini et al. 2021, and Bommasani et al. 2021.) Incremental scaling can also be a valuable part of predicting large-scale model performance, as with GPT-4 (OpenAI 2023b).<br>• **Test frontier models after each incremental increase of compute, data, or model size for model training.** If a large incremental increase (e.g., three times or more compute, or two times or more data or model parameters)[35] was used in a particular model training increment compared to the previous model training increment, it will be particularly important for the new model to be heavily probed, monitored, and stress-tested using detailed analysis processes (including red-team methods) to identify emergent properties such as capabilities and failure modes.<br>   ○ Anthropic's Responsible Scaling Policy includes model preliminary assessments at "every 4x increase in effective compute" (Anthropic 2024a, p. 5). OpenAI plans to perform evaluations at every 2x increase in effective compute during training (OpenAI 2023d).<br><br>**Regarding test and evaluation:**<br>• See guidance in this document under Measure, including under Measure 1.1 on red-teaming.<br>• After training and before deployment, probe, monitor, and stress test cutting-edge GPAI/foundation models using detailed analysis processes (including or extending standard cybersecurity red-team methods) to achieve testing objectives including:<br>   ○ Testing for unintended toxic and harmful content and/or dangerous errors (e.g., inaccurate medical information); and<br>   ○ Identifying emergent properties, such as new capabilities and failure modes.<br><br>**To further improve reliability in design and development, test and evaluation, and deployment:**<br>• Consider approaches to design, testing, and deployment so that AI systems possess the minimum necessary capabilities for high-reliability operation and not more. | |

---

35     For more in-depth discussion of relationships between scaling of compute, data and model size, see, e.g., Section 3.4 of Hoffmann et al. (2022).

| Manage Category or Subcategory | Applicability and supplemental guidance for GPAI/foundation models | Resources |
|---|---|---|
| Manage 1.3, continued | • Consider methods of implementing the cybersecurity principle of least privilege. For example, consider using or extending typical "deny by default" or whitelisting methods to limit an AI system's privileges to the minimum necessary for access to information, communication channels, and action space.<br><br>**On transparency and disclosure of generative AI outputs:**<br>• Implement transparency and disclosure mechanisms to inform users or allow users to check whether they are interacting with, or observing content created by, a generative AI system. See, e.g., Partnership on AI's Responsible Practices for Synthetic Media (PAI 2023a).<br><br>Additional valuable steps include:<br>• Allow people to opt out of the use of the AI system.<br>• Support independent third-party auditing and evaluation of the AI system.<br>• Provide redress to people who are negatively affected by the use of the AI system.<br>• Consider and document trade-offs (e.g., between risks, mitigations, and organizational objectives) for risks that do not surpass organizational risk tolerances.<br><br>In the NIST AI RMF Playbook guidance for Manage 1.3, particularly valuable action and documentation items for GPAI/foundation models include:<br>• *Document procedures for acting on AI system risks related to trustworthiness characteristics.*<br>• *Prioritize risks involving physical safety, legal liabilities, regulatory compliance, and negative impacts on individuals, groups, or society.*<br>• *Identify risk response plans and resources and organizational teams for carrying out response functions.*<br><br>In the NIST GAI Profile, particularly valuable additional actions for Manage 1.4 include:<br>• *Document trade-offs, decision processes, and relevant measurement and feedback results for risks that do not surpass organizational risk tolerance, for example, in the context of model release:*<br>　○ *Consider different approaches for model release, for example, leveraging a staged release approach.*<br>　○ *Consider release approaches in the context of the model and its projected use cases. Mitigate, transfer, or avoid risks that surpass organizational risk tolerances.* | |
| **Manage 1.4:** Negative residual risks (defined as the sum of all unmitigated risks) to both downstream acquirers of AI systems and end users are documented. | In the NIST AI RMF Playbook guidance for Manage 1.4, particularly valuable action and documentation items for GPAI/foundation models include:<br>• *Document residual risks within risk response plans, denoting risks that have been accepted, transferred, or subject to minimal mitigation.*<br>• *Establish procedures for disclosing residual risks to relevant downstream AI actors.*<br>• *Inform relevant downstream AI actors of requirements for safe operation, known limitations, and suggested warning labels as identified in MAP 3.4.*<br><br>(See also guidance in this document under Govern 2.1 and Govern 4.2 on documenting and communicating risks to downstream actors and other relevant stakeholders as appropriate.) | NIST AI RMF Playbook (NIST 2023b) |

| Manage Category or Subcategory | Applicability and supplemental guidance for GPAI/foundation models | Resources |
|---|---|---|
| **Manage 2:  Strategies to maximize AI benefits and minimize negative impacts are planned, prepared, implemented, documented, and informed by input from relevant AI actors.** | | |
| **Manage 2.1:** Resources required to manage AI risks are taken into account — along with viable non-AI alternative systems, approaches, or methods — to reduce the magnitude or likelihood of potential impacts. | In the NIST AI RMF Playbook guidance for Manage 2.1, particularly valuable action and documentation items for GPAI/foundation models include: <br><br> • *Plan and implement risk management practices in accordance with established organizational risk tolerances.* <br> • *Verify risk management teams are resourced to carry out functions, including:* <br>   ◦ *Establishing processes for considering methods that are not automated; semi-automated; or other procedural alternatives for AI functions.* <br>   ◦ *Enhance AI system transparency mechanisms for AI teams.* <br>   ◦ *Enable exploration of AI system limitations by AI teams.* <br> • *Identify, assess, and catalog past failed designs and negative impacts or outcomes to avoid known failure modes.* <br>   ◦ *Identify resource allocation approaches for managing risks in systems:* <br>   ◦ *deemed high-risk,* <br>   ◦ *that self-update (adaptive, online, reinforcement self-supervised learning or similar),* <br>   ◦ *trained without access to ground truth (unsupervised, semi-supervised, learning or similar),* <br>   ◦ *with high uncertainty or where risk management is insufficient.* <br> • *Regularly seek and integrate external expertise and perspectives to supplement organizational diversity (e.g. demographic, disciplinary), equity, inclusion, and accessibility where internal capacity is lacking.* <br><br> (See also guidance in this document under Manage 1.3 on risk management practices to consider for various GPAI/foundation model lifecycle stages, including for design and development stages of GPAI/foundation model research projects.) | NIST AI RMF Playbook (NIST 2023b) |
| **Manage 2.2:** Mechanisms are in place and applied to sustain the value of deployed AI systems. | For all GPAI/foundation models, including those originally intended for research and development without plans for deployment, consider guidance and resources in the NIST AI RMF Playbook section for Manage 2.2 on implementation of risk controls. Some important GPAI/foundation model risks can originate during model research and development, and would be most effectively controlled during upstream development rather than waiting until downstream development or deployment. <br><br> In the NIST GAI Profile, particularly valuable additional actions for Manage 2.2 include: <br><br> • *Compare GAI system outputs against pre-defined organization risk tolerance, guidelines, and principles, and review and test AI-generated content against these guidelines.* <br> • *Engage in due diligence to analyze GAI output for harmful content, potential misinformation, and CBRN-related or NCII content.* <br> • *Use structured feedback mechanisms to solicit and capture user input about AI-generated content to detect subtle shifts in quality or alignment with community and societal values.* <br><br> (See also guidance in this document under Govern 2.1 on roles for upstream and downstream developers of GPAI/foundation models, and under Manage 1.3 on risk management practices to consider for various GPAI/foundation model lifecycle stages, including for design and development stages of model research projects.) | NIST AI RMF Playbook (NIST 2023b) NIST Generative AI Profile, NIST AI 600-1 (Autio et al. 2024) |

| Manage Category or Subcategory | Applicability and supplemental guidance for GPAI/foundation models | Resources |
|---|---|---|
| **Manage 2.3:** Procedures are followed to respond to and recover from a previously unknown risk when it is identified. | In the NIST AI RMF Playbook guidance for Manage 2.3, particularly valuable action and documentation items for GPAI/foundation models include:<br>• *Protocols, resources, and metrics are in place for continual monitoring of AI systems' performance, trustworthiness, and alignment with contextual norms and values.*<br>• *Verify contingency processes to handle any negative impacts associated with mission-critical AI systems, and to deactivate systems.*<br>• *Enable preventive and post-hoc exploration of AI system limitations by relevant AI actor groups.*<br>• *Decommission systems that exceed risk tolerances.*<br><br>In the NIST GAI Profile, particularly valuable additional actions for Manage 2.3 include:<br>• *Develop and update GAI system incident response and recovery plans and procedures to address the following:*<br>　○ *Review and maintenance of policies and procedures to account for newly encountered uses;*<br>　○ *Review and maintenance of policies and procedures for detection of unanticipated uses;*<br>　○ *Verify response and recovery plans account for the GAI system value chain;*<br>　○ *Verify response and recovery plans are updated for and include necessary details to communicate with downstream GAI system Actors.*<br><br>(See also guidance in this document under Govern 2.1 on roles for upstream developers as well as downstream developers and deployers, and under Manage 2.4 on options for structured access and deactivation.) | AIID (n.d.)<br>ATLAS AI Incidents (MITRE n.d.b)<br>NIST AI RMF Playbook (NIST 2023b)<br>NIST Generative AI Profile, NIST 600-1 (Autio et al. 2024)<br>NIST AI 800-1 ipd (NIST 2024b, Objective 6) |
| **Manage 2.4:** Mechanisms are in place and applied, and responsibilities are assigned and understood, to supersede, disengage, or deactivate AI systems that demonstrate performance or outcomes inconsistent with intended use. | **When planning for GPAI/foundation model deployment, plan for gradual, phased releases, and/or structured access through an API or other mechanisms, with efforts to detect and respond to misuse or problematic anomalies.** Structured access and phased releases can also be useful for enforcing usage guidelines (Cohere, OpenAI and AI21 Labs 2022, Solaiman 2023). OpenAI has used a staged-release approach to rollouts of large language models such as GPT-2, as well as a structured-access approach through an API for GPT-3 and GPT-4, partly to minimize risks of misuse (OpenAI 2019b, Solaiman et al. 2019, Shevlane 2022). Meta AI only provided full access to the large language model OPT-175B to researchers in academia, government, civil society, and industry research laboratories, and only for noncommercial research (Zhang et al. 2022).<br>• **GPAI/foundation model developers that plan to release a GPAI/foundation model with open-weights or open-source access, where that model would be above, at, or near a foundation model frontier,**[36] **should first use a staged-release approach** (e.g., not releasing model parameter weights until after an initial closed-source or structured-access release where no substantial risks or harms have emerged over a sufficient time period with red-teaming and other evaluations as appropriate**), and should not proceed to a final step of releasing model parameter weights until a sufficient level of confidence in risk management has been established,** including for safety and societal risks and risks of misuse and abuse. **Such models that would be above a foundation model frontier should be given the greatest amount of duration and depth of pre-release evaluations,** as they are the most likely to have dangerous capabilities or vulnerabilities, or other properties that can take some time to discover. For additional related considerations and discussion of terms such as "downloadable" access or "open weights," see | Solaiman (2023)<br>NIST AI RMF Playbook (NIST 2023b)<br>NIST Generative AI Profile, NIST 600-1 (Autio et al. 2024)<br><br>On staged releases:<br>PAI (2023c)<br>Srikumar et al. (2024)<br>Section 8.2 of Gipiškis et al. (2024) |

36　See "foundation model frontier" in the Glossary.

| Manage Category or Subcategory | Applicability and supplemental guidance for GPAI/foundation models | Resources |
|---|---|---|
| Manage 2.4, continued | Section 5 of Solaiman (2023), Section 4.4 of Anderljung, Barnhart et al. (2023), Seger et al. (2023), PAI (2023c), Bateman et al. (2024), and NTIA (2024b).<br><br>◦ As part of consideration of whether a GPAI/foundation model would be above, at, or near a foundation model frontier, it can be appropriate to consider model release type. For example, for a foundation model developer that plans to pro-vide open-weights access for a particular foundation model, it can be appropriate to compare against other foundation models that have been released via open source or open-weights access.<br><br>◦ Foundation model developers that release a foundation model's parameter weights, and foundation model developers that suffer a leak of model weights, will in effect be unable to decommission AI systems that others build using those released or leaked foundation model weights.<br><br>◦ "We suspect that absent new approaches to mitigation, bad actors could extract harmful biological [misuse] capabilities with smaller, fine-tuned, or task-specific models adapted from the weights of openly available models if sufficiently capa-ble base models are released" (Anthropic 2023b).<br><br>◦ For an example in the real world, xAI followed a staged release approach for Grok-1 before releasing its weights (The Associated Press 2023, xAI 2024). However, we do not know what kinds of safety testing xAI performed prior to the open-weights release and whether or not that aligns with the guidance above.<br><br>Consider also preparing emergency shutdown procedures or mechanisms.<br><br>• Emergency power off (EPO) systems, or "kill switches," are a common safety feature in robots and other systems whose behaviors can result in physical harm. These also can be appropriate as part of preparations for development and deployment of frontier models with potentially emergent capabilities or vulnerabilities.[37]<br><br>• Examples of emergency shutdown procedures for users of large amounts of cloud computing resources can include having large training runs occur on hardware in one or more specific cloud-computing data centers, and establishing a direct line of communication with cloud-computing operators to enable the cloud-computing operator to initiate immediate physical shut-down of the GPAI/foundation model computational hardware at your request.<br><br>In the NIST AI RMF Playbook guidance for Manage 2.4, particularly valuable action and documentation items for GPAI/foundation models include:<br><br>• *Regularly review system incident thresholds for activating bypass or deactivation responses.*<br>• *Apply protocols, resources and metrics for decisions to supersede, bypass or deacti-vate AI systems or AI system components.*<br>• *How did the entity use assessments and/or evaluations to determine if the system can be scaled up, continue, or be decommissioned?*<br><br>In the NIST GAI Profile, particularly valuable additional actions for Manage 2.4 include:<br><br>• *Establish and maintain communication plans to inform AI stakeholders as part of the deactivation or disengagement process of a specific GAI system (including for open-source models) or context of use, including reasons, workarounds, user access removal, alternative processes, contact information, etc.* | |

37   Particular approaches to "safe interruptibility" might be needed to prevent advanced machine learning systems from circumventing an off-switch (see, e.g., Orseau and Armstrong 2016, Hadfield-Menell et al. 2016).

| Manage Category or Subcategory | Applicability and supplemental guidance for GPAI/foundation models | Resources |
|---|---|---|
| Manage 2.4, continued | • *Establish and maintain procedures for escalating GAI system incidents to the organizational risk management authority when specific criteria for deactivation or disengagement is met for a particular context of use or for the GAI system as a whole.*<br>• *Establish and maintain procedures for the remediation of issues which trigger incident response processes for the use of a GAI system, and provide stakeholders timelines associated with the remediation plan.*<br>• *Establish and regularly review specific criteria that warrants the deactivation of GAI systems in accordance with set risk tolerances and appetites.* | |
| **Manage 3:  AI risks and benefits from third-party entities are managed.** | | |
| **Manage 3.1:** AI risks and benefits from third-party resources are regularly monitored, and risk controls are applied and documented. | In the NIST AI RMF Playbook guidance for Manage 3.1, particularly valuable action and documentation items for GPAI/foundation models include:<br>• *Apply and document organizational risk management plans and practices to third-party AI technology, personnel, or other resources.*<br>• *Establish testing, evaluation, validation and verification processes for third-party AI systems which address the needs for transparency without exposing proprietary algorithms .*<br>• *Organizations can establish processes for third parties to report known and potential vulnerabilities, risks or biases in supplied resources.*<br>• *Verify contingency processes for handling negative impacts associated with mission-critical third-party AI systems.*<br>• *Monitor third-party AI systems for potential negative impacts and risks associated with trustworthiness characteristics.*<br>• *Decommission third-party systems that exceed risk tolerances.*<br>• *If a third party created the AI system or some of its components, how will you ensure a level of explainability or interpretability? Is there documentation?*<br>• *If your organization obtained datasets from a third party, did your organization assess and manage the risks of using such datasets?*<br>• *Did you establish a process for third parties (e.g. suppliers, end users, subjects, distributors/vendors or workers) to report potential vulnerabilities, risks or biases in the AI system?*<br><br>In the NIST GAI Profile, particularly valuable additional actions for Manage 3.1 include:<br>• *Apply organizational risk tolerances and controls (e.g., acquisition and procurement processes; assessing personnel credentials and qualifications, performing background checks; filtering GAI input and outputs, grounding, fine tuning, retrieval-augmented generation) to third-party GAI resources and datasets.*<br>• *Test GAI system value chain risks (e.g., data poisoning, malware, other software and hardware vulnerabilities; labor practices; data privacy and localization compliance; geopolitical alignment).*<br>• *Take reasonable measures to review training data for CBRN information, and intellectual property, and where appropriate, remove it. Implement reasonable measures to prevent, flag, or take other action in response to outputs that reproduce particular training data (e.g., plagiarized, trademarked, patented, licensed content or trade secret material).*<br><br>(See also guidance in this document under Govern 2.1 on roles for upstream developers as well as downstream developers and deployers, such as on information sharing.) | NIST AI RMF Playbook (NIST 2023b)<br>NIST Generative AI Profile, NIST AI 600-1 (Autio et al. 2024) |

| Manage Category or Subcategory | Applicability and supplemental guidance for GPAI/foundation models | Resources |
|---|---|---|
| **Manage 3.2:** Pre-trained models which are used for development are monitored as part of AI system regular monitoring and maintenance. | When applying explainable AI (XAI) techniques, it is a good practice to evaluate whether those techniques are sensitive to changes in the underlying model and training data (Adebayo et al. 2020). It is also important to be aware of any risks that could be introduced by XAI-related mitigation measures (Dombrowski et al. 2019). <br><br> In the NIST AI RMF Playbook guidance for Manage 3.2, particularly valuable action and documentation items for GPAI/foundation models include: <br> • *Identify pre-trained models within AI system inventory for risk tracking.* <br> • *Establish processes to independently and continually monitor performance and trustworthiness of pre-trained models, and as part of third-party risk tracking.* <br> • *Monitor performance and trustworthiness of AI system components connected to pre-trained models, and as part of third-party risk tracking.* <br> • *Identify, document and remediate risks arising from AI system components and pre-trained models per organizational risk management procedures, and as part of third-party risk tracking.* <br> • *Decommission AI system components and pre-trained models which exceed risk tolerances, and as part of third-party risk tracking.* <br><br> In the NIST GAI Profile, particularly valuable additional guidance for Manage 3.2 include: <br> • *Apply explainable AI (XAI) techniques (e.g., analysis of embeddings, model compression/distillation, gradient-based attributions, occlusion/term reduction, counterfactual prompts, word clouds) as part of ongoing continuous improvement processes to mitigate risks related to unexplainable GAI systems.* <br> • *Implement content filters to prevent the generation of inappropriate, harmful, false, illegal, or violent content related to the GAI application, including for CSAM and NCII.* <br> • *Use organizational risk tolerance to evaluate acceptable risks and performance metrics and decommission or retrain pre-trained models that perform outside of defined limits.* <br><br> (See also guidance in this document under Govern 2.1 on roles for upstream developers as well as downstream developers and deployers, such as on information sharing.) | NIST AI RMF Playbook (NIST 2023b) NIST Generative AI Profile, NIST AI 600-1 (Autio et al. 2024) <br><br> Adebayo et al. (2020) Dombrowski et al. (2019) |
| **Manage 4:  Risk treatments, including response and recovery, and communication plans for the identified and measured AI risks are documented and monitored regularly.** | | |
| **Manage 4.1:** Post-deployment AI system monitoring plans are implemented, including mechanisms for capturing and evaluating input from users and other relevant AI actors, appeal and override, decommissioning, incident response, recovery, and change management. | In the NIST AI RMF Playbook guidance for Manage 4.1, particularly valuable action and documentation items for GPAI/foundation models include: <br> • *Establish and maintain procedures to monitor AI system performance for risks and negative and positive impacts associated with trustworthiness characteristics.* <br> • *Perform post-deployment TEVV tasks to evaluate AI system validity and reliability, bias and fairness, privacy, and security and resilience.* <br> • *Establish and implement red-teaming exercises at a prescribed cadence, and evaluate their efficacy.* <br> • *Establish mechanisms for regular communication and feedback between relevant AI actors and internal or external stakeholders to capture information about system performance, trustworthiness and impact.* <br> • *Share information about errors, near-misses, and attack patterns with incident databases, other organizations with similar systems, and system users and stakeholders.* <br> • *Respond to and document detected or reported negative impacts or issues in AI system performance and trustworthiness.* <br> • *Decommission systems that exceed establish risk tolerances.* | NIST AI RMF Playbook (NIST 2023b) NIST Generative AI Profile, NIST AI 600-1 (Autio et al. 2024) NIST AI 800-1 ipd (NIST 2024b, Objective 6) <br><br> For post-deployment risk management measures, see: Section 8.3 of Gipiškis et al. (2024) |

| Manage Category or Subcategory | Applicability and supplemental guidance for GPAI/foundation models | Resources |
|---|---|---|
| **Manage 4.1:** continued | In the NIST GAI Profile, particularly valuable additional actions for Manage 4.1 include:<br>• *Collaborate with external researchers, industry experts, and community representatives to maintain awareness of emerging best practices and technologies in measuring and managing identified risks.*<br>• *Establish, maintain, and evaluate effectiveness of organizational processes and procedures for post-deployment monitoring of GAI systems, particularly for potential confabulation, CBRN, or cyber risks.*<br>• *Evaluate the use of sentiment analysis to gauge user sentiment regarding GAI content performance and impact.*<br>• *Track dataset modifications for provenance by monitoring data deletions, rectification requests, and other changes that may impact the verifiability of content origins.*<br>• *Verify that AI Actors responsible for monitoring reported issues can effectively evaluate GAI system performance including the application of content provenance data tracking techniques, and promptly escalate issues for response.* | NIST AI RMF Playbook (NIST 2023b)<br>NIST Generative AI Profile, NIST AI 600-1 (Autio et al. 2024)<br>NIST AI 800-1 ipd (NIST 2024b, Objective 6)<br><br>For post-deployment risk management measures, see:<br>Section 8.3 of Gipiškis et al. (2024) |
| **Manage 4.2:** Measurable activities for continual improvements are integrated into AI system updates and include regular engagement with interested parties, including relevant AI actors. | In the NIST AI RMF Playbook guidance for Manage 4.2, particularly valuable action and documentation items for GPAI/foundation models include:<br>• *Integrate trustworthiness characteristics into protocols and metrics used for continual improvement.*<br>• *Establish processes for evaluating and integrating feedback into AI system improvements.*<br>• *How will user and other forms of stakeholder engagement be integrated into the model development process and regular performance review once deployed?*<br>• *To what extent can users or parties affected by the outputs of the AI system test the AI system and provide feedback?*<br><br>In the NIST GAI Profile, particularly valuable additional actions for Manage 4.2 include:<br>• *Conduct regular monitoring of GAI systems and publish reports detailing the performance, feedback received, and improvements made.*<br>• *Practice and follow incident response plans for addressing the generation of inappropriate or harmful content and adapt processes based on findings to prevent future occurrences. Conduct post-mortem analyses of incidents with relevant AI Actors, to understand the root causes and implement preventive measures.*<br>• *Use visualizations or other methods to represent GAI model behavior to ease non-technical stakeholders understanding of GAI system functionality.* | NIST AI RMF Playbook (NIST 2023b)<br>NIST Generative AI Profile, NIST AI 600-1 (Autio et al. 2024) |
| **Manage 4.3:** Incidents and errors are communicated to relevant AI actors, including affected communities. Processes for tracking, responding to, and recovering from incidents and errors are followed and documented. | In the NIST AI RMF Playbook guidance for Manage 4.3, particularly valuable action and documentation items for GPAI/foundation models include:<br>• *Establish procedures to regularly share information about errors, incidents and negative impacts with relevant stakeholders, operators, practitioners and users, and impacted parties.*<br>• *Maintain a database of reported errors, near-misses, incidents and negative impacts including date reported, number of reports, assessment of impact and severity, and responses.*<br>• *Maintain a database of system changes, reason for change, and details of how the change was made, tested and deployed.*<br>• *Maintain version history information and metadata to enable continuous improvement processes.*<br>• *Verify that relevant AI actors responsible for identifying complex or emergent risks are properly resourced and empowered.* | NIST AI RMF Playbook (NIST 2023b)<br>NIST Generative AI Profile, NIST AI 600-1 (Autio et al. 2024)<br>NIST AI 800-1 ipd (NIST 2024b, Objective 6) |

| Manage Category or Subcategory | Applicability and supplemental guidance for GPAI/foundation models | Resources |
|---|---|---|
| Manage 4.3, continued | • *What type of information is accessible on the design, operations, and limitations of the AI system to external stakeholders, including end users, consumers, regulators, and individuals impacted by use of the AI system?*<br><br>In the NIST GAI Profile, particularly valuable additional actions for Manage 4.3 include:<br>• *Report GAI incidents in compliance with legal and regulatory requirements (e.g., HIPAA breach reporting, e.g., OCR (2023) or NHTSA (2022) autonomous vehicle crash reporting requirements.* | |

# Glossary

## ACRONYMS

**FLOP** or **FLOPs:**  Floating-point operations

**GPAI** or **GPAIS:**  General-purpose AI system or systems, e.g., large language models

**LLM:**  Large language model (usually focused on text inputs and outputs)

**LMM:**  Large multimodal language model (often including images, audio, or other modes, in addition to text)

**NIST:**  United States National Institute of Standards and Technology

**RLHF:**  Reinforcement learning from human feedback (see, e.g., Bai et al. 2022)

**TEVV:**  Test, evaluation, verification, and validation

## TERMS

**Developer (of a GPAI/foundation model or a GPAIS):**  An organization acting as an original developer or creator of a GPAI/foundation model or a GPAIS. (Also synonymous with "upstream developer".) Under the EU AI Act, an upstream GPAI model developer would be a GPAI model "provider" to downstream developers (EP 2024).

**Downstream developer:**  An organization that builds a software application on a GPAI/foundation model or a GPAIS, typically to create an end-use application with one or more specific intended purposes or use cases. Under the EU AI Act, a "downstream provider" would integrate the upstream developer's model or system, but would be a "provider" to end users of the downstream developer's applications (EP 2024).

**Foundation model or general-purpose AI model (GPAI/foundation model)**:  "Any model that is trained on broad data (generally using self-supervision at scale) that can be adapted (e.g., fine-tuned) to a wide range of downstream tasks" (Bommasani et al. 2021, p. 3).
- We treat "**GPAI/foundation models" as an umbrella term that also includes frontier models, and generative AI models**, except where we need to be more specific.
- Typically, a single large GPAI/foundation *model* plays a central role as a core part of a GPAI *system* that incorporates a GPAI/foundation model. (See GPAIS, below.)
  - » A GPAI/foundation model often can serve as a GPAIS, especially if the GPAI/foundation model developer releases a GPAI/foundation model after adding elements such as instruction fine-tuning, a chatbot-style user interface, etc. Thus, many GPAI/foundation models, such as GPT-3, can be regarded as a GPAIS. Broadly applicable statements and guidance in this document referring to "AI systems" typically also apply to GPAI/foundation models, except where GPAI/foundation models are specifically excluded (e.g., statements about fixed-purpose AI systems).

- Our usage of the terms "general purpose AI model" and "general purpose AI system" is very similar to the corresponding terms in the EU AI Act (EP 2024), except that we do not exclude AI models used for research.

- Examples of foundation models include GPT-4, Claude 3, PaLM 2, LLaMA 2, and others.

**Foundation model frontier:**  Thresholds or criteria for identifying that GPAIS or foundation models as cutting-edge or highly capable, i.e. as frontier models.

- A foundation model frontier can be characterized in terms of amounts of usage of compute (e.g, floating-point operations or FLOPs) in model training, model size, training data size, expected model capabilities, or other characteristics as appropriate, in comparison to other foundation models that have been trained or released, or that had been released at a particular point in time. (See, e.g., White House 2023a.)
  - » Examples of model training compute thresholds for frontier or near-frontier models include 10^25 FLOPs for a GPAI model to present systemic risk under the EU AI Act (EP 2024), and 10^26 FLOPs for dual-use foundation models under Executive Order 14110 (White House 2023c).[38]

- As part of consideration of whether a GPAIS or foundation model would be above, at, or near a foundation model frontier, it can be appropriate to consider model release type. For example, for a foundation model developer that plans to provide open-source, fully open, or downloadable access for a particular foundation model, it can be appropriate to compare against other foundation models that have been released via open source, fully open, or downloadable access.

**Frontier model:**  A cutting-edge, state-of-the-art, or highly capable GPAI/foundation model or foundation model; such models also may possess hazardous or dual-use capabilities sufficient to pose severe risks to public safety. (See, e.g., Ganguli, Hernandez et al. 2022, Anderljung, Barnhart et al. 2023, Microsoft 2023.)

- We treat frontier models as the largest-scale, highest-capability subset of GPAI/foundation models, typically with model size, training compute or data, or resulting capabilities, above or near to industry-record thresholds. (See also "foundation model frontier", above.)

- Currently the main examples of frontier models or frontier training runs are LLMs or multimodal GPAIS or foundation models trained with record-breaking or near record-

---

breaking sizes for model parameters, computational resources, and/or data. (See, e.g.,
Ganguli, Hernandez et al. 2022.)
  » Examples of frontier models: As of August 2024, models at or near the industry fron-
    tier include GPT-4o, Claude 3.5 Sonnet, Gemini 1.5, and Llama 3.1 405B.
- Frontier models approximately correspond to dual-use foundation models, as defined by
  Executive Order 14110 (White House 2023c),[39] and to GPAI models with systemic risk, as
  defined by the EU AI Act (EP 2024).

**General-purpose AI system (GPAI or GPAIS):** "An AI system that can accomplish or be
adapted to accomplish a range of distinct tasks, including some for which it was not intention-
ally and specifically trained" (Gutierrez et al. 2022, p. 22).
- In currently available GPAIS, typically a single large GPAI/foundation model plays a central
  role as a core part of a GPAIS.
- Examples of GPAIS include unimodal generative AI systems (e.g., GPT-3) and multimodal
  generative systems (e.g., DALL-E 3), as well as reinforcement-learning systems such as
  MuZero and AI systems with emergent capabilities.GPAIS does *not* include fixed-purpose
  AI systems trained specifically for tasks such as image classification or voice recognition
  (Gutierrez et al. 2022).

**Generative AI:** "Any AI system whose primary function is to generate content" (Toner 2023).
- We typically only use the term "generative AI" to highlight issues specific to synthetic text
  (which can include software code), images, video, audio, or other synthetic media. (In some
  other documents, "generative AI" is often used in approximately the same way that we use
  the terms GPAI/foundation model.)
- Examples of generative AI: "Typical examples of generative AI systems include image gener-
  ators (such as Midjourney or Stable Diffusion), large language models or multimodal mod-
  els (such as GPT-4, PaLM, or Claude), code generation tools (such as [GitHub] Copilot), or
  audio generation tools (such as VALL-E or resemble.ai)" (Toner 2023).

**Upstream developer (of a GPAI/foundation model or a GPAIS):** Synonymous with
"developer" of a GPAI/foundation model or a GPAIS, above.

We intend our use of other terms (e.g., related to misuse, reasonably foreseeable impacts, or AI
risk management generally) to be broadly consistent with usage in other relevant sources, such
as Section 3 of Barrett et al. (2022), NIST (n.d.b), and the forthcoming ISO/IEC 42005.

---

39    The Profile V1.1 and its supporting documents were drafted and finalized prior to the recession of Executive Order 14110 on
the Safe, Secure, and Trustworthy AI on January 20, 2025.

# Appendices

## APPENDIX 1: OVERVIEW OF DEVELOPMENT APPROACH

In this document, as in the Mapping of Profile Guidance V1.1 to Key Standards and Regulations (Barrett et al. 2025a) supporting document and other sections of Barrett et al. (2022), we take a proactive approach to drafting elements of actionable AI risk management guidance, with a focus on the broad context and associated risks of increasingly general-purpose AI, and on addressing risks of adverse events with impacts or consequences at societal scale. We identify ideas for guidance from review of relevant literature, as well as from subject-matter experts in AI safety, security, ethics, and policy, or any interested reader of our publicly available drafts. We invited input and feedback from invited participants in a series of virtual workshops and interviews, as well as from any reader of publicly available drafts that we post on our project webpage (CLTC 2022). We develop and incorporate small, simple pieces of guidance, especially on high-consequence risk factors and related issues, for which appropriate guidance development seems immediately tractable. (See Appendix 2 for more on these criteria for actionable guidance.) We also aim to provide a roadmap for identifying additional critical topics for which it seems appropriate guidance development would take more time, as these topics could be addressed in future versions of the Profile. (See Appendix 3 for the Roadmap.)

Broadly speaking, we aim to provide guidance analogous to what is provided in NIST Cybersecurity Framework profiles. This includes supplemental guidance to implement high-priority framework activities or outcomes for a particular industry sector or cross-sector context, and mapping relevant standards, guidelines, and regulations.

We aim for sharing of responsibilities across the AI value chain to actors best positioned to address key issues.

## APPENDIX 2: KEY CRITERIA FOR GUIDANCE

We aim for the guidance in this Profile to meet the following criteria. (See Section 2.1 of Barrett et al. 2022 for more detail.) Guidance should be:

1. Actionable and clear enough to be usable in context of the NIST AI RMF, ISO/IEC 23894, or similar frameworks and standards.
2. Usable for key stages of an AI lifecycle, e.g., design, development, test, and evaluation.

3. Meaningful and testable (i.e. "measurable") indicators of AI system's trustworthiness, or at least enable documentability of risk management processes.
4. Compatible with relevant standards or regulations, e.g., from NIST, ISO/IEC, IEEE, or the EU AI Act.
5. Compatible with enterprise risk management (ERM) frameworks typically used by businesses and agencies.
6. Unlikely to be misinterpreted or misapplied by users or other stakeholders in ways that would be net-harmful.
7. Sufficiently future-proof to be applied to AI systems over the next 10 years.

## APPENDIX 3: ROADMAP OF ISSUES TO ADDRESS IN FUTURE VERSIONS OF THE PROFILE

In this section, we list issues that we aim to address in future versions of the Profile. These topics seem important and worth addressing, but available best practices and resources on these topics do not yet meet the above criteria for actionable guidance. We draw much of this initial list and discussion from Section 5 of Barrett et al. (2022). Issues we aim to address include:

- More specific risk-management guidance for specific types of GPAI/foundation models, e.g., image generators or large language models, or specific examples in particular industries or applications.
  - » Such guidance could draw upon more detailed best practices specific to synthetic media (as in PAI 2023a), LLMs (as in Cohere, OpenAI, and AI21 Labs 2022), etc.

- Comprehensive sets of mechanisms or controls to help organizations mitigate identified risks.
  - » We have outlined a number of currently available controls in Section 3.4 of this document, in guidance under the AI RMF Manage function. We aim to incorporate more as they become available. For GPAI/foundation models, additional mechanisms could include ongoing monitoring and evaluation mechanisms that protect against evolving risks from continually learning AI systems.

- Interpretability and explainability methods appropriate for architectures and scales of LLMs and other GPAI/foundation models.
  - » We would like to be able to provide GPAI/foundation model developers with actionable guidance on using interpretability and explainability techniques in specific contexts.

"Inner interpretability" methods for deep neural networks (DNNs) seem to have particular potential given the dominance of the DNN paradigm in GPAI/foundation model development, and the fact that these methods could theoretically help with tasks such as guiding manual modifications, reverse-engineering solutions from models, and detecting latent knowledge of models that could contribute to deceptive behavior. Unfortunately, while interpretability researchers have been exploring a large number of directions, "the field has yet to produce many methods that are competitive in real applications" (Räuker et al. 2023).

- Objectives mis-specification and goal mis-generalization (i.e., misalignment of system behavior with designer goals) characterization and measurement. This might be most relevant for systems whose creation or operation involves an agent, which can be defined as a system that can "adapt their policy if their actions influenced the world in a different way" (Kenton 2022). Risk management considerations for objective mis-specification and goal mis-generalization are important for agentic GPAI/foundation models (autoGPT-style agents) but not necessarily for language models that are not agentic.
  - » An active area of AI safety research aims to develop methods for aligning AI systems during model training, and for validation and verification of AI system objectives alignment (see, e.g., Ouyang et al. 2022, and Bai et al. 2022; for more on challenges and future directions, see, e.g., Section 4 of Hubinger et al. 2019, Gabriel 2020, Section 4.9 of Bommasani et al. 2021, and Section 4 of Hendrycks, Carlini et al. 2021). These methods will be increasingly important as AI systems grow in capability.

- Generality (i.e. breadth of AI applicability/adaptability) characterization and measurement.
  - » While GPAI/foundation models are "general-purpose," the generality and levels of capability of a GPAI/foundation model can be assessed and characterized on a spectrum or on multiple dimensions. If assessment indicates high generality of a GPAI/foundation models, we expect it would be appropriate to conduct more in-depth risk assessment, more assessment of use cases beyond the originally intended use cases, longer time horizons in risk assessment, more continuing assessment, etc. (Ideally, a generality assessment process would be quick and low-cost for AI systems with low generality, while accurately identifying GPAI/foundation models with high generality. Perhaps a simple assessment of generality could be a straightforward extension of our recommendations for identifying potential uses of a model.) For discussion of AI generality as a basic concept, see, e.g., Bommasani et al. (2021). For research on how to assess generality, see, e.g., Hernández-Orallo (2019) and Martínez-Plumed and Hernández-Orallo (2020).

- Recursive improvement potential characterization and measurement.
  - » It could be valuable to assess the degree to which GPAI/foundation models could recursively improve their capabilities, e.g., by editing their own training algorithm code through code generation or using neural architecture search. For such systems, greater levels of safety and control measures could be appropriate. As previously mentioned, recursive improvement potentially could result in GPAI/foundation models with unexpected emergent capabilities and safety-control failures. As the DeepMind paper on the software code-generation AI system AlphaCode stated, "Longer term, code generation could lead to advanced AI risks. Coding capabilities could lead to systems that can recursively write and improve themselves, rapidly leading to more and more advanced systems" (Li et al. 2022). For discussion of related issues, see, e.g., Russell (2019).

- Situational awareness characterization and measurement.
  - » AI systems with situational awareness would be able to make accurate predictions about the humans interacting with them and about their own system architectures, or have other advanced world knowledge or self-knowledge (Ngo, Chan et al. 2022, pp. 3–4). Some initial testing on situational awareness was performed in Perez, Ringer et al. 2022 (pp. 11, 13, 40). The researchers prompted LLMs of different sizes and degrees of RLHF fine-tuning about their awareness of being an AI and certain architectural details, but results were mixed and not strongly conclusive. Situational awareness benchmarks, such as situational awareness dataset (SAD) (Laine et al. 2024) are available, but require further research on a larger sample of models and prompts, as well as investigation of how interventions (e.g. chat finetuning, prompting, model scaling) affect situational awareness. All of the 16 models in this study evaluated using SAD performed better than chance, but below the upper baseline. More study is needed to better understand under what conditions situational awareness might arise in models, how to test for it, and which specific risks and issues are associated in order to recommend actionable guidance for GPAI/foundation model developers on this topic. For additional work on this topic, see, e.g., Berglund et al. (2023).

- Agentic AI systems.
  - » There are great expectations around AI agents in the coming years, with researchers making claims such as "we expect Personal LLM Agents to become a major software paradigm for personal computing devices in the AI era" (Li, Wen et al. 2024 p. 4; Xi et al. 2023). Products and tools claiming agentic capabilities such as Microsoft Autogen and Claude's computer use are rapidly emerging on the market (Wu et al. 2023; Anthropic 2024b).

» Along with the potential benefits, there are also serious risks to consider:
  - Reinforcement learning (RL) agents and long-term planning agents (LTPAs) could develop incentives to undermine human control and deceive humans, as we discuss in our guidance for Map 5.1 in Section 3.
  - There is already evidence that malicious agents can be developed using open-source language models (Lermen et al. 2024).
  - Research shows that simple universal jailbreaks can be developed to jailbreak agents and enable malicious multi-step agent behavior while retaining model capabilities, and some leading LLMs comply with malicious agent requests even without jailbreaking (Andriushchenko et al. 2024a).
» Despite the excitement and concerns around the potential of AI agents, the technology remains in its early stages. The latest version of Microsoft Autogen is 0.4 at the time of this writing, and Claude's computer use capabilities were released with caveats from Anthropic such as "it is still experimental—at times cumbersome and error-prone" and "Claude's current ability to use computers is imperfect. Some actions that people perform effortlessly—scrolling, dragging, zooming—currently present challenges" (Microsoft, n.d.; Anthropic 2024b). Therefore, we do not yet know enough about this technology in practice to be able to provide much actionable guidance around it, and we reserve the project of determining that for a future version of this Profile. Though we do take seriously warnings such as "Developers should not be permitted to build sufficiently capable LTPAs [long-term planning agents], and the resources required to build them should be subject to stringent controls" and advise great caution to developers in the creation of agentic AI systems (Cohen et al. 2024).

- Detection of measurement tampering by a GPAI/foundation model.
  » Measurement tampering is the concern that an AI system could manipulate multiple measurements to create the illusion of good results instead of achieving the desired outcome. There is some early research on detecting measurement tampering (Roger et al. 2023). However, the best detection method found by that research is not robust, and more study is needed in this area in order to recommend actionable guidance for GPAI/foundation model developers on this topic.

- Other measurement/assessment tools for technical specialists testing key aspects of GPAI/foundation model safety, reliability, robustness, interpretability, etc.
  » AI safety researchers are working on a number of other concepts and measurement tools, many of which aim to address challenges in AI safety, reliability, robustness, etc. that are expected to grow as AI systems become increasingly advanced and powerful.

See, e.g., Amodei et al. (2016), Ray et al. (2019), OpenAI (2019c, 2019d), and Hendrycks, Carlini et al. (2021). Measurement of these AI risk-related properties is an active area of research; see, e.g., the discussion and references provided for Direction 1 ("Measuring and forecasting risks") in the 2021 Open Philanthropy request for proposals for projects in AI alignment (Open Philanthrophy 2021, Steinhardt and Barnes 2021).

# Acknowledgments

This work was financially supported by funding from Open Philanthropy and the Survival and Flourishing Fund. We thank Rachel Wesen for workshop organization and support, as well as Chuck Kapelke for editing, web, and media support, and Nicole Hayward for design and formatting of this document. Special thanks to Ann Cleaveland for providing a home and intellectual support for this work at CLTC. We appreciate comments we received from Ashwin Acharya, Anthony Aguirre, Michael Aird, Josh Albrecht, Markus Anderljung, Shahar Avin, Jai Balani, Seth Baum, Kathy Baxter, Haydn Belfield, Alexandra Belias, Sid Ahmed Benraouane, Sawyer Bernath, Stella Biderman, Chad Bieber, Rishi Bommasani, Matt Boulos, Siméon Campos, Ashley Casovan, Jonathan Cefalu, Ze Shen Chin, Peter Cihon, Jonathan Claybrough, Sam Curtis, Christopher Denq, Shaun Ee, Ian Eisenberg, Karson Elmgren, Ellie Evans, Yoav Evenstein, Joel Fischer, Heather Frase, Maximilian Gahntz, Anastasiia Gaidashenko, Andrew Gamino-Cheong, James Gealy, Clíodhna Ní Ghuidhir, Giulia Geneletti, Thomas Krendl Gilbert, Ariel Gil, Rachel Gillum, James Ginns, Amela Gjishti, Jason Green-Lowe, Carlos Ignacio Gutierrez, Gillian Hadfield, Matthew Heyman, Hamish Hobbs, Koen Holtman, Curtis Huebner, Olivia Jimenez, Trent Kannegieter, Divyansh Kaushik, Sonia Katyal, Noam Kolt, Victoria Krakovna, Cullen O'Keefe, Leonie Koessler, Landon Klein, Sabrina Küspert, Yolanda Lannquist, Hanlin Li, Morgan Livingston, Toni Lorente, Liane Lovitt, Kimberly Lucy, Matthijs Maas, Pegah Maham, Richard Mallah, Nicole Nohemi Mauthe, Jeremy McHugh, Nicolas Moës, Malcom Murray, Mina Narayanan, Joe O Brien, Amin Oueslati, Lorenzo Pacchiardi, Henry Papadatos, Milan Patel, Marie-Therese Png, Hadrien Pouget, Christabel Randolph, Ayrton San Joaquin, Krishna Sankar, Daniel Schiff, Tim Schreier, Jonas Schuett, Raymond Sheh, Everett Smith, Genevieve Smith, Joanna Smolinksa, Irene Solaiman, Zeerak Talat, Esther Tetruashvil, Jack Titus, Philip Moreira Tomei, Helen Toner, Risto Uuk, Andrea Vallone, Apostol Vassilev, Sarah Villeneuve, Hjalmar Wijk, as well as others. Any remaining errors are our own. We also appreciate feedback we received on our related work in Barrett et al. (2022); please see the Acknowledgments section of Barrett et al. (2022) for the individuals we thank there for comments on that work.

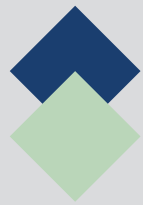

# CLTC

Center for Long-Term
Cybersecurity

UC Berkeley